\PassOptionsToPackage{table,dvipsnames,rgb}{xcolor}
\documentclass{maskwam}

\usepackage{amsmath,amsfonts,bm}

\def\eqref#1{equation~\ref{#1}}

\def\1{\bm{1}}

\DeclareMathAlphabet{\mathsfit}{\encodingdefault}{\sfdefault}{m}{sl}
\SetMathAlphabet{\mathsfit}{bold}{\encodingdefault}{\sfdefault}{bx}{n}

\usepackage{mathtools}
\usepackage{enumitem}
\usepackage{algorithm}
\usepackage{algorithmic}
\usepackage{wrapfig}
\usepackage{adjustbox}
\usepackage{tabularx}
\usepackage{makecell}
\usepackage{siunitx}
\usepackage{float}
\usepackage{diagbox}
\usepackage{dsfont}
\usepackage{multicol}
\usepackage{pifont}
\usepackage{marvosym}
\usepackage{colortbl}

\let\originalincludegraphics\includegraphics
\renewcommand{\includegraphics}[2][]{%
  \IfFileExists{figs_optimized/#2}
    {\originalincludegraphics[#1]{figs_optimized/#2}}
    {\originalincludegraphics[#1]{#2}}%
}

\extrafloats{100}

\newcommand{\cmark}{\textcolor{green!55!black}{\ding{51}}}
\newcommand{\xmark}{\textcolor{red}{\ding{55}}}
\newcommand{\tabcite}[1]{\cite{#1}}

\definecolor{cvprblue}{rgb}{0.21,0.49,0.74}
\definecolor{bestcell}{HTML}{D7E5F2}
\definecolor{seccell}{HTML}{EBF2F8}
\newcommand{\gbest}[1]{\cellcolor{bestcell}{#1}}
\newcommand{\gbestu}[1]{\cellcolor{bestcell}\underline{#1}}
\newcommand{\gsec}[1]{\cellcolor{seccell}#1}


\renewcommand{\floatpagefraction}{0.80}
\titlespacing*{\section}{0pt}{1.6ex plus 0.4ex minus 0.2ex}{0.7ex plus 0.2ex}
\titlespacing*{\subsection}{0pt}{1.3ex plus 0.3ex minus 0.2ex}{0.5ex plus 0.1ex}
\titlespacing*{\subsubsection}{0pt}{1.0ex plus 0.2ex minus 0.1ex}{0.4ex plus 0.1ex}
\titlespacing*{\paragraph}{0pt}{0.5ex plus 0.2ex minus 0.1ex}{0.8em}

\titlelogo{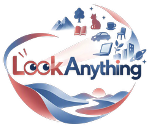}
\title{LookAnything: A Foundation Model for Holistic Gaze Following}

\author[1,2,*,\protect\ddagger]{Jiayu Tang}
\author[2,*,\protect\dagger]{Yuchen Zhou}
\author[3]{Hongsheng Wu}
\author[2]{Hanjing Ye}
\author[4]{Ke Huang}
\author[1]{Cheng Wang}
\author[5]{Yuqin Dai}
\author[2]{Yufei Huang}
\author[2]{Yonggen Ling}
\author[1,\protect\dagger]{Chao Gou}

\affiliation[1]{Sun Yat-Sen University}
\affiliation[2]{Tencent Robotics X}
\affiliation[3]{Futian Laboratory}
\affiliation[4]{Hong Kong Polytechnic University}
\affiliation[5]{Tsinghua University}

\authnote{$^{*}$Equal Contribution.\enspace $^{\dagger}$Corresponding authors.}

\metadata[Project Page]{\url{https://look-anything.github.io/}}
\metadata[GitHub]{\url{https://github.com/Look-Anything/LookAnything}}

\abstract{
    Human gaze provides a rich window into attention and intent, making its understanding essential for applications ranging from social signal analysis to human-robot interaction. Existing methods typically employ highly customized architectures relying on auxiliary inputs like head crops or depth maps, causing cascading errors and inference overhead. Trained on single domains and confined to isolated gaze following with heatmap outputs, they lack flexibility and generalization for diverse interactive scenarios. To bridge this gap, we present \textbf{LookAnything}, a foundation model that unifies diverse gaze-centric tasks for holistic gaze following. Benefiting from large-scale multi-task and cross-domain training, LookAnything flexibly adapts to various settings, generating diverse outputs (points, boxes, heatmaps, and language) and supporting broad gaze-centric localization, description, and reasoning. At its core, we introduce learnable gaze-prior queries to promote gaze-aware representation learning, eliminating the need for auxiliary external inputs. To facilitate this paradigm, we construct \textbf{HoliGaze}, a comprehensive corpus and benchmark comprising 311K question-answer pairs across 5 capability dimensions and 9 distinct tasks. Experiments demonstrate that LookAnything consistently outperforms 33 leading vision-language models on holistic gaze following, while remaining highly competitive with dedicated specialists across 5 conventional target detection benchmarks. Evaluation on unseen datasets and deployment on a real robot further validate its strong out-of-domain generalization and flexible adaptation in real-world scenarios.
}

\begin{document}
\maketitle
\makeatletter
\begingroup
\long\def\@makefntext#1{\noindent\footnotesize\makebox[1.8em][l]{\@makefnmark}#1}%
\def\@thefnmark{\ensuremath{\ddagger}}%
\@footnotetext{Work done during his internship at Tencent Robotics X.}%
\endgroup
\makeatother

\section{Introduction}
\label{sec:intro}
Human gaze is a fundamental nonverbal cue that offers a rich window into human attention, cognitive state, and intent \citep{kleinke1986gaze,frischen2007gaze,grossmann2017eyes}. As a cognitive signal, gaze behavior reveals underlying preferences and social dynamics, which is essential in diverse domains such as social signal analysis and human-robot interaction~\citep{kendon1967some,zuo2026gaze2act, saran2018human}. Driven by these demands, the core visual task of gaze following has emerged to identify the exact spatial location a person is focusing on \citep{recasens2015they, chong2020detecting}. This capability is crucial for developing responsive embodied agents and enabling human-centric interactive applications.

\begin{figure}[!t]
  \centering
  \includegraphics[width=1.0\textwidth]{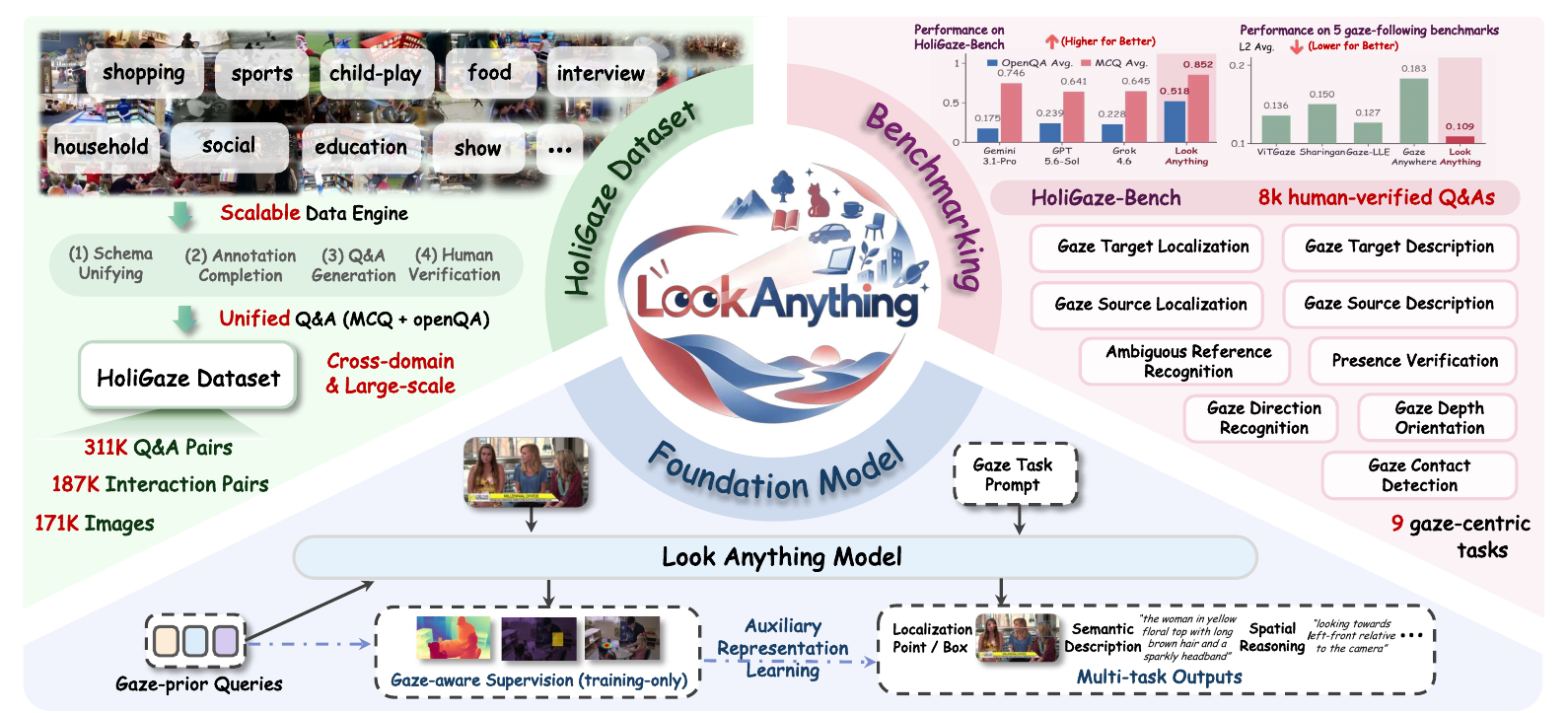}
  \caption{\textbf{Overview.} \textbf{(a) Left:} Sourced from diverse scenarios, we design a scalable data engine to transform heterogeneous annotations into a unified Q\&A format. The resulting large-scale \textit{HoliGaze} dataset provides the essential data foundation for holistic gaze following. \textbf{(b) Down:} \textit{LookAnything} is a generalist foundation model that unifies diverse gaze-centric tasks. By introducing learnable gaze-prior queries, it eliminates the need for computationally expensive external inputs and flexibly generates versatile outputs (points, boxes, heatmaps, and language) for broad localization, description, and reasoning. \textbf{(c) Right:} We establish \textit{HoliGaze-Bench} across 9 distinct gaze-centric tasks to systematically evaluate 33 leading VLMs. \textit{LookAnything} demonstrates superior performance over other VLMs on this comprehensive benchmark, while achieving highly competitive results against dedicated specialists on 5 conventional gaze following datasets.}
  \label{fig:teaser}
\end{figure}

Existing specialist methods typically employ highly customized architectures relying on auxiliary inputs like head crops or depth maps~\citep{tonini2022multimodal, gupta2022modular}, which introduce cascading errors and inference overhead~\citep{ryan2025gaze, tu2022end}. They are often trained on single domains and strictly confined to the isolated task of predicting gaze target heatmaps \citep{miao2023patch, song2024vitgaze}. Such coarse outputs lack instance-level semantics, rendering them inflexible and difficult to generalize across diverse interactive scenarios and downstream reasoning tasks. While recent efforts have incorporated gaze to vision-language models (VLMs) \citep{wang2025vl4gaze, mathew2026gazevlm}, they remain limited in task scope and underexploit gaze-centric priors.

To bridge this gap, we present \textbf{LookAnything}, a foundation model that unifies diverse gaze-centric tasks within a single multimodal framework. Benefiting from large-scale multi-task and cross-domain training, LookAnything flexibly adapts to various settings, generating diverse outputs (including points, bounding boxes, heatmaps, and natural language) while supporting broad gaze-centric localization, description, and reasoning.
As shown in Figure~\ref{fig:teaser}, rather than relying on computationally expensive external inputs, we introduce a novel set of learnable gaze-prior queries. These queries interact with visual features to implicitly internalize scene depth, source-head orientation, and target spatial geometry directly into the latent space. This design elegantly promotes intrinsic gaze-aware representation learning while maintaining the efficiency of foundation models.

\begin{wraptable}[8]{l}{0.48\textwidth}
\vspace{-6pt}
\centering
\caption{Comparison with existing gaze-follow datasets.}
\label{tab:holigaze_dataset_comparison}
\setlength{\tabcolsep}{3.5pt}
\renewcommand{\arraystretch}{1.04}
\footnotesize
\resizebox{\linewidth}{!}{%
\begin{tabular}{@{}lcc|cc|ccc@{}}
\toprule
\multirow{2}{*}{Dataset} &
  \multirow{2}{*}{\#Samples} &
  \multirow{2}{*}{\#Tasks} &
  \multicolumn{2}{c|}{Reference Type} &
  \multicolumn{3}{c}{Output} \\
\cmidrule(lr){4-5} \cmidrule(lr){6-8}
 & & & Text & Box & Box & Point & Text \\
\midrule
GazeFollow~\tabcite{recasens2015they}
  & 130K & 1
  & \xmark & \cmark
  & \xmark & \cmark & \xmark \\
VAT~\tabcite{chong2020detecting}
  & 165K & 2
  & \xmark & \cmark
  & \xmark & \cmark & \xmark \\
VL4Gaze~\tabcite{wang2025vl4gaze}
  & 489K & 4
  & \cmark & \xmark
  & \xmark & \cmark & \cmark \\
Gaze-Co~\tabcite{cao2026gaze}
  & 120K & 1
  & \cmark & \xmark
  & \cmark & \cmark & \xmark \\
\midrule
\rowcolor{blue!10}
\textbf{HoliGaze (Ours)}
  & \textbf{311K} & \textbf{9}
  & \cmark & \cmark
  & \cmark & \cmark & \cmark \\
\bottomrule
\end{tabular}%
}
\vspace{-4pt}
\end{wraptable}
Facilitating this unified paradigm demands comprehensive data and evaluation protocols. However, existing benchmarks remain fragmented and limited in scope~\citep{recasens2015they, cao2026gaze, wang2026enhancing}. To address this, we construct \textbf{HoliGaze}, a comprehensive corpus and benchmark for holistic gaze following. HoliGaze standardizes 311K high-quality Q$\&$A pairs across 5 distinct capability dimensions (e.g., gaze target understanding, source understanding, spatial reasoning, social interaction, and query answerability) and 9 diverse gaze-centric tasks. This provides the critical foundation for training and systematically evaluating gaze-aware generalist models.

Extensive experiments demonstrate that LookAnything substantially outperforms 33 leading VLMs on HoliGaze-Bench, while remaining highly competitive with dedicated specialists across 5 conventional benchmarks. Furthermore, it demonstrates impressive out-of-domain generalization on two unseen datasets. Crucially, real-world deployment on a Unitree G1 robot further validates its robust generalization and flexible adaptation to complex physical scenarios. 
Our key contributions are as follows:
\begin{itemize}
    \item We introduce LookAnything, a foundation model that leverages learnable gaze-prior queries for intrinsic gaze-aware representation learning. Coupled with large-scale multi-task and cross-domain training, it enables flexible adaptation across various scenarios, task settings, and output formats without external auxiliary inputs.
    \item We construct HoliGaze, to the best of our knowledge, the most comprehensive corpus and benchmark for holistic gaze following to date. By standardizing 311K Q$\&$A pairs across 5 capability dimensions and 9 distinct gaze-centric tasks, it addresses the fragmentation of existing datasets to facilitate large-scale training and systematic evaluation.
    \item LookAnything consistently outperforms 33 leading VLMs on HoliGaze-Bench and remains highly competitive with specialists across 5 conventional benchmarks. Both quantitative evaluation on unseen datasets and real-world robotic deployment further validate its robust generalization and flexible adaptation to complex physical scenarios.
\end{itemize}

\begin{figure*}[t]
    \centering
    \includegraphics[width=1.0\textwidth]{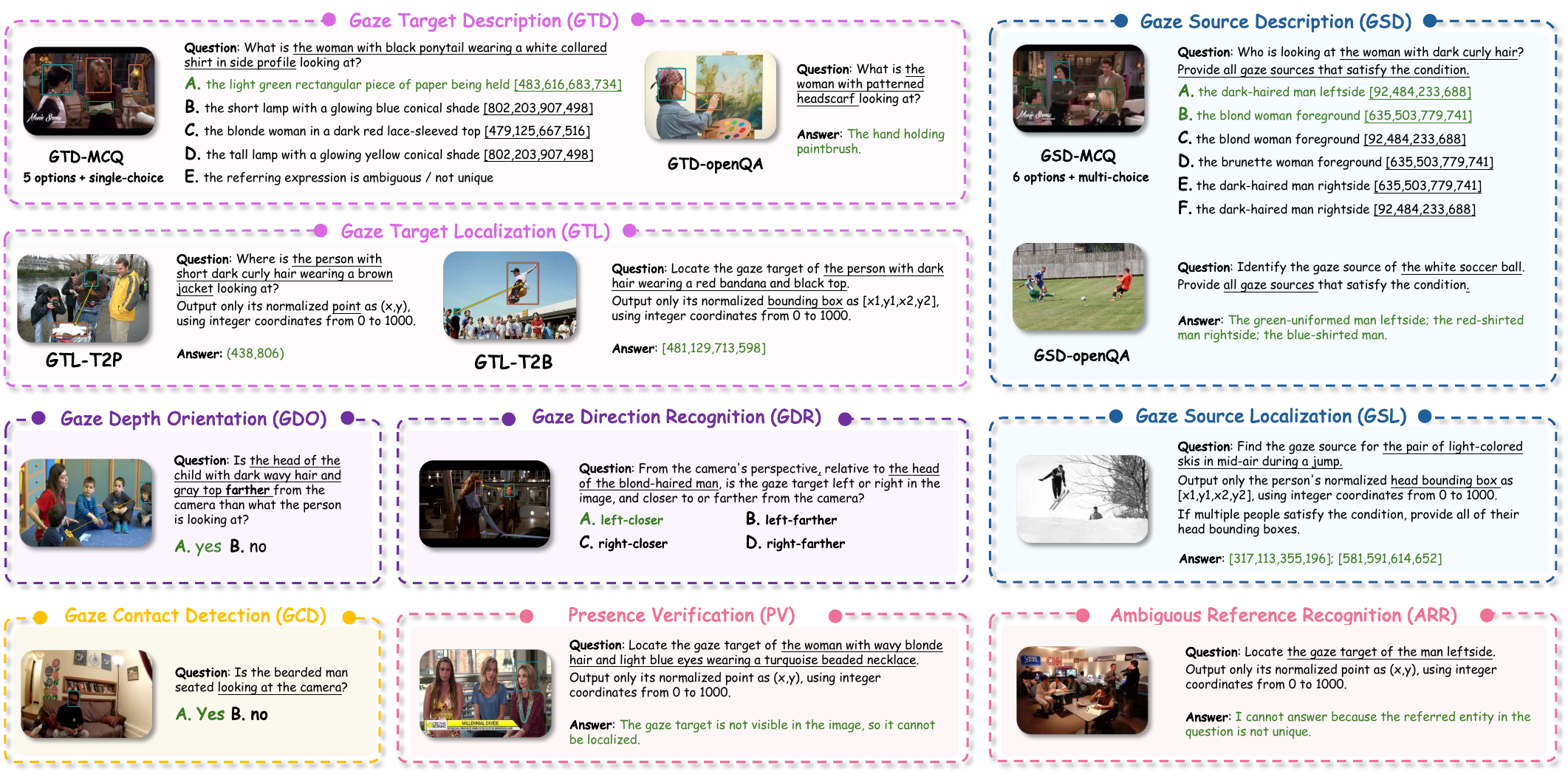}
        \caption{\textbf{Task taxonomy and Q$\&$A examples of HoliGaze.} HoliGaze covers 5 capability dimensions and 9 gaze-centric tasks, including both multiple-choice Q$\&$A (MCQ) and open-ended Q$\&$A (openQA). Each panel represents a type of gaze-centric task, with panels of the same color belonging to the same capability dimension.}
        \label{fig:task_taxonomy}
        \vspace{-4mm}
    \end{figure*}
    
    \section{The HoliGaze Dataset}
    \label{sec:benchmark}
    We propose HoliGaze, a diverse and comprehensive dataset for holistic gaze following. It comprises \textbf{HoliGaze-Corpus} (303K training samples) and \textbf{HoliGaze-Bench} (8K human-verified evaluation samples), totaling 311K Q$\&$A pairs.
    
    \subsection{Task Taxonomy}
    \label{subsec:taxonomy}
    As illustrated in Figure~\ref{fig:task_taxonomy}, HoliGaze defines 9 tasks grouped into 5 dimensions. Open-ended Q$\&$A (openQA) covers generative localization and description, while multiple-choice Q$\&$A (MCQ) supports stable automatic scoring.
    
    \begin{enumerate}
        \setlength{\itemsep}{0pt}
        \setlength{\parsep}{0pt}
        \setlength{\topsep}{0.25em}
        \setlength{\partopsep}{0pt}
        \item \textbf{Gaze Target Understanding:} Given a textual or box reference of the gaze source person, \textit{Gaze Target Localization (GTL)} grounds the target as a 2D point (GTL-T2P) or bounding box (GTL-T2B). \textit{Gaze Target Description (GTD)} requires a natural-language description of the attended entity. Training additionally includes a box-referred variant (GTL-B2P).
        \item \textbf{Gaze Source Understanding:} Given a textual reference to the gaze target, \textit{Gaze Source Localization (GSL)} predicts the head boxes of all observers, and \textit{Gaze Source Description (GSD)} describes their visual attributes.
        \item \textbf{Gaze Spatial Reasoning:} \textit{Gaze Depth Orientation (GDO)} asks whether the observer's head is closer to or farther from the camera than the target. \textit{Gaze Direction Recognition (GDR)} classifies the gaze into four camera-relative directions.
        \item \textbf{Gaze Social Interaction:} \textit{Gaze Contact Detection (GCD)} is a binary MCQ asking whether a specified person is making eye contact with the camera wearer.
        \item \textbf{Gaze Query Answerability:} Models must refuse ill-posed queries. \textit{Ambiguous Reference Recognition (ARR)} requires refusal when the referring expression matches multiple entities, and \textit{Presence Verification (PV)} requires refusal when the target lies outside the frame.
    \end{enumerate}

    \subsection{Scalable Data Engine}
    \label{subsec:data_engine}
\begin{figure*}[t]
      \centering
        \includegraphics[width=1.0\textwidth]{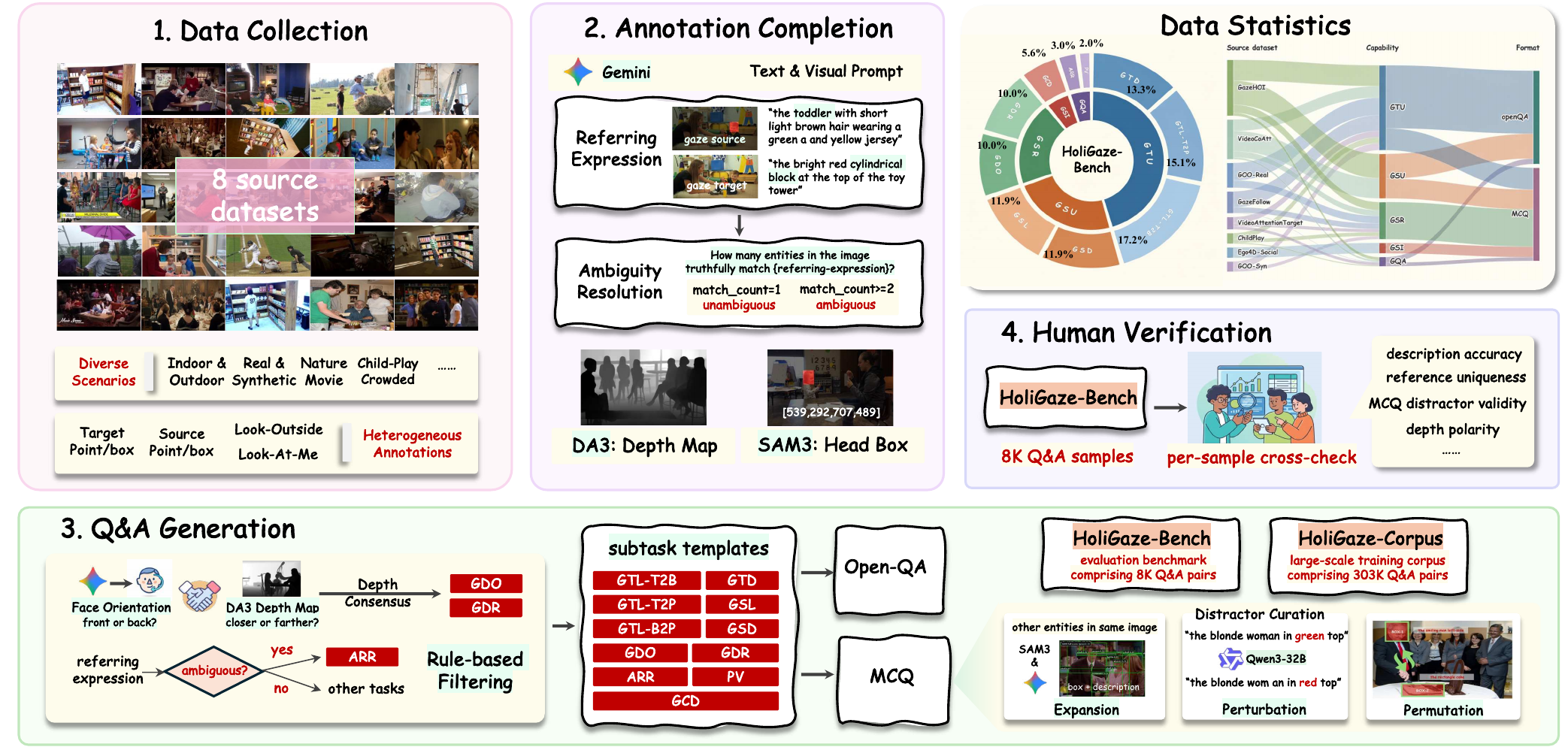}
          \caption{\textbf{Data engine and dataset statistics of HoliGaze.} The data engine converts 8 source datasets into unified Q$\&$A samples via four stages: (1) unifying heterogeneous records into a shared schema, (2) completing missing annotations using off-the-shelf models, (3) generating openQA and MCQ pairs via task-specific templates, and (4) conducting human verification. \textbf{Top-right:} Statistics of HoliGaze-Bench, which comprises 8K evaluation pairs quota-sampled across 9 tasks with a balanced 1:1 ratio of openQA to MCQ.}
          \label{fig:data_pipeline}
      \end{figure*}

    We design a multi-stage data engine that converts heterogeneous gaze annotations from 8 source datasets into unified Q$\&$A pairs, as illustrated in Figure~\ref{fig:data_pipeline}. First, we unify records into a shared schema and use off-the-shelf models (SAM3~\citep{carion2026sam}, Depth Anything v3~\citep{lin2025depth}, and Gemini~\citep{comanici2025gemini}) to complete missing fields such as head masks, scene depth, and referring expressions. Next, task-specific templates instantiate the schema into openQA and MCQ formats. We synthesize challenging MCQ distractors via entity expansion, attribute perturbation, and box permutation, while filtering ambiguous or out-of-frame targets for answerability tasks (ARR/PV). Finally, to prevent data leakage, we strictly follow each source's native split to curate the training corpus and evaluation benchmark. HoliGaze-Bench undergoes rigorous instance-level human verification, yielding a low modification rate ($0.70\%$) that indicates strong agreement between the automated outputs and human judgment. \textit{We defer the comprehensive details of data engine to Appendix~D.}
    
    \subsection{Dataset Statistics}
    \label{subsec:stats_metrics}
    In total, HoliGaze comprises 311K Q$\&$A pairs, 171K images, and 187K distinct gaze interactions. The dataset is systematically partitioned into two subsets: (1) HoliGaze-Corpus, providing 303K training samples, and (2) HoliGaze-Bench, containing 8K evaluation pairs over 6,143 images. To ensure comprehensive evaluation, the benchmark is quota-sampled across 9 distinct gaze-centric tasks with a balanced 1:1 ratio of open-ended to multiple-choice questions. As compared in Table~\ref{tab:holigaze_dataset_comparison}, HoliGaze is the only dataset that simultaneously supports multiple referring types and diverse output formats across all task dimensions. \textit{Detailed statistical distributions are deferred to Appendix~C.}
\section{The LookAnything Model}

\subsection{Model Architecture}
\label{sec:model_architecture}

We formulate holistic gaze following as instruction-conditioned autoregressive generation. Given an image $I$ and a language prompt $P$, LookAnything models the conditional distribution over an open-form answer $Y$: $p_\theta(Y\mid I,P)=\prod_{t=1}^{N_Y}p_\theta(y_t \mid I,P,y_{<t})$, where $Y$ unifies localization, description, and reasoning into a single token sequence.

\begin{figure*}[t]
  \centering
      \includegraphics[width=1.0\textwidth]{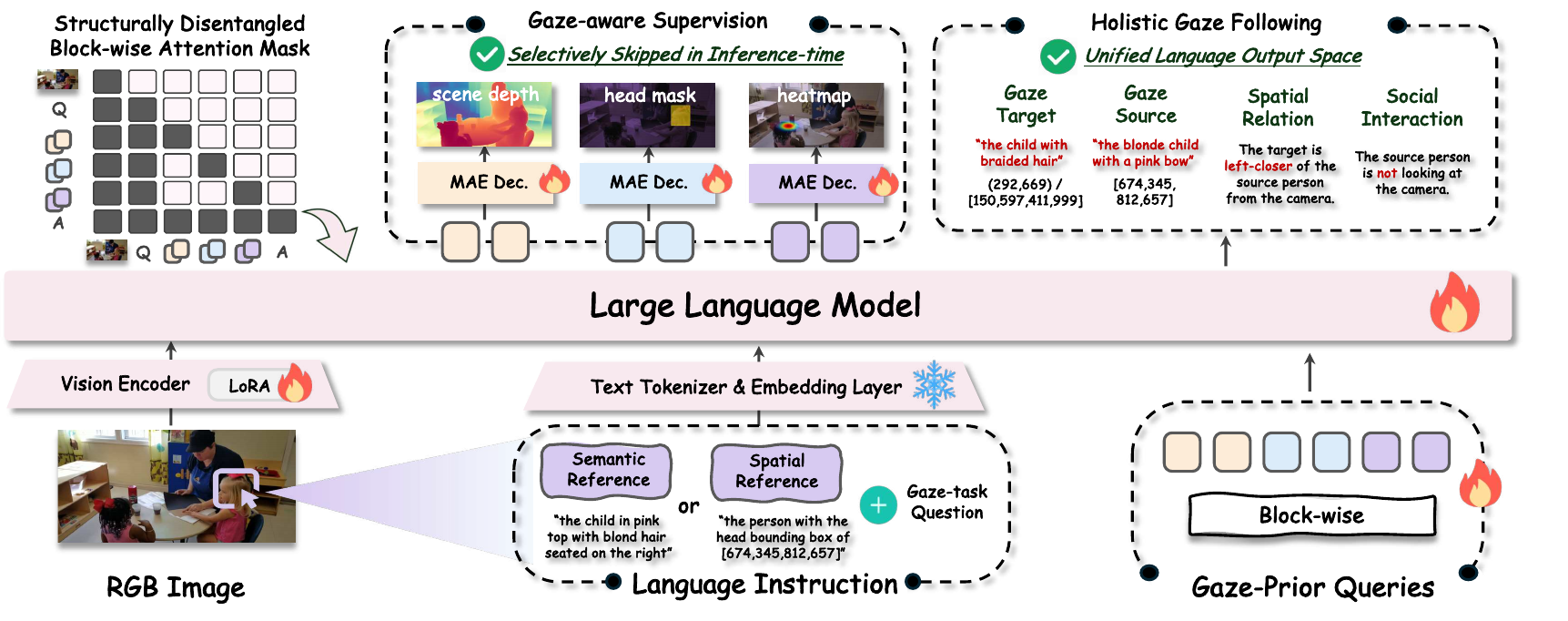}
      \caption{\textbf{Model architecture overview.} LookAnything appends learnable gaze-prior queries to the multimodal input sequence to learn gaze-aware representations via auxiliary geometric supervision (scene depth, source head grounding, and gaze target heatmap). The language head then autoregressively decodes the answer in a unified output space. The auxiliary branches can be selectively skipped during inference. \textbf{Top-left:} The structurally disentangled block-wise attention mechanism isolates functional specialization by preventing heterogeneous query groups from attending to one another.}
      \label{fig:framework}
      \vspace{-2mm}
  \end{figure*}

As illustrated in Figure~\ref{fig:framework}, LookAnything builds upon the Qwen3-VL \citep{bai2025qwen3} architecture. Modality-specific encoders first map the image and prompt into latent tokens. To explicitly inject gaze-aware spatial priors, we introduce learnable continuous queries $\mathcal{Q} = [Q^d, Q^h, Q^g]$ designated for scene depth, source head grounding, and target spatial geometry. These queries are appended to the multimodal embeddings. 

The large language model backbone $\mathcal{M}$ integrates these inputs into compact latent gaze-context representations $H = \mathcal{M}(\mathcal{E}_{v}(I), \mathcal{E}_{t}(P) \mid \mathcal{Q})$. A set of lightweight auxiliary decoders $\mathcal{D}$ projects these query states into explicit geometric predictions, while the language head autoregressively generates the answer $Y$. During inference, the auxiliary decoders can be selectively skipped to save computation. \textit{Detailed architectural configurations are provided in Appendix~B.3.}

\begin{table*}[t]
\centering
\caption{\textbf{Results on HoliGaze-Bench.} Group-wise best/second-best are in \colorbox{bestcell}{color}/\colorbox{seccell}{color}; overall best/second-best are \textbf{bolded}/\underline{underlined}.}
\label{tab:holigaze8k_mllm}
\setlength{\tabcolsep}{2.4pt}
\renewcommand{\arraystretch}{1.00}
\resizebox{\textwidth}{!}{%
\begin{tabular}{l cc cc ccc c ccc ccc cc}
\toprule
\multirow{3}{*}{\textbf{Methods}} & \multicolumn{2}{c}{\textbf{Avg.}} & \multicolumn{2}{c}{GTL} & \multicolumn{3}{c}{GTD} & GSL & \multicolumn{3}{c}{GSD} & GDO & GDR & GCD & ARR & PV \\
\cmidrule(lr){2-3} \cmidrule(lr){4-5} \cmidrule(lr){6-8} \cmidrule(lr){9-9} \cmidrule(lr){10-12} \cmidrule(lr){13-13} \cmidrule(lr){14-14} \cmidrule(lr){15-15} \cmidrule(lr){16-16} \cmidrule(lr){17-17}
& \multirow{2}{*}{\textbf{openQA}} & \multirow{2}{*}{\textbf{MCQ}} & T2P & T2B & MCQ & \multicolumn{2}{c}{openQA} & openQA & MCQ & \multicolumn{2}{c}{openQA} & MCQ & MCQ & MCQ & openQA & openQA \\
\cmidrule(lr){4-4} \cmidrule(lr){5-5} \cmidrule(lr){6-6} \cmidrule(lr){7-8} \cmidrule(lr){9-9} \cmidrule(lr){10-10} \cmidrule(lr){11-12} \cmidrule(lr){13-13} \cmidrule(lr){14-14} \cmidrule(lr){15-15} \cmidrule(lr){16-16} \cmidrule(lr){17-17}
& & & $L_2$$\downarrow$ & mIoU$\uparrow$ & Acc$\uparrow$ & R-L$\uparrow$ & B-4$\uparrow$ & mIoU$\uparrow$ & F1$\uparrow$ & R-L$\uparrow$ & B-4$\uparrow$ & Acc$\uparrow$ & Acc$\uparrow$ & Acc$\uparrow$ & F1$\uparrow$ & F1$\uparrow$ \\
\midrule
\rowcolor{cvprblue!8}
\multicolumn{17}{c}{\textit{Closed-source MLLMs}} \\
Gemini-2.5-Flash~\tabcite{comanici2025gemini} & 0.101 & 0.537 & 0.237 & 0.032 & 0.432 & 0.152 & 0.065 & 0.074 & 0.493 & 0.267 & 0.109 & 0.557 & 0.385 & 0.816 & 0.000 & 0.111 \\
Gemini-2.5-Pro~\tabcite{comanici2025gemini} & 0.178 & 0.644 & 0.189 & 0.085 & 0.532 & 0.159 & 0.065 & 0.197 & 0.588 & 0.227 & 0.109 & 0.792 & 0.575 & 0.731 & 0.049 & 0.529 \\
Gemini-3.7-Flash~\tabcite{comanici2025gemini} & \gbestu{0.265} & \gbestu{0.770} & \gbestu{0.131} & 0.038 & \gbestu{0.609} & \gsec{0.237} & 0.149 & 0.047 & \gsec{0.717} & \gsec{0.276} & \gbestu{0.124} & \gbestu{0.901} & \gbestu{0.785} & \gbestu{0.840} & \gbestu{0.621} & 0.630 \\
Gemini-3.1-Pro~\tabcite{comanici2025gemini} & 0.175 & \gsec{0.746} & 0.194 & 0.029 & \gsec{0.581} & 0.137 & 0.102 & 0.040 & \gbestu{0.742} & 0.204 & 0.107 & \gsec{0.875} & \gsec{0.734} & 0.800 & 0.140 & 0.642 \\
DeepSeek-V4-Flash~\tabcite{xu2026deepseek} & 0.161 & 0.394 & 0.341 & 0.023 & 0.258 & 0.125 & 0.115 & 0.026 & 0.314 & 0.137 & 0.067 & 0.615 & 0.292 & 0.493 & 0.000 & \gsec{0.797} \\
GLM-5.3-Flash~\tabcite{zeng2026glm} & 0.160 & 0.639 & 0.254 & \gsec{0.090} & 0.515 & 0.175 & 0.097 & \gsec{0.247} & 0.606 & \gbestu{0.277} & 0.117 & 0.770 & 0.666 & 0.640 & 0.095 & 0.180 \\
Grok-4.6~\tabcite{xiaimodelcard_card-4p6-4cd2dc57} & 0.228 & 0.645 & 0.241 & 0.076 & 0.505 & 0.236 & \gsec{0.153} & 0.128 & 0.586 & 0.205 & 0.092 & 0.785 & 0.642 & 0.707 & \gsec{0.404} & 0.533 \\
Qwen3-Max~\tabcite{bai2025qwen3} & 0.171 & 0.405 & 0.342 & 0.019 & 0.271 & 0.161 & 0.096 & 0.014 & 0.306 & 0.189 & 0.086 & 0.580 & 0.274 & 0.593 & 0.000 & \gbestu{0.800} \\
GPT-5.6-Sol~\tabcite{singh2025openai} & \gsec{0.239} & 0.641 & \gsec{0.176} & \gbest{0.116} & 0.510 & \gbest{0.257} & \gbest{0.179} & \gbest{0.248} & 0.513 & 0.267 & \gsec{0.122} & 0.785 & 0.573 & \gsec{0.824} & 0.048 & 0.672 \\
\midrule
\rowcolor{cvprblue!8}
\multicolumn{17}{c}{\textit{Open-source MLLMs}} \\
Qwen3-VL-2B~\tabcite{bai2025qwen3} & 0.081 & 0.465 & 0.387 & 0.040 & 0.394 & 0.210 & 0.116 & 0.081 & 0.483 & 0.128 & 0.061 & 0.550 & 0.237 & 0.660 & 0.000 & 0.011 \\
Qwen3.5-2B~\tabcite{bai2025qwen3} & 0.133 & 0.421 & 0.372 & 0.070 & 0.377 & 0.241 & 0.153 & 0.104 & 0.340 & 0.140 & 0.045 & 0.461 & 0.209 & 0.720 & \gsec{0.182} & 0.126 \\
InternVL3-2B~\tabcite{zhu2025internvl3} & 0.090 & 0.453 & 0.610 & 0.020 & 0.370 & 0.212 & 0.119 & 0.045 & 0.444 & 0.235 & 0.092 & 0.532 & 0.249 & 0.671 & 0.000 & 0.000 \\
InternVL3.5-2B~\tabcite{wang2025internvl3} & 0.100 & 0.463 & 0.536 & 0.005 & 0.434 & 0.195 & 0.123 & 0.177 & 0.495 & 0.218 & 0.081 & 0.590 & 0.196 & 0.602 & 0.000 & 0.000 \\
RynnBrain-2B~\tabcite{dang2026rynnbrain} & 0.128 & 0.557 & 0.369 & 0.000 & 0.494 & 0.153 & 0.115 & 0.000 & 0.540 & 0.043 & 0.053 & 0.711 & 0.334 & 0.704 & 0.000 & 0.658 \\
Embodied-R1-3B~\tabcite{yuan2026embodiedr1} & 0.170 & 0.469 & 0.765 & 0.004 & 0.389 & 0.136 & 0.080 & 0.022 & 0.438 & 0.160 & 0.069 & 0.546 & 0.292 & 0.682 & 0.115 & \gsec{0.777} \\
Qwen3-VL-4B~\tabcite{bai2025qwen3} & 0.136 & 0.550 & 0.329 & 0.058 & 0.496 & 0.223 & 0.127 & 0.121 & 0.568 & 0.223 & 0.102 & 0.545 & 0.352 & 0.789 & 0.000 & 0.233 \\
Qwen2.5-VL-7B~\tabcite{bai2025qwen3} & 0.126 & 0.506 & 0.591 & 0.008 & 0.403 & 0.142 & 0.066 & 0.038 & 0.445 & 0.174 & 0.074 & 0.522 & 0.385 & 0.773 & \gbest{0.261} & 0.244 \\
MiMo-Embodied-7B~\tabcite{hao2025mimo} & 0.100 & 0.210 & 0.859 & 0.004 & 0.120 & 0.101 & 0.042 & 0.005 & 0.001 & 0.121 & 0.031 & 0.268 & 0.044 & 0.616 & 0.000 & 0.496 \\
Pelican-VL-7B~\tabcite{zhang2025pelican} & 0.077 & 0.532 & 0.732 & 0.000 & 0.379 & 0.189 & 0.107 & 0.000 & 0.450 & 0.128 & 0.064 & 0.721 & 0.352 & 0.758 & 0.118 & 0.012 \\
RoboBrain-2.0-7B~\tabcite{team2025robobrain} & 0.072 & 0.561 & 0.511 & 0.004 & 0.391 & 0.184 & 0.095 & 0.002 & 0.391 & 0.180 & 0.082 & \gbest{0.757} & 0.535 & 0.731 & 0.025 & 0.000 \\
VeBrain-7B~\tabcite{luo2025visual} & 0.201 & 0.496 & 0.476 & 0.026 & 0.399 & \gbestu{0.285} & 0.148 & 0.058 & 0.508 & 0.121 & 0.052 & 0.449 & 0.362 & 0.760 & 0.157 & 0.760 \\
InternVL3-8B~\tabcite{zhu2025internvl3} & 0.213 & 0.548 & 0.326 & 0.044 & 0.468 & 0.200 & 0.109 & 0.179 & 0.490 & \gbest{0.264} & \gsec{0.107} & 0.551 & 0.449 & 0.780 & 0.025 & \gsec{0.777} \\
InternVL3.5-8B~\tabcite{wang2025internvl3} & 0.192 & 0.467 & 0.227 & 0.088 & 0.452 & 0.159 & 0.108 & 0.177 & 0.304 & 0.183 & 0.088 & 0.613 & 0.241 & 0.724 & 0.000 & 0.734 \\
Embodied-R1.5-8B~\tabcite{yuan2026embodied} & 0.186 & 0.617 & 0.246 & 0.068 & 0.508 & 0.215 & 0.128 & 0.241 & \gsec{0.611} & 0.164 & 0.091 & 0.725 & 0.506 & 0.736 & 0.000 & 0.583 \\
RoboBrain-2.5-8B~\tabcite{tan2026robobrain} & 0.171 & 0.569 & 0.319 & 0.049 & 0.488 & 0.132 & 0.053 & 0.237 & 0.566 & 0.210 & 0.081 & 0.589 & 0.411 & 0.793 & 0.000 & 0.603 \\
RynnBrain-9B~\tabcite{dang2026rynnbrain} & 0.145 & 0.322 & 0.687 & 0.007 & 0.154 & 0.091 & 0.036 & 0.068 & 0.149 & 0.131 & 0.049 & 0.527 & 0.198 & 0.580 & 0.000 & \gbest{0.779} \\
Qwen3.5-27B~\tabcite{bai2025qwen3} & 0.216 & 0.498 & \gbest{0.173} & \gsec{0.137} & 0.339 & 0.262 & 0.167 & 0.340 & 0.060 & 0.210 & 0.095 & 0.724 & \gbest{0.576} & 0.791 & 0.049 & 0.467 \\
Qwen3.6-27B~\tabcite{bai2025qwen3} & \gbest{0.223} & 0.487 & 0.229 & 0.112 & 0.340 & 0.270 & \gbestu{0.181} & \gbestu{0.395} & 0.048 & 0.211 & 0.095 & 0.724 & 0.534 & 0.789 & 0.025 & 0.498 \\
Qwen3.8-27B~\tabcite{bai2025qwen3} & \gsec{0.222} & 0.401 & \gsec{0.223} & \gbestu{0.152} & 0.159 & \gsec{0.282} & \gsec{0.179} & 0.294 & 0.032 & 0.221 & 0.085 & 0.640 & 0.365 & \gsec{0.811} & 0.025 & 0.539 \\
Hy-Embodied-30B~\tabcite{wang2026hy} & 0.054 & 0.603 & 0.262 & 0.000 & 0.434 & 0.079 & 0.100 & 0.002 & 0.542 & 0.171 & 0.079 & 0.748 & 0.500 & 0.791 & 0.000 & 0.000 \\
Qwen3-VL-32B~\tabcite{bai2025qwen3} & 0.213 & \gsec{0.641} & 0.255 & 0.108 & \gbest{0.569} & 0.221 & 0.135 & \gsec{0.366} & \gbest{0.667} & 0.239 & 0.103 & 0.693 & 0.482 & 0.796 & 0.000 & 0.529 \\
InternVL3-38B~\tabcite{zhu2025internvl3} & 0.155 & \gbest{0.655} & 0.336 & 0.092 & \gsec{0.552} & 0.185 & 0.112 & 0.253 & 0.592 & \gsec{0.259} & \gbest{0.113} & \gsec{0.755} & \gsec{0.557} & \gbest{0.818} & 0.025 & 0.197 \\
InternVL3.5-38B~\tabcite{wang2025internvl3} & 0.168 & 0.577 & 0.694 & 0.092 & 0.462 & 0.205 & 0.098 & 0.290 & 0.478 & 0.210 & 0.088 & 0.728 & 0.429 & 0.787 & 0.000 & 0.358 \\
\midrule
\rowcolor{cvprblue!15}
\textbf{LookAnything (2B, Ours)} & \textbf{0.518} & \textbf{0.852} & \textbf{0.094} & \textbf{0.333} & \textbf{0.764} & \textbf{0.459} & \textbf{0.283} & \textbf{0.610} & \textbf{0.865} & \textbf{0.466} & \textbf{0.260} & \textbf{0.924} & \textbf{0.863} & \textbf{0.847} & \textbf{0.805} & \textbf{0.932} \\
\bottomrule
\end{tabular}%
}
\end{table*}

\subsection{Gaze Context-Aware Representation Learning}
\label{sec:representation_learning}

Instead of relying on explicit external inputs (e.g., depth maps or head crops) which introduce cascading errors and latency, LookAnything internalizes critical gaze priors into latent representations via auxiliary supervision. \textit{Detailed loss formulations are deferred to Appendix~B.4.}

\paragraph{Scene Depth Estimation.}
Scene depth provides essential 3D structural context, which is crucial for understanding the spatial relationship between the gaze source and the target. The depth query states $H^d$ are projected by a lightweight MAE-style~\citep{he2022masked} decoder $\mathcal{D}_d$ to predict the relative monocular scene depth $\hat{D} = \mathcal{D}_d(H^d)$. This prediction is supervised using a scale-invariant logarithmic loss to emphasize relative structural depth.

\paragraph{Source Head Grounding.}
Accurate head localization is a prerequisite for gaze target estimation. The head query states $H^h$ are decoded by an MAE-style decoder $\mathcal{D}_h$ to output a binary source head mask $\hat{M} = \mathcal{D}_h(H^h)$, which is optimized via a combination of Binary Cross-Entropy and soft Dice loss to mitigate foreground-background imbalance.

\paragraph{Target Spatial Prediction.}
The spatial distribution of the gaze target is a vital geometric cue. By interacting with the contextualized visual features $F_v$ via cross-attention, the gaze queries $H^g$ are decoded by an MAE-style decoder into a dense gaze heatmap $\hat{A} = \mathcal{D}_a(H^g, F_v)$, supervised by KL-divergence and Mean Squared Error. Concurrently, dedicated regression heads predict geometric primitives, specifically a normalized target point $\hat{p} = \mathcal{D}_p(H^g)$ and a bounding box $\hat{b} = \mathcal{D}_b(H^g)$, optimized via Huber and GIoU losses.

\paragraph{Structurally Disentangled Block-wise Attention.}
If heterogeneous query groups freely attended to one another, dense depth gradients or low-level head contours might interfere with gaze target representations, creating shortcut dependencies that degrade semantic generalization. To strictly isolate functional specialization, we introduce a block-wise structured attention mask (top-left of Figure~\ref{fig:framework}). Each query group independently attends to the shared multimodal prompt but remains isolated from other query groups. Crucially, the downstream autoregressive language tokens jointly attend to all contextualized gaze representations, elegantly integrating them for holistic gaze-follow reasoning.

\section{Experiments}
\label{sec:experiments}

\subsection{Experimental Settings}
\label{subsec:experimental_settings}

\paragraph{Evaluation Metrics.}
For HoliGaze-Bench, point and box localization are evaluated using average $L_2$ distance and mean IoU (mIoU), respectively. OpenQA generation is measured by ROUGE-L and BLEU-4, while multiple-choice questions (MCQs) are assessed via accuracy, with set-F1 applied for multi-select formats. For native benchmarks, following established protocols~\citep{recasens2015they, tafasca2024toward}, we report average $L_2$ distance and heatmap AUC, alongside dataset-specific metrics such as Min $L_2$ on GazeFollow and GazeAcc on GazeHOI. For LookAnything, AUC is computed using the auxiliary query-based heatmap output, while all other metrics are derived solely from the generated text.

\paragraph{Implementation Details.}
LookAnything is built upon Qwen3-VL-2B \citep{bai2025qwen3}. We adapt the vision encoder via LoRA \citep{hu2021lora} (rank=16, $\alpha$=32) and jointly train the language backbone, gaze-prior queries, and auxiliary decoders. We optimize the model using AdamW \citep{loshchilov2017decoupled} with a global batch size of 32 for 2 epochs on 8 NVIDIA H20 GPUs. The learning rate schedule uses a linear warmup (3\% of total steps) followed by cosine decay, with a peak learning rate of $1\times 10^{-5}$ for the language backbone and $1\times 10^{-4}$ for the auxiliary heads and vision LoRA. We use 303K samples from HoliGaze-Corpus for training, which comprises data from GazeFollow \citep{recasens2015they}, VAT \citep{chong2020detecting}, ChildPlay \citep{tafasca2023childplay}, GazeHOI \citep{tafasca2024toward}, Ego4D-Social \citep{grauman2022ego4d}, VideoCoAtt \citep{fan2018inferring}, and GOO-syn \citep{tomas2021goo}. \textit{More implementation details are provided in Appendix~B.}

\subsection{Results on HoliGaze-Bench}
\label{subsec:results_holigaze}
As depicted in Table~\ref{tab:holigaze8k_mllm}, we compare LookAnything with leading closed-source and open-source MLLMs (e.g., Gemini-3.1-Pro \citep{comanici2025gemini}, GPT-5.6-Sol \citep{singh2025openai}, etc.), including generalist and embodied models. We highlight three observations:

\paragraph{\textbf{\textit{Geometric grounding is the primary gap for generalist MLLMs.}}} While closed-source models like Gemini-3.7-Flash~\citep{comanici2025gemini} excel on descriptive MCQs, most of them struggle on localization. The strongest open-source localizers (Qwen3.8-27B~\citep{bai2025qwen3}) lag LookAnything by large margins (e.g., 0.152 vs 0.333 GTL-T2B mIoU). Broad language and reasoning capability therefore does not intrinsically yield precise gaze-conditioned grounding.

\paragraph{\textbf{\textit{Embodied pretraining does not automatically transfer.}}} Embodied models (e.g., RoboBrain~\citep{team2025robobrain, tan2026robobrain}) perform adequately on GCD/GDO but remain weak on point and box localization, indicating that generic spatial or action-centric supervision cannot substitute for gaze-specific priors.

\begin{figure}[!t]
\centering
\includegraphics[width=0.92\textwidth]{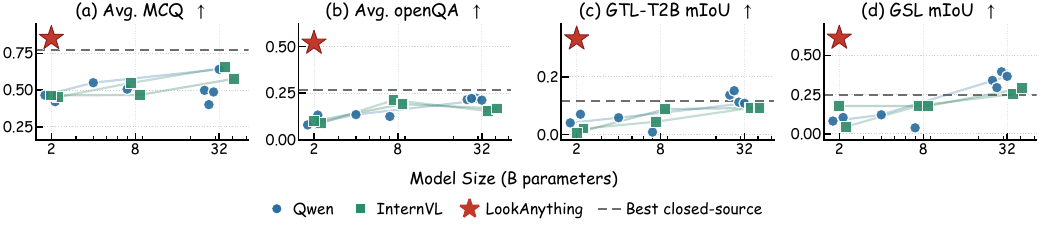}
\caption{\textbf{HoliGaze-Bench performance vs. model size.} Our 2B model (red star) achieves superior performance despite its compact scale.}
\label{fig:holigaze_vs_size}
\end{figure}

\paragraph{\textbf{\textit{LookAnything delivers broad gains at a compact scale.}}} Our 2B model outperforms all baselines across all metrics. Compared to its Qwen3-VL-2B backbone, it reduces GTL-T2P $L_2$ from 0.387 to 0.094, validating the effectiveness of large-scale multi-task training and gaze-aware representation learning on HoliGaze-Corpus. Figure~\ref{fig:holigaze_vs_size} further shows that these gains are not the result of model scale: LookAnything lies well above the open-source scaling trend, and even 27B--38B generalists fall far short of its localization averages.

\subsection{Results on Native Gaze Target Detection}
\label{subsec:results_native}
To verify that task unification does not compromise conventional gaze target detection, we evaluate LookAnything on the official test splits of GazeFollow \citep{recasens2015they}, VAT \citep{chong2020detecting}, ChildPlay \citep{tafasca2023childplay}, GazeHOI \citep{tafasca2024toward}, and GOO-Real \citep{tomas2021goo}, against representative specialist models. Following native task settings, all models take head boxes or crops as input. LookAnything is evaluated across all benchmarks using a single set of weights without domain-specific fine-tuning. Other models are first pre-trained and evaluated on GazeFollow, and then individually fine-tuned for other domains (e.g., fine-tuned on the VAT training set for VAT evaluation). For GOO-Real, baselines are fine-tuned on GOO-Syn. \textit{More details about baselines and evaluation protocols are provided in Appendix~B}.

As shown in Table~\ref{tab:external_native_bench}, LookAnything achieves the best overall performance with an average $L_2$ of 0.109 and AUC of 0.942 across all 5 benchmarks. Specifically, it achieves the best $L_2$ and AUC on 4 benchmarks (VAT, ChildPlay, GazeHOI, and GOO-Real). On GazeFollow, it obtains the best average $L_2$ despite being trained jointly across diverse tasks. Overall, LookAnything matches or exceeds dedicated specialists on core localization.

\begin{table*}[t]
\centering
\caption{\textbf{Results on 5 gaze-following benchmarks.} Best and second-best are \textbf{bold} and \underline{underlined}.}
\label{tab:external_native_bench}
\vspace{-1mm}
\setlength{\tabcolsep}{3pt}
\renewcommand{\arraystretch}{1.02}
\resizebox{\textwidth}{!}{%
\begin{tabular}{l cc cc ccc cc ccc cc}
\toprule
& \multicolumn{2}{c}{\textbf{Avg.}} & \multicolumn{2}{c}{VAT} & \multicolumn{3}{c}{GazeFollow} & \multicolumn{2}{c}{ChildPlay} & \multicolumn{3}{c}{GazeHOI} & \multicolumn{2}{c}{GOO-Real} \\
\cmidrule(lr){2-3} \cmidrule(lr){4-5} \cmidrule(lr){6-8} \cmidrule(lr){9-10} \cmidrule(lr){11-13} \cmidrule(lr){14-15}
Method & \textbf{L2}$\downarrow$ & \textbf{AUC}$\uparrow$ & L2$\downarrow$ & AUC$\uparrow$ & L2$\downarrow$ & Min L2$\downarrow$ & AUC$\uparrow$ & L2$\downarrow$ & AUC$\uparrow$ & L2$\downarrow$ & AUC$\uparrow$ & GazeAcc$\uparrow$ & L2$\downarrow$ & AUC$\uparrow$ \\
\midrule
\rowcolor{cvprblue!6} Recasens et al.~\tabcite{recasens2015they} & 0.253 & 0.809 & 0.242 & 0.789 & 0.227 & 0.155 & 0.857 & 0.292 & 0.754 & 0.259 & 0.817 & 0.371 & 0.245 & 0.829 \\
Lian et al.~\tabcite{lian2018believe} & 0.184 & 0.884 & 0.157 & 0.883 & 0.189 & 0.119 & 0.885 & 0.167 & 0.908 & 0.200 & 0.877 & 0.561 & 0.205 & 0.868 \\
\rowcolor{cvprblue!6} Chong et al.~\tabcite{chong2020detecting} & 0.172 & 0.857 & 0.126 & 0.864 & 0.138 & 0.077 & 0.923 & 0.128 & 0.858 & 0.197 & 0.869 & 0.564 & 0.272 & 0.770 \\
HGTTR~\tabcite{tu2022end} & 0.224 & 0.825 & 0.266 & 0.816 & 0.169 & 0.109 & 0.847 & 0.204 & 0.859 & 0.216 & 0.824 & 0.528 & 0.265 & 0.780 \\
\rowcolor{cvprblue!6} MADGTD~\tabcite{tonini2022multimodal} & 0.172 & 0.857 & 0.128 & 0.860 & 0.138 & 0.077 & 0.925 & 0.133 & 0.866 & 0.195 & 0.870 & 0.578 & 0.265 & 0.762 \\
Gupta et al.~\tabcite{gupta2022modular} & 0.181 & 0.867 & 0.128 & 0.898 & 0.156 & 0.091 & 0.914 & 0.144 & 0.893 & 0.198 & 0.867 & 0.556 & 0.279 & 0.765 \\
\rowcolor{cvprblue!6} Miao et al.~\tabcite{miao2023patch} & 0.170 & 0.877 & 0.117 & 0.906 & 0.132 & 0.072 & 0.928 & 0.122 & 0.919 & 0.194 & 0.864 & 0.584 & 0.287 & 0.768 \\
Tafasca et al.~\tabcite{tafasca2023childplay} & -- & -- & 0.123 & 0.904 & 0.128 & 0.065 & 0.935 & 0.112 & 0.924 & -- & -- & -- & -- & -- \\
\rowcolor{cvprblue!6} ViTGaze~\tabcite{song2024vitgaze} & 0.136 & 0.900 & 0.103 & 0.908 & 0.105 & 0.047 & 0.949 & 0.112 & 0.899 & 0.163 & 0.880 & \underline{0.714} & 0.196 & 0.862 \\
Sharingan~\tabcite{tafasca2024sharingan} & 0.150 & 0.900 & 0.104 & 0.920 & 0.112 & 0.055 & 0.944 & 0.110 & 0.930 & 0.177 & 0.886 & 0.633 & 0.248 & 0.819 \\
\rowcolor{cvprblue!6} Gaze-LLE (ViT-B)~\tabcite{ryan2025gaze} & 0.132 & 0.930 & 0.105 & 0.932 & 0.104 & \underline{0.045} & \underline{0.956} & 0.115 & 0.945 & 0.162 & 0.906 & 0.705 & 0.175 & 0.909 \\
Gaze-LLE (ViT-L)~\tabcite{ryan2025gaze} & \underline{0.127} & \underline{0.933} & \underline{0.100} & \underline{0.937} & \underline{0.098} & \textbf{0.041} & \textbf{0.958} & \underline{0.102} & \underline{0.950} & \underline{0.160} & \underline{0.907} & 0.710 & 0.173 & 0.915 \\
\rowcolor{cvprblue!6} GazeAnywhere~\tabcite{cao2026gaze} & 0.183 & 0.897 & 0.236 & 0.845 & 0.179 & 0.115 & 0.923 & 0.139 & 0.916 & 0.196 & 0.885 & 0.625 & \underline{0.165} & \underline{0.918} \\
\midrule
\rowcolor{cvprblue!15} \textbf{LookAnything (Ours)} & \textbf{0.109} & \textbf{0.942} & \textbf{0.098} & \textbf{0.941} & \textbf{0.091} & 0.048 & 0.945 & \textbf{0.094} & \textbf{0.956} & \textbf{0.130} & \textbf{0.940} & \textbf{0.771} & \textbf{0.132} & \textbf{0.927} \\
\bottomrule
\end{tabular}%
}
\vspace{-2mm}
\end{table*}

\subsection{Ablation Study}
\label{subsec:ablation}

In this section, we conduct extensive ablation studies to investigate the following questions.

\begin{figure}[!htbp]
  \centering
  \begin{minipage}[t]{0.485\textwidth}
    \centering
    \includegraphics[width=\linewidth]{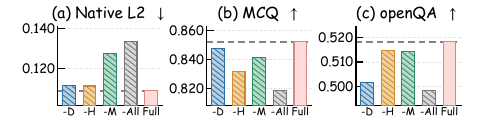}
    \caption{\textbf{Ablation of gaze-prior queries.} ``-D'', ``-H'', and ``-M'' denote removing the depth, head mask, and target heatmap queries, respectively. ``-All'' removes all three groups.}
    \label{fig:ablation_queries}
  \end{minipage}\hfill
  \begin{minipage}[t]{0.485\textwidth}
    \centering
    \includegraphics[width=\linewidth]{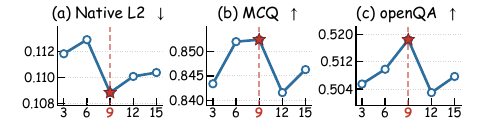}
    \caption{\textbf{Ablation of query-token budget.} $k \in \{3, 6, 9, 12, 15\}$; the red star and vertical line mark the adopted setting $k=9$.}
    \label{fig:ablation_tokens}
  \end{minipage}
\end{figure}

\paragraph{Q1: What is the contribution of gaze-prior queries?}
We ablate the query-based supervisions individually and jointly. As shown in Figure~\ref{fig:ablation_queries}, discarding all queries (i.e., training the bare Qwen3-VL-2B) causes the most severe degradation, validating our overall design. Individually, omitting target heatmaps severely degrades localization by hindering spatial representation learning. Discarding head masks primarily hampers MCQ reasoning, underscoring the need for reliable source head grounding. Removing the depth prior impairs open-ended generation, as 3D context is vital for resolving spatial ambiguities. Thus, these geometric priors offer complementary benefits.

\paragraph{Q2: How does the number of query tokens affect performance?}
We sweep the token budget $k \in \{3, 6, 9, 12, 15\}$ per prior group (Figure~\ref{fig:ablation_tokens}). A restricted budget ($k \le 6$) degrades localization and open-ended generation, while $k \ge 12$ causes slight regressions likely due to attention dilution. The model achieves the optimal trade-off at $k = 9$, providing sufficient representational bandwidth without overwhelming the language backbone.

\begin{wraptable}[11]{r}{0.48\textwidth}
\vspace{-10pt}
\centering
\caption{\textbf{Component ablations.} Each ablated row changes only the named component. The full model uses structured attention, separated query groups, and vision-encoder LoRA.}
\label{tab:component_ablation}
\setlength{\tabcolsep}{4pt}
\renewcommand{\arraystretch}{1.05}
\footnotesize
\resizebox{\linewidth}{!}{%
\begin{tabular}{l|ccc|c}
\toprule
\multirow{2}{*}{\textbf{Setting}} & \multicolumn{3}{c|}{\textbf{HoliGaze-Bench}} & \textbf{Native} \\
 & \begin{tabular}{@{}c@{}}Avg. \\ MCQ $\uparrow$\end{tabular} & \begin{tabular}{@{}c@{}}Avg. \\ openQA $\uparrow$\end{tabular} & \begin{tabular}{@{}c@{}}GTL-T2P \\ $L_2$ $\downarrow$\end{tabular} & \begin{tabular}{@{}c@{}}Avg. \\ $L_2$ $\downarrow$\end{tabular} \\
\midrule
Causal attention mask & 0.848 & 0.498 & 0.097 & 0.111 \\
Shared query group & 0.845 & 0.493 & 0.098 & 0.112 \\
Frozen vision encoder & 0.774 & 0.467 & 0.130 & 0.138 \\
\rowcolor{cvprblue!15}
\textbf{Full model} & \textbf{0.852} & \textbf{0.518} & \textbf{0.094} & \textbf{0.109} \\
\bottomrule
\end{tabular}%
}
\vspace{-4pt}
\end{wraptable}
\paragraph{Q3: Is block-wise structured attention necessary?}
Replacing our structured attention (Figure~\ref{fig:framework}) with a standard causal mask allows later query groups to attend to earlier ones. This shortcut dependency introduces gradient noise and degrades semantic generalization (Table~\ref{tab:component_ablation}). Instead, our structured design ensures each domain independently extracts relevant features from the shared context, preserving representational purity and improving overall accuracy.

\paragraph{Q4: Are specialized queries necessary for different geometric priors?}
Using a unified query group to simultaneously predict depth, head masks, and heatmaps consistently underperforms (Table~\ref{tab:component_ablation}). Forcing a single token set to capture disparate geometric properties causes feature entanglement and feeds conflicting signals to the language head. Disentangling queries into specialized groups maintains clean, domain-specific representations.

\paragraph{Q5: Is fine-tuning the vision encoder necessary?}
While MLLMs possess strong visual representations, fine-grained gaze understanding requires capturing subtle cues like eye orientation. As shown in Table~\ref{tab:component_ablation}, freezing the vision encoder severely degrades all metrics (e.g., $L_2$ from 0.109 to 0.138). This confirms that task-specific adaptation of the visual backbone is indispensable for extracting the nuanced geometric features necessary for accurate gaze target detection.

\subsection{Out-of-domain Evaluation}
\label{subsec:ood_evaluation}

To evaluate the out-of-domain (OOD) generalization of LookAnything, we conduct both quantitative evaluation on unseen benchmarks and qualitative deployment in real-world physical environments.

\begin{wraptable}[9]{r}{0.46\textwidth}
\vspace{-6pt}
\centering
\caption{\textbf{OOD Evaluation.} Avg $L_2$ ($\downarrow$) on GFIE and 4D-OR.}
\label{tab:ood_bench}
\setlength{\tabcolsep}{6pt}
\renewcommand{\arraystretch}{1.05}
\footnotesize
\resizebox{\linewidth}{!}{%
\begin{tabular}{lcc}
\toprule
\textbf{Method} & \textbf{GFIE}~\citep{hu2023gfie} & \textbf{4D-OR}~\citep{chen2026they} \\
\midrule
Sharingan~\citep{tafasca2024sharingan} & 0.0537 & 0.095 \\
ViTGaze~\citep{song2024vitgaze} & 0.0534 & 0.093 \\
Gaze-LLE (ViT-L/14)~\citep{ryan2025gaze} & 0.0518 & 0.084 \\
\midrule
\rowcolor{cvprblue!15}
\textbf{LookAnything (Ours)} & \textbf{0.0464} & \textbf{0.076} \\
\bottomrule
\end{tabular}%
}
\vspace{-4pt}
\end{wraptable}
\paragraph{Quantitative Results on OOD Benchmarks.}
We evaluate LookAnything on the GFIE~\citep{hu2023gfie} and 4D-OR~\citep{chen2026they} test sets to assess its zero-shot generalization capability. Specifically, GFIE features complex indoor scenarios, while 4D-OR captures surgical activities in operating rooms; the data domains of both datasets are strictly unseen during our training phase. As shown in Table~\ref{tab:ood_bench}, LookAnything consistently achieves the lowest average $L_2$ distance, significantly outperforming existing specialist models such as Gaze-LLE~\citep{ryan2025gaze}, ViTGaze~\citep{song2024vitgaze}, and Sharingan~\citep{tafasca2024sharingan}. This demonstrates that our model, trained on the diverse HoliGaze-Corpus, acquires robust gaze-following priors that generalize effectively to novel domains without any dataset-specific fine-tuning.

\begin{wrapfigure}[19]{l}{0.48\textwidth}
  \centering
  \includegraphics[width=\linewidth]{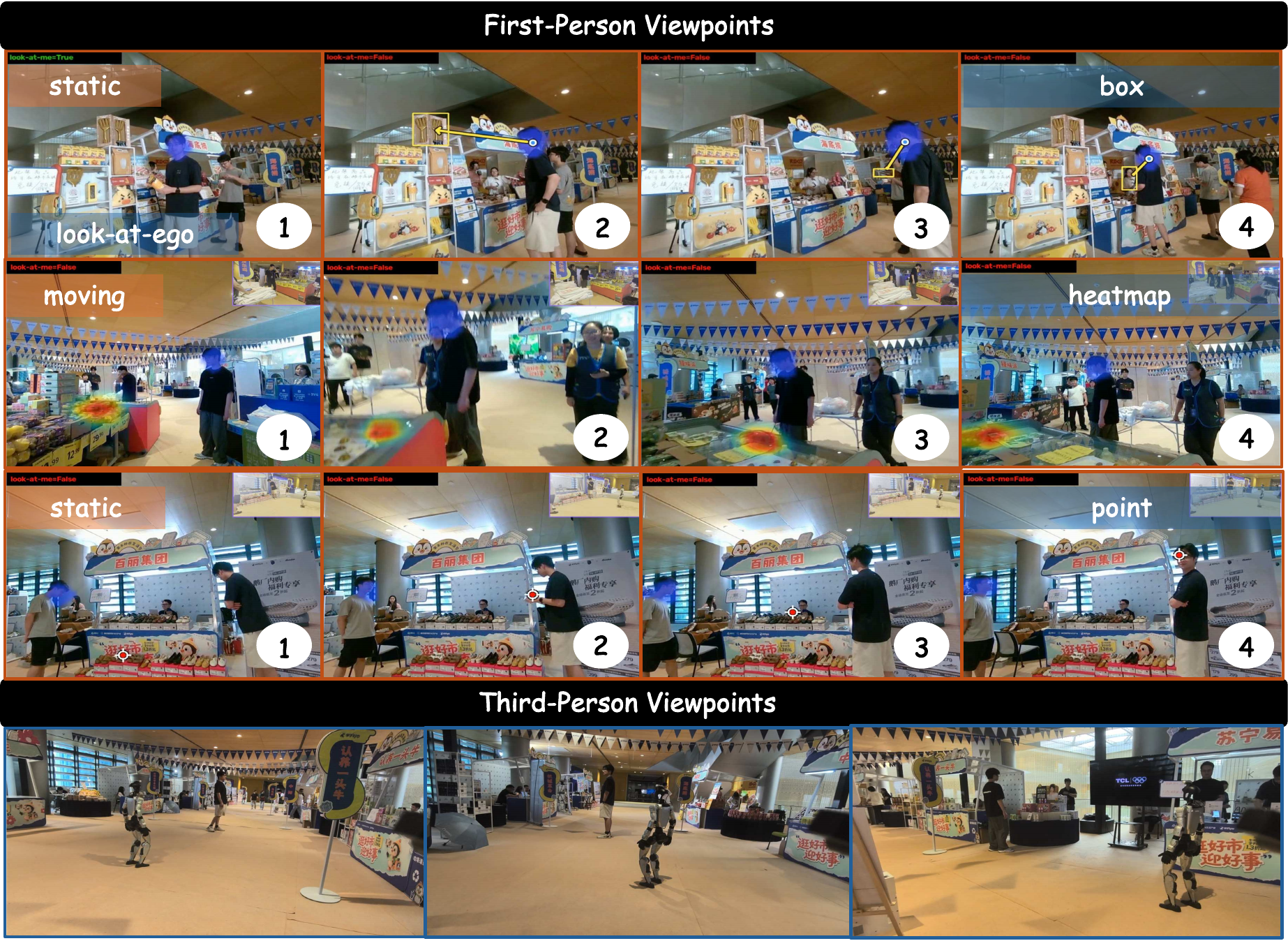}
  \caption{Real-world Experiments. \textit{The demo video in the supplementary material provides more visualizations.}}
  \label{fig:robotic_deploy}
\end{wrapfigure}
\paragraph{Real-world Experiments.}
We deploy LookAnything on a Unitree G1 robot in real-world physical environments. Figure~\ref{fig:robotic_deploy} illustrates the experimental setup and qualitative visualization results. From the robot's egocentric perspective, the 3 rows depict 3 different subjects in various shopping scenarios, with the model predicting gaze targets in the form of bounding boxes, heatmaps, and points, respectively. The blue overlays indicate the predicted head masks of the source persons, while the top-left corner displays the gaze contact detection results (i.e., whether the source person is looking at the ego-camera). As observed, LookAnything accurately grounds the source person in multi-person scenes, reliably recognizes social gaze signals, and precisely tracks gaze targets under both static and moving perspectives. These results demonstrate its strong adaptability and robustness for real-world embodied applications. \textit{The real-world demo video in the supplementary material provides further visualizations}, showcasing LookAnything's performance across typical scenarios such as human-robot interaction, shopping assistance, and desktop, covering various output formats under diverse perspectives.

\section{Conclusion}
In this paper, we introduced LookAnything, a foundation model that unifies diverse gaze-centric tasks for holistic gaze following. By internalizing gaze priors via large-scale multi-task and cross-domain learning, our framework bypasses explicit auxiliary inputs, enabling flexible adaptation across various scenarios, task settings, and output formats. To facilitate this paradigm, we construct HoliGaze, a comprehensive corpus and benchmark standardizing 311K Q\&A samples across 9 distinct tasks. Empirical results confirm that LookAnything surpasses leading VLM in holistic gaze following while remaining highly competitive with specialists in conventional gaze target detection. Quantitative evaluation on unseen datasets and real-world deployment on a Unitree G1 robot further underscore its robust out-of-domain generalization in complex and diverse physical scenarios. 
We believe LookAnything and HoliGaze will inspire future advances in human-centric interactive applications and embodied AI, envisioning them as a foundational step toward socially intelligent embodied agents capable of seamless non-verbal cognitive alignment with humans.

\clearpage
\appendix
\begin{center}
\centering
{\Large\sffamily\bfseries LookAnything: A Foundation Model for Holistic Gaze Following}\\[0.5em]
{\large\sffamily Supplementary Material}
\end{center}
\addcontentsline{toc}{part}{Supplementary Material}
\bigskip
\setcounter{topnumber}{1}
\setcounter{bottomnumber}{1}
\setcounter{totalnumber}{2}
\renewcommand{\floatpagefraction}{0.58}

\section{Related Work}
\label{sec:appendix_related}
\textbf{Gaze Following.} Gaze following aims to infer where people look in visual scenes, providing vital cues for attention estimation \citep{zhou2025towards, kasahara2022look, zhou2024learning, zhou2025behavior, tang2026letp}, intent understanding \citep{fan2018inferring, peng2026eye}, and human–robot interaction \citep{zuo2026gaze2act, saran2018human, yuan2026embodiedr1}. Pioneered by \citet{recasens2015they}, classical frameworks formulate this task as predicting a 2D spatial heatmap by fusing cropped head representations with full-scene context in dual-stream architectures \citep{lian2018believe, chong2020detecting, fang2021dual}. Subsequent studies advance this line of research by incorporating scene depth priors \citep{gupta2022modular, tonini2022multimodal}, patch-level distributions \citep{miao2023patch}, and transformer-based head–target associations \citep{tu2022end, tafasca2024sharingan, lin2025gazehta, song2024vitgaze, mi2026multi}. More recently, methods such as Gaze-LLE \citep{ryan2025gaze} and GazeMoE \citep{dai2026gazemoe} scale up visual representations using large pretrained vision backbones or mixture-of-experts designs. Despite steady progress, these specialists rely on highly customized pipelines with pre-extracted crops or depth maps that induce cascading errors. Moreover, their coarse heatmap outputs lack instance-level semantics \citep{tonini2023object, tafasca2024toward}, hindering seamless integration with downstream interactive and embodied tasks \citep{saran2018human, zuo2026gaze2act, pani2026gaze, lai2026fam}. In contrast, LookAnything reformulates gaze localization and holistic understanding within a generalist vision-language framework, supporting both instance-level grounding and semantic description.

\paragraph{\textbf{Vision-Language Models (VLMs).}} VLMs have demonstrated remarkable capabilities in open-vocabulary recognition, visual reasoning, and conversational interaction \citep{bai2025qwen3, comanici2025gemini, tang2026spatialsv, zhou2026logic}. Motivated by these advances, recent studies adapt VLMs to human gaze tasks \citep{wang2026enhancing, mathew2026gazevlm}. Data-driven fine-tuning approaches like GazeVLM \citep{mathew2026gazevlm} and VL4Gaze \citep{wang2025vl4gaze} formulate gaze following as standard visual question answering without incorporating architectural gaze prior contexts.
In contrast, LookAnything injects gaze priors (e.g., head masks and scene depth) into latent representations via learnable context queries and, scaled with diverse gaze corpora, establishes a foundation model for holistic gaze following.

\paragraph{\textbf{Benchmarks for Gaze Following and Understanding.}} Foundational gaze following benchmarks, such as GazeFollow \citep{recasens2015they}, VideoAttentionTarget \citep{chong2020detecting}, ChildPlay \citep{tafasca2023childplay}, and GOO \citep{tomas2021goo}, predominantly focus on gaze target detection in both static images and dynamic videos across diverse domains, including unconstrained everyday scenes, child-centric settings, and specialized retail environments. 
Subsequently, research has expanded to capture more complex gaze dynamics, such as shared attention, where VideoCoAtt \citep{fan2018inferring} enables the study of multi-person co-gaze behaviors. 
More recently, to advance beyond pure gaze target localization, Gaze-Co \citep{cao2026gaze} incorporates semantic object descriptions, and VL4Gaze \citep{wang2026enhancing} introduces vision-language question answering for gaze understanding. Despite these efforts, current benchmarks remain fragmented, each isolating specific gaze facets within narrow task definitions. To bridge this gap, we propose HoliGaze, a unified dataset and benchmark encompassing 5 capability dimensions across 9 distinct gaze-centric tasks, providing the most comprehensive evaluation suite for holistic gaze following to date.

\section{Additional Implementation Details}
\label{sec:appendix_implementation}

\subsection{Evaluation Protocols for HoliGaze-Bench}
\label{subsec:appendix_model_pool}

\subsubsection{Overview of Evaluated Models}
To comprehensively benchmark holistic gaze following capabilities, we curate a diverse evaluation pool comprising 33 distinct models. These models are systematically categorized into three primary families: proprietary multimodal APIs, general-purpose open-weight models, and specialized embodied foundation models. All evaluations are conducted in a zero-shot setting, utilizing either the official provider APIs or the canonical checkpoints hosted on Hugging Face.

\paragraph{Proprietary Multimodal APIs.}
This category encompasses state-of-the-art closed-source models accessed via commercial endpoints, which typically leverage massive scale and proprietary training recipes to set the upper bound for general multimodal reasoning. Within the Google Gemini lineage \citep{comanici2025gemini}, we evaluate multiple generations: Gemini-3.7-Flash serves as the highly efficient, reasoning-optimized endpoint supporting a massive 1M-token context window and interleaved multimodal inputs, while Gemini-3.1-Pro represents the higher-compute tier designed for complex, multi-step spatial and logical reasoning. We additionally include the previous-generation Gemini-2.5-Flash and Gemini-2.5-Pro to track cross-generational capability improvements. From OpenAI, we evaluate GPT-5.6-Sol \citep{singh2025openai}, the flagship vision-language model exhibiting exceptional zero-shot spatial grounding and instruction-following capabilities. To provide a comprehensive view of the commercial landscape, we also incorporate DeepSeek-V4-Flash \citep{xu2026deepseek}, a highly optimized Mixture-of-Experts (MoE) model; GLM-5.3-Flash \citep{zeng2026glm}; Grok-4.6 \citep{xiaimodelcard_card-4p6-4cd2dc57}; and Qwen3-Max \citep{bai2025qwen3}. These models provide robust alternatives with strong visual grounding, spatial relationship understanding, and state-of-the-art performance on broad vision-language benchmarks.

\paragraph{General-Purpose Open-Weight Models.}
This group consists of publicly accessible vision-language models that allow for local inference and transparent architectural analysis, varying significantly in parameter scale and architectural design. We comprehensively evaluate the Qwen ecosystem \citep{bai2025qwen3}, including the latest Qwen3-VL series (2B, 4B, 32B), which features dynamic image resolution and temporal-aware positional encoding, making them highly adept at fine-grained spatial localization. To analyze the scaling laws of gaze understanding within a unified architectural family, we also assess various intermediate checkpoints: the dense Qwen3.5-2B and Qwen3.5-27B, the MoE-based Qwen3.6-27B, and the advanced Qwen3.8-27B. The earlier Qwen2.5-VL-7B is included as a baseline for instruction-tuned dynamic-resolution VLMs. Furthermore, we evaluate the InternVL family \citep{zhu2025internvl3, wang2025internvl3}, specifically InternVL3 (2B, 8B, 38B) and the subsequently aligned InternVL3.5 (2B, 8B, 38B). Known for their high-resolution patch processing and strong visual encoders, these models excel in capturing intricate visual details, which is theoretically advantageous for detecting subtle gaze cues and small target objects.

\paragraph{Embodied Foundation Models.}
Given the strong connection between gaze understanding and human-robot interaction, we evaluate a suite of VLMs specifically fine-tuned on embodied, ego-centric, and robotic manipulation datasets. This allows us to investigate whether action-centric spatial supervision naturally transfers to human gaze target detection. We evaluate vision-language-action (VLA) models such as Embodied-R1-3B \citep{yuan2026embodiedr1} and Embodied-R1.5-8B \citep{yuan2026embodied}, which are optimized for embodied control and spatial grounding. We also include the RoboBrain series (RoboBrain-2.0-7B \citep{team2025robobrain} and RoboBrain-2.5-8B \citep{tan2026robobrain}), tailored for 3D spatial reasoning and complex robotic manipulation tasks. To assess lightweight, edge-deployable models focusing on efficient visual-motor policies, we incorporate RynnBrain-2B and RynnBrain-9B \citep{dang2026rynnbrain}. Finally, specialized robotic VLMs including MiMo-Embodied-7B \citep{hao2025mimo}, Pelican-VL-7B \citep{zhang2025pelican}, VeBrain-7B \citep{luo2025visual}, and Hy-Embodied-30B \citep{wang2026hy} are evaluated to provide a comprehensive view of how current embodied foundation models, trained on diverse large-scale robotic datasets, handle human-centric cognitive signals like gaze.

\subsubsection{Evaluation Settings and Coordinate Protocols}
All evaluated models define the image coordinate system with the origin at the top-left corner, the x-axis pointing right (width), and the y-axis pointing down (height). However, they differ significantly in coordinate ordering ($x$-$y$ vs. $y$-$x$), format, and scaling (absolute pixels vs. relative coordinates). Specifically, we adapt the evaluation to the following protocols for the 33 evaluated VLMs:
\begin{itemize}
    \item \textbf{The Qwen2.5 Ecosystem and Derivatives:} The Qwen2.5-VL-7B model directly uses absolute pixels corresponding to the actual dimensions of the input image, adopting an $[x_{\text{min}}, y_{\text{min}}, x_{\text{max}}, y_{\text{max}}]$ bounding box format and $(x, y)$ point format. Consequently, specialized embodied foundation models built upon the Qwen2.5 architecture—such as Embodied-R1-3B, RoboBrain-2.0-7B, MiMo-Embodied-7B, Pelican-VL-7B, and VeBrain-7B—inherit this absolute pixel scaling with the $x$-$y$ ordering.
    \item \textbf{The Qwen3 Ecosystem and Derivatives:} Starting from the Qwen3 series (including Qwen3-VL-2B/4B/32B), the protocol shifts to relative coordinates normalized to the range of $[0, 1000]$, while retaining the $[x_{\text{min}}, y_{\text{min}}, x_{\text{max}}, y_{\text{max}}]$ box and $(x, y)$ point formats. Embodied models derived from this newer architecture—such as Embodied-R1.5-8B, RoboBrain-2.5-8B, RynnBrain-2B/9B, and Hy-Embodied-30B—explicitly adopt this normalized $[0, 1000]$ scale with the $x$-$y$ ordering.
    \item \textbf{The InternVL Family:} Models in the InternVL lineage (InternVL3 and InternVL3.5 across 2B, 8B, and 38B scales) consistently employ the $[x_{\text{min}}, y_{\text{min}}, x_{\text{max}}, y_{\text{max}}]$ box format normalized to $[0, 1000]$, typically enclosed within specific spatial tokens (e.g., \texttt{<box>}).
    \item \textbf{Proprietary APIs and Other Generalist Models:} The Google Gemini lineage (Gemini-2.5-Flash/Pro, Gemini-3.1-Pro, Gemini-3.7-Flash) natively employs a $y$-$x$ ordering, representing bounding boxes as $[y_{\text{min}}, x_{\text{min}}, y_{\text{max}}, x_{\text{max}}]$ and points as $[y, x]$, both normalized to $[0, 1000]$. GLM-5.3-Flash follows the $x$-$y$ convention ($[x_{\text{min}}, y_{\text{min}}, x_{\text{max}}, y_{\text{max}}]$ and $(x, y)$) scaled to $[0, 1000]$, while DeepSeek-V4-Flash utilizes the $x$-$y$ convention scaled to $[0, 999]$. For OpenAI's GPT-5.6-Sol and Grok-4.6, which lack a rigid native spatial protocol in their official documentation, we explicitly prompt them to output the standard $x$-$y$ relative coordinates normalized to $[0, 1000]$.
\end{itemize}

To ensure rigorous and fair zero-shot evaluations, we dynamically adapt the question prompts in HoliGaze-Bench to strictly align with the native coordinate protocol of each specific model. The spatial instructions within the prompts are customized to elicit the exact coordinate format and scaling the model was trained on. During inference, the models generate spatial predictions following their inherent formatting rules. Before computing the final evaluation metrics, we parse and convert these heterogeneous outputs back into a unified, standardized format: relative coordinates normalized to $[0, 1000]$ with the $x$-$y$ ordering (i.e., $[x_{\text{min}}, y_{\text{min}}, x_{\text{max}}, y_{\text{max}}]$ and $(x, y)$). Specifically, for models outputting absolute pixels, the coordinates are divided by the actual image width and height, and then scaled by 1000; for models using the $[0, 999]$ scale, coordinates are divided by 999 and scaled by 1000; and for Gemini models, the $y$ and $x$ axes are swapped. This meticulous alignment prevents artificial performance degradation caused by format mismatch and accurately reflects the true spatial grounding capabilities of each model.

\subsection{Evaluation Protocols for Gaze Target Detection}
\label{subsec:appendix_baselines}

\subsubsection{Overview of Baseline Models}
We comprehensively evaluate our proposed LookAnything against 11 representative specialist baselines, which cover the principal architectural families and methodological advancements in gaze target detection:
\begin{itemize}
    \item \textbf{Foundational Dual-Stream CNNs:} We include the pioneering work by \citet{recasens2015they}, which first formulated gaze following using a dual-pathway architecture to fuse head crops and full-scene images. We also evaluate \citet{lian2018believe}, which explicitly models gaze direction and depth cues, and \citet{chong2020detecting}, which extends the dual-stream paradigm to the spatio-temporal domain to capture dynamic gaze behaviors in videos.
    \item \textbf{Transformer-Based Specialists:} To represent attention-driven architectures, we evaluate HGTTR \citep{tu2022end}, which formulates gaze target detection as an end-to-end set prediction problem, bypassing the need for pre-extracted head crops. We also include ViTGaze \citep{song2024vitgaze}, which leverages Vision Transformers to effectively model global context and head-target interactions, and Sharingan \citep{tafasca2024sharingan}, a transformer architecture specifically tailored for multi-person gaze following and shared attention.
    \item \textbf{Depth- and Multimodal-Enhanced Approaches:} To assess methods utilizing explicit physical priors, we compare against MADGTD \citep{tonini2022multimodal}, which incorporates monocular depth maps to address cross-domain gaze target detection, and \citet{gupta2022modular}, which proposes a modular multimodal architecture fusing scene, depth, and pose features. Furthermore, we include PDP \citep{miao2023patch}, which predicts gaze distributions at the patch level rather than the pixel level to enhance spatial reasoning.
    \item \textbf{Recent Large-Scale and Concept-Driven Models:} We evaluate the most recent state-of-the-art specialists, including Gaze-LLE \citep{ryan2025gaze}, which adapts large-scale pretrained visual encoders (e.g., DINOv2) to achieve robust gaze representations. Finally, we include GazeAnywhere \citep{cao2026gaze}, which introduces a promptable gaze estimation framework that allows identifying the target subject via flexible concept prompts (e.g., appearance or location) in addition to traditional spatial coordinates.
\end{itemize}

\subsubsection{Training and Evaluation Protocols}
To ensure fair and consistent comparisons, all models are provided with explicit head location cues (e.g., bounding boxes or cropped head images) to identify the gaze source during evaluation. Notably, LookAnything utilizes the head bounding box as the spatial condition to ground the source person, while GazeAnywhere employs the center point of the head bounding box, strictly aligning with its native settings.

Regarding the training paradigms, LookAnything undergoes multi-task, cross-domain joint training on the HoliGaze-Corpus (data composition is detailed in Table~\ref{tab:app_source}). In contrast, the specialist baselines follow a sequential domain adaptation strategy. They are initially trained and evaluated on the native GazeFollow \citep{recasens2015they} dataset. These pretrained weights are then used to initialize fine-tuning on the respective training splits of VAT \citep{chong2020detecting}, ChildPlay \citep{tafasca2023childplay}, GazeHOI \citep{tafasca2024toward}, and GOO-Syn \citep{tomas2021goo}, prior to evaluation on their corresponding test sets. For the GOO \citep{tomas2021goo} dataset, models fine-tuned on GOO-Syn are evaluated on the GOO-Real test set, ensuring that the real-world GOO-Real images remain strictly unseen to all models during training. Furthermore, consistent with its evaluation protocol, the training pipeline for GazeAnywhere also uniformly adopts the center point of the head bounding box as the standard spatial reference.

To guarantee reproducibility, we directly evaluate the official checkpoints released by the respective authors whenever available for a given dataset. In cases where official weights or specific training recipes for a target domain are absent, we utilize the corresponding official codebases to train or fine-tune the models on the designated data sources following their canonical configurations.

\subsection{LookAnything Model Architecture}
\label{subsec:appendix_architecture}

LookAnything is built upon the generalist multimodal large language model, Qwen3-VL \citep{bai2025qwen3}. The architecture integrates a vision encoder, a language model backbone, learnable gaze-prior queries, and specialized geometric decoders. 
Specifically, the vision encoder (ViT) processes the input image into a sequence of visual patch tokens, which are then concatenated with textual tokens from the prompt. To inject gaze-aware spatial inductive biases, we introduce three groups of learnable continuous queries: depth queries $Q^d$, head queries $Q^h$, and gaze target queries $Q^g$. Each group consists of $k=9$ tokens. These queries are appended to the multimodal embeddings and processed by the language backbone. To ensure strict functional specialization, we implement a block-wise structured attention mask as an additive 4D causal mask in PyTorch Scaled Dot-Product Attention (SDPA). This mask allows each query group to independently attend to the shared multimodal context (image and text tokens) and itself, but strictly masks out attention to other query groups, preventing shortcut dependencies such as dense depth gradients interfering with gaze target representations.

The contextualized query states are then projected into explicit geometric predictions by lightweight auxiliary decoders. The depth decoder $\mathcal{D}_d$ and head mask decoder $\mathcal{D}_h$ are instantiated as MAE-style Transformer decoders \citep{he2022masked}. They take the 9 query tokens and 256 ($16 \times 16$) learnable mask tokens with 2D sin-cos positional embeddings, process them through 2 self-attention layers (hidden dimension 512, 8 heads), and project the output to reconstruct $64 \times 64$ resolution maps. The target spatial distribution decoder also uses this MAE-style architecture but is augmented with 2 layers of cross-attention, where the spatial mask tokens attend to the visual patch tokens from the vision encoder to output a $64 \times 64$ dense gaze heatmap. 

Simultaneously, the target coordinate regressor directly predicts continuous coordinates. The point head applies 1 layer of cross-attention to the visual tokens, mean-pools the 9 gaze queries, and processes them through a 2-layer MLP (Linear $\rightarrow$ GELU $\rightarrow$ Linear) to regress the normalized continuous point $(x, y)$. The bounding box head similarly mean-pools the queries and uses a 2-layer MLP to regress the box coordinates $[x_1, y_1, x_2, y_2]$. During inference, these auxiliary decoders can be entirely skipped to reduce computational overhead, allowing the language head to autoregressively generate the final answers based on the compact gaze-context representations. The detailed model architecture hyperparameters are summarized in Table~\ref{tab:arch_hyperparams}.

\begin{table}[!htbp]
\centering
\caption{\textbf{Architecture Hyperparameters of LookAnything.}}
\label{tab:arch_hyperparams}
\setlength{\tabcolsep}{12pt}
\renewcommand{\arraystretch}{1.08}
\small
\begin{tabular}{lc}
\toprule
\textbf{Hyperparameter} & \textbf{Value} \\
\midrule
LLM Backbone & Qwen3-VL-2B \\
Vision Encoder & ViT \\
Visual Patch Size & 16 \\
\midrule
Number of Depth Queries ($Q^d$) & 9 \\
Number of Head Queries ($Q^h$) & 9 \\
Number of Gaze Target Queries ($Q^g$) & 9 \\
\midrule
MAE Decoder Mask Tokens & $16 \times 16 = 256$ \\
MAE Decoder Hidden Dimension & 512 \\
MAE Decoder Attention Heads & 8 \\
Depth Decoder Layers ($\mathcal{D}_d$) & 2 Self-Attention \\
Head Mask Decoder Layers ($\mathcal{D}_h$) & 2 Self-Attention \\
Target Heatmap Decoder & 2 Self-Attention + 2 Cross-Attention \\
Target Point Regressor & 1 Cross-Attention + Mean-Pool + 2-layer MLP \\
Target Box Regressor & Mean-Pool + 2-layer MLP \\
Output Spatial Resolution & $64 \times 64$ \\
\bottomrule
\end{tabular}
\end{table}

\subsection{Training Objectives and Loss Formulations}
\label{subsec:appendix_training}

We train LookAnything end-to-end on the HoliGaze-Corpus, which comprises 303K training samples. During training, the language backbone, the newly introduced gaze-prior queries, and the geometric decoders are jointly optimized, while the vision encoder is adapted using Low-Rank Adaptation (LoRA) \citep{hu2021lora} to capture fine-grained gaze and geometric features. The LoRA rank is set to $r=16$ and the scaling factor is $\alpha=32$.

\paragraph{Loss Formulations.}
LookAnything is trained end-to-end with a composite objective combining the gaze context-aware geometric losses and an autoregressive language loss:
\begin{equation}
\mathcal{L}_{\text{total}} = \mathcal{L}_{\text{AR}} + \lambda_d \mathcal{L}_{\text{depth}} + \lambda_h \mathcal{L}_{\text{head}} + \lambda_t \mathcal{L}_{\text{target}},
\end{equation}
where $\mathcal{L}_{\text{AR}}$ is the token-level cross-entropy over the open-form answer $Y$, given the input image $I$ and text prompt $P$:
\begin{equation}
\mathcal{L}_{\text{AR}} = -\frac{1}{N_Y} \sum_{t=1}^{N_Y} \log p_\theta\big(y_t \mid I, P, y_{<t}\big).
\end{equation}

The auxiliary geometric supervision utilizes specific loss formulations for each prior. The depth representation is optimized using a scale-invariant logarithmic (SiLog) loss to emphasize relative structural depth while remaining robust to global scale ambiguity:
\begin{equation}
\mathcal{L}_{\text{depth}} = \sqrt{\frac{1}{|\Omega|} \sum_{i \in \Omega} e_i^2 - \lambda \Big(\frac{1}{|\Omega|} \sum_{i \in \Omega} e_i\Big)^2},
\end{equation}
where $e_i = \log \hat{D}_i - \log D^*_i$ is the logarithmic error for valid pixels $\Omega$, with $\hat{D}$ being the predicted depth map and $D^*$ being the ground-truth depth map.

The predicted head mask $\hat{M}$ is supervised against the ground-truth mask $M^*$ via a combination of Binary Cross-Entropy (BCE) and soft Dice loss to mitigate foreground-background pixel imbalance:
\begin{equation}
\mathcal{L}_{\text{head}} = \mathcal{L}_{\text{BCE}}(\hat{M}, M^*) + \mathcal{L}_{\text{Dice}}(\hat{M}, M^*).
\end{equation}

The predicted target spatial distribution $\hat{A}$ is supervised using a combination of spatial KL-divergence and Mean Squared Error (MSE) against a Gaussian ground-truth distribution $A^*$ ($\sigma=3.0$), while the predicted continuous coordinates are optimized against their ground truths via Huber loss ($\delta=0.05$) for points ($\hat{p}, p^*$) and a joint L1 and Generalized IoU (GIoU) loss for bounding boxes ($\hat{b}, b^*$):
\begin{equation}
\mathcal{L}_{\text{target}} = \mathcal{L}_{\text{KL\_MSE}}(\hat{A}, A^*) + \mathcal{L}_{\text{Huber}}(\hat{p}, p^*) + \mathcal{L}_{\text{box}}(\hat{b}, b^*).
\end{equation}
The loss weights for the auxiliary geometric supervision are empirically set to $\lambda_d=0.02$ for scene depth, $\lambda_h=0.1$ for source head grounding, and $\lambda_t=1.0$ for target spatial distribution. 

\paragraph{Optimization Details.}
We optimize the model using the AdamW optimizer \citep{loshchilov2017decoupled} with a weight decay of 0.01 and a maximum gradient norm of 1.0. The learning rate follows a cosine decay schedule with a linear warmup ratio of 0.03. The language backbone uses a peak learning rate of $1 \times 10^{-5}$, while the newly initialized geometric heads and vision LoRA use a higher learning rate of $1 \times 10^{-4}$. The training is conducted on 8 NVIDIA H20 GPUs with a per-device batch size of 2 and gradient accumulation steps of 2, resulting in an effective global batch size of 32. The model is trained for 2 epochs. The detailed training hyperparameters are provided in Table~\ref{tab:train_hyperparams}.

\begin{table}[!htbp]
\centering
\caption{\textbf{Training Hyperparameters of LookAnything.}}
\label{tab:train_hyperparams}
\setlength{\tabcolsep}{12pt}
\renewcommand{\arraystretch}{1.08}
\small
\begin{tabular}{lc}
\toprule
\textbf{Hyperparameter} & \textbf{Value} \\
\midrule
Training Data & HoliGaze-Corpus (303K) \\
Optimizer & AdamW \\
Optimizer Hyperparameters & $\beta_1=0.9, \beta_2=0.999, \epsilon=10^{-8}$ \\
Weight Decay & 0.01 \\
Gradient Clipping & 1.0 \\
Learning Rate Schedule & Cosine Decay \\
Backbone Peak Learning Rate & $1 \times 10^{-5}$ \\
Geometric Heads Peak Learning Rate & $1 \times 10^{-4}$ \\
Warmup Ratio & 0.03 \\
Global Batch Size & 32 \\
Training Epochs & 2 \\
\midrule
Vision Encoder LoRA Rank ($r$) & 16 \\
Vision Encoder LoRA Alpha ($\alpha$) & 32 \\
\midrule
Depth Loss Weight ($\lambda_d$) & 0.02 \\
Head Mask Loss Weight ($\lambda_h$) & 0.1 \\
Target Loss Weight ($\lambda_t$) & 1.0 \\
Target Heatmap Gaussian $\sigma$ & 3.0 \\
Target Point Huber $\delta$ & 0.05 \\
\midrule
Hardware & 8 $\times$ NVIDIA H20 GPUs \\
\bottomrule
\end{tabular}
\end{table}

\section{Additional Analysis of HoliGaze}
\label{sec:appendix_analysis}

\subsection{Detailed Task Taxonomy}
To provide a deeper understanding of the capability dimentions evaluated in HoliGaze, we expand upon the task taxonomy introduced in Section~2.1. Each task captures a specific facet of human gaze behavior and addresses distinct practical challenges in real-world applications.

\paragraph{GTL (Gaze Target Localization).}
This task identifies the spatial location of the attended entity for a described observer. We represent this grounding as either a 2D point (GTL-T2P) or a bounding box (GTL-T2B). While point-level localization captures the fundamental direction of attention, bounding box prediction isolates the spatial extent of the target. This object-centric grounding is essential for downstream tasks such as gaze-guided object detection and robotic manipulation.

\paragraph{GTD (Gaze Target Description).}
Going beyond spatial coordinates, GTD requires a natural-language description of the attended target. This task shifts the focus from simply locating \textit{where} someone is looking to comprehending \textit{what} they are looking at. Extracting semantic meaning from the gaze target is crucial for applications like intent inference and conversational agents that need to interpret human attention in context.

\paragraph{GSL (Gaze Source Localization).}
Reversing the traditional gaze-following paradigm, GSL localizes the heads of all observers who are actively looking at a specified target. Understanding ``who is looking at this object'' is fundamental for analyzing shared attention in retail analytics, classroom education, and social crowd dynamics.

\paragraph{GSD (Gaze Source Description).}
This task provides natural-language descriptions of the visual attributes of people looking at a specific target. Going beyond mere localization, GSD identifies the demographic or visual characteristics of observers. It is highly valuable for social scene understanding, where characterizing active observers enables higher-level behavioral insights.

\paragraph{GDO (Gaze Depth Orientation).}
Because gaze is inherently a 3D physical ray, 2D image coordinates alone cannot capture its true depth. GDO analyzes whether the target is closer to or farther from the camera than the observer. This ensures models develop 3D structural awareness, which is vital for embodied agents to safely navigate physical spaces.

\paragraph{GDR (Gaze Direction Recognition).}
This task classifies the coarse 3D direction of the observer's gaze relative to the camera. Explicit direction recognition bridges 2D visual appearance with 3D pose estimation. It serves as a robust orientation prior, which is highly effective for intent prediction in autonomous driving and human--computer interaction.

\paragraph{GCD (Gaze Contact Detection).}
GCD determines whether an observer is making direct eye contact with the camera lens. Eye contact is a primal social signal of engagement and communication readiness. Recognizing mutual gaze is therefore a core prerequisite for social robots and wearable devices to initiate human interactions naturally.

\paragraph{ARR (Ambiguous Reference Recognition).}
In real-world deployments, users may issue vague queries. ARR evaluates whether a model can recognize when a query is ill-posed due to multiple indistinguishable entities and subsequently refuse to guess. This capability prevents hallucinations, ensuring reliability in critical applications like autonomous surveillance.

\paragraph{PV (Presence Verification).}
Human gaze frequently falls outside the camera's field of view. Forcing a model to predict a location for an unseen target causes catastrophic failures. PV verifies whether the attended target actually resides within the visible frame. This endows models with the crucial ability to declare a target ``out of frame,'' thereby enhancing operational safety.

\subsection{Dataset Statistics and Distributions}
In this subsection, we present additional statistics of HoliGaze.

\begin{figure}[!ht]
\centering
    \includegraphics[width=0.92\textwidth]{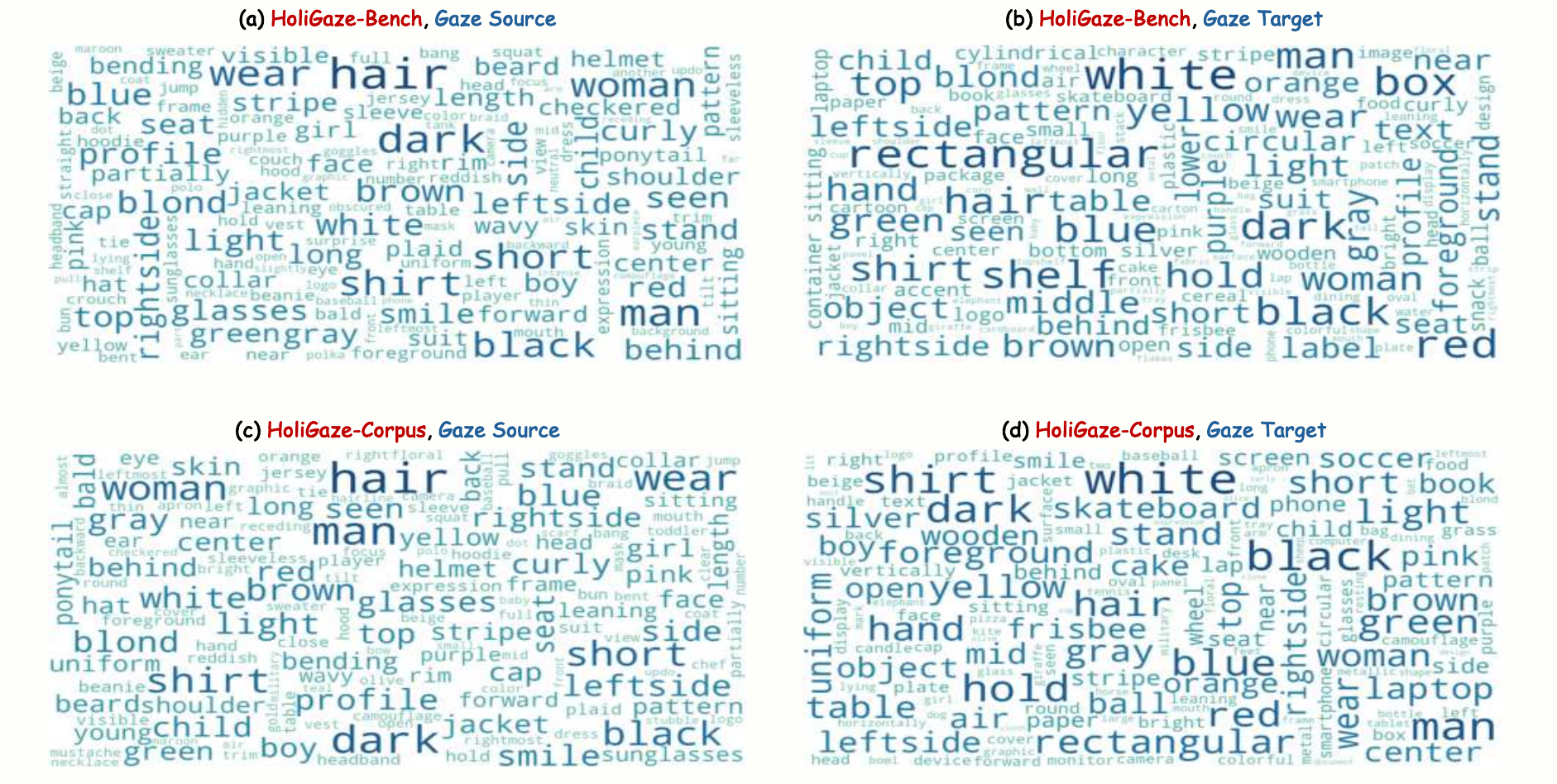}
    \caption{Word clouds of semantic descriptions of gaze sources and targets.}
    \label{fig:wordclouds}
\end{figure}

\paragraph{Semantic Diversity of Gaze Sources and Targets.}
Figure~\ref{fig:wordclouds} presents the word clouds of the semantic descriptions for gaze sources and targets. The source descriptions are predominantly centered around human subjects (e.g., ``man'', ``woman'', ``boy'', ``girl''), reflecting the dataset's focus on human gaze behaviors. In contrast, the target descriptions exhibit a much broader vocabulary, encompassing various objects (e.g., ``ball'', ``phone'', ``book''), scene elements (e.g., ``ground'', ``table''), and other people. This semantic richness ensures that models trained on HoliGaze can generalize to unconstrained, real-world gaze targets.

\begin{figure}[!ht]
\centering
    \includegraphics[width=0.92\textwidth]{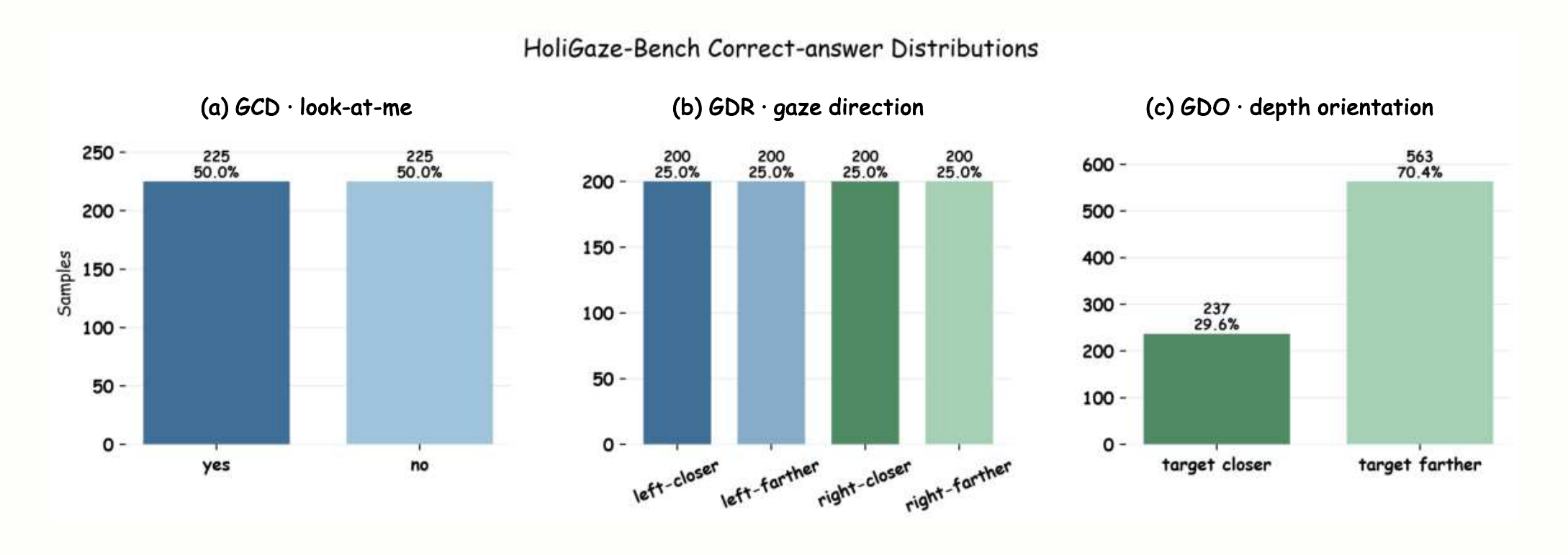}
    \caption{Distributions of correct answers across GCD, GDR, and GDO on HoliGaze-Bench.}
    \label{fig:gcdgdogdr_dist}
\end{figure}

\paragraph{Distributions of Source-Target Spatial Relationship.}
Figure~\ref{fig:gcdgdogdr_dist} reports the answer distributions of Gaze Direction Recognition (GDR) and Gaze Depth Orientation (GDO) on HoliGaze-Bench. GDR is balanced across the four camera-relative directional quadrants (left-closer, left-farther, right-closer, right-farther), covering diverse horizontal and depth orientations. GDO likewise balances targets that are closer to or farther from the camera than the gaze source, reducing the risk of trivial depth biases.

\begin{figure}[!ht]
\centering
    \includegraphics[width=0.88\textwidth]{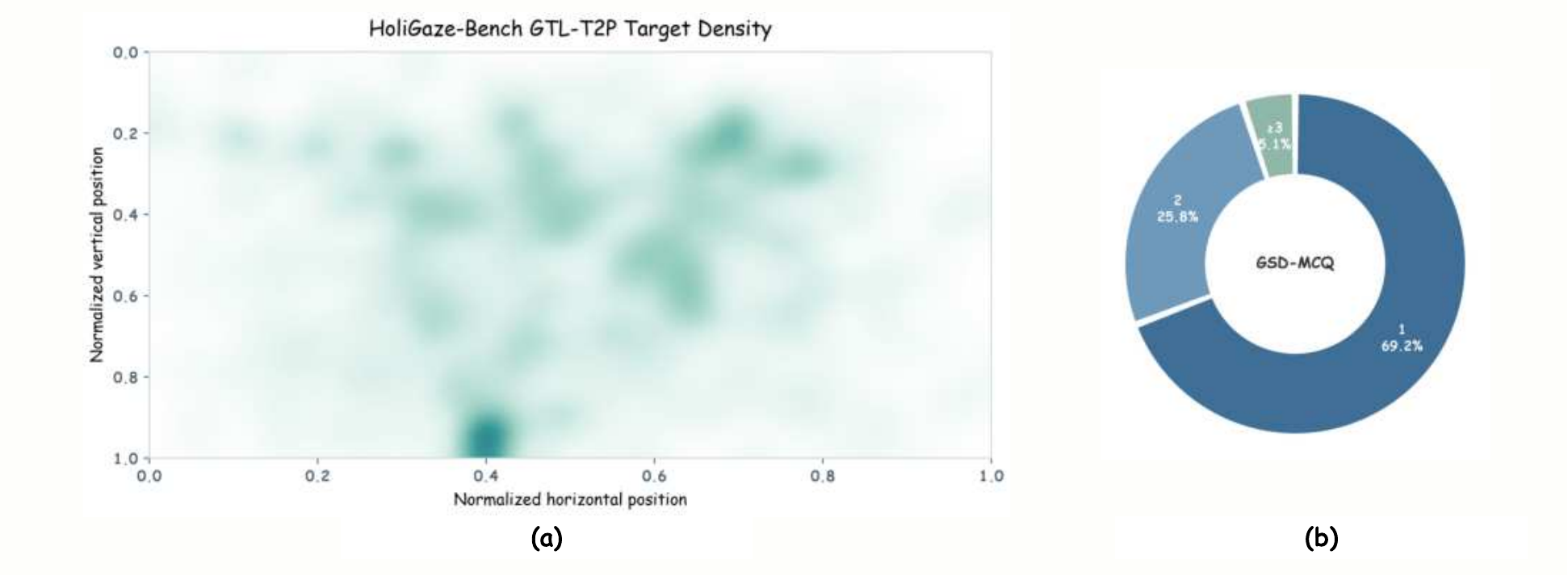}
    \caption{(a) Spatial distribution of gaze targets on HoliGaze-Bench. (b) Distribution of correct option counts on HoliGaze-Bench.}
    \label{fig:gsdgtl_dist}
\end{figure}

\paragraph{Spatial Distribution and Shared Attention.}
Figure~\ref{fig:gsdgtl_dist}(a) visualizes the spatial distribution of gaze targets in HoliGaze-Bench. The targets are widely distributed across the image plane, with a noticeable concentration towards the center and lower-middle regions, which is typical for human-object interactions and social scenarios. Furthermore, Figure~\ref{fig:gsdgtl_dist}(b) illustrates the distribution of correct option counts for the Gaze Source Description (GSD) task. While most targets are attended by a single person, a significant portion of targets are looked at by two or more people, highlighting the presence of shared attention and complex social dynamics.

\begin{figure}[!ht]
\centering
    \includegraphics[width=0.90\textwidth]{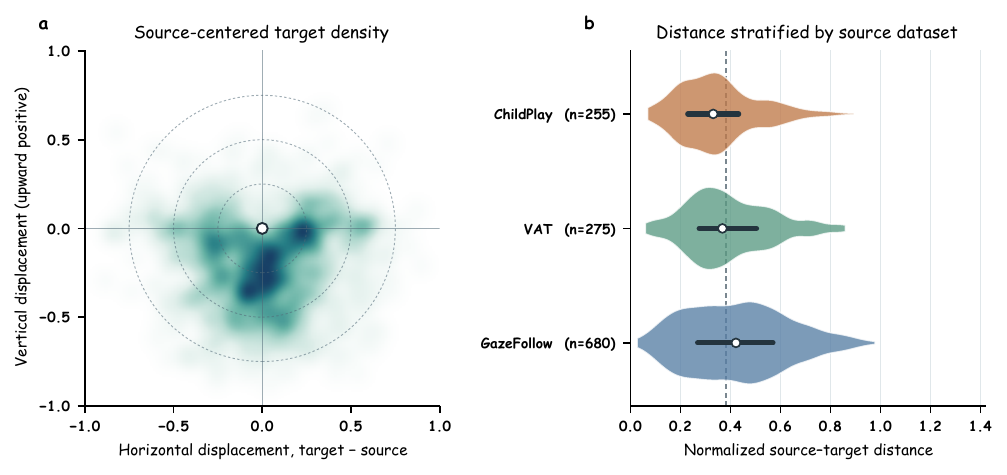}
    \caption{\textbf{Source--target joint geometry on HoliGaze-Bench.}
    \textbf{(a)} Log-scaled density of target displacement relative to the source-head center for 1,210 answerable GTL-T2P items (normalized coordinates; positive $y$ is above the source).
    \textbf{(b)} Source-target distances by dataset, with medians, IQRs, and the overall median marked.}
    \label{fig:source_target_joint_geometry}
\end{figure}

\paragraph{Source-Target Joint Geometry.}
Beyond the marginal target distribution in Figure~\ref{fig:gsdgtl_dist}, we examine the relative geometry between each gaze source and its target. Figure~\ref{fig:source_target_joint_geometry} places the source-head center at the origin and plots target displacement in normalized image coordinates for all 1,210 answerable GTL-T2P items. The horizontal distribution is nearly symmetric (48.5\% of targets lie to the left of the source), whereas 78.6\% of targets lie below the source head. This downward bias matches common gaze-following compositions, where faces appear above attended hands, objects, or interaction regions. The median source--target distance is 0.381 in normalized image coordinates, with dataset-wise medians of 0.422, 0.368, and 0.331 for GazeFollow, VideoAttentionTarget, and ChildPlay. The broad spreads across all three sources indicate substantial variation in both gaze direction and spatial extent.

\begin{figure}[!ht]
\centering
    \includegraphics[width=0.86\textwidth]{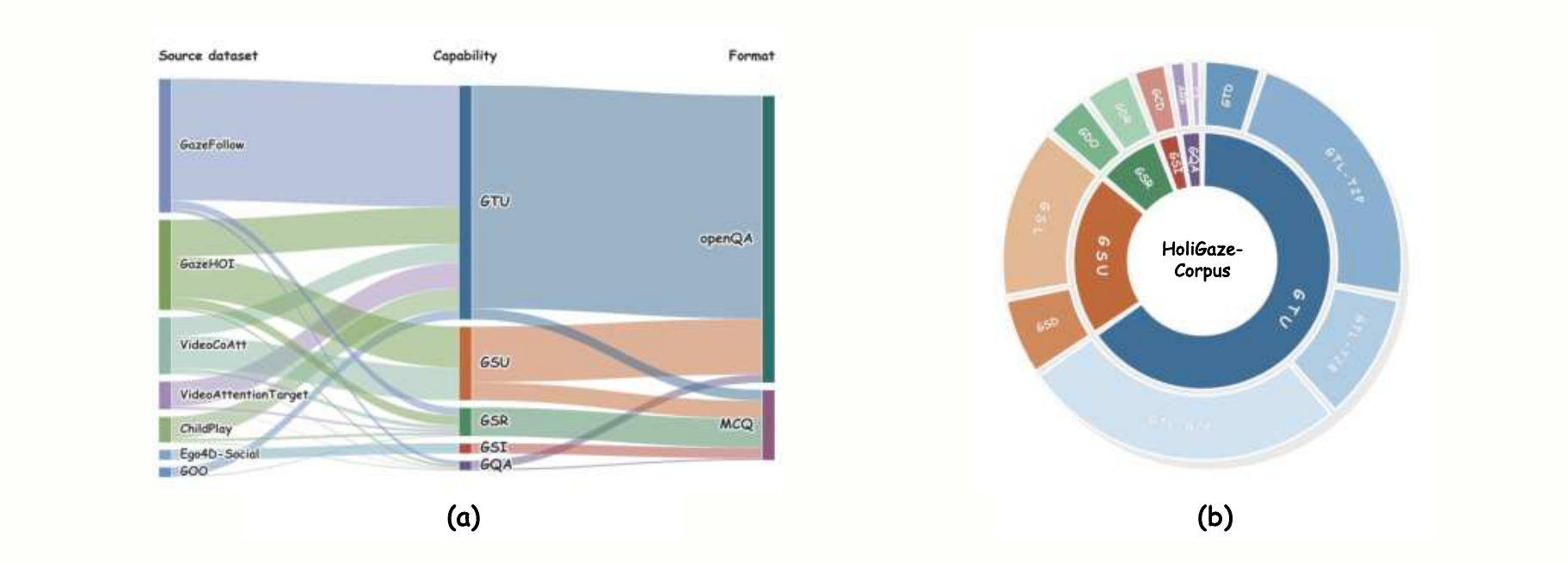}
    \caption{Statistics of HoliGaze-Corpus.}
    \label{fig:corpus_statistics}
\end{figure}

\begin{figure}[p]
  \centering
      \includegraphics[width=1.0\textwidth]{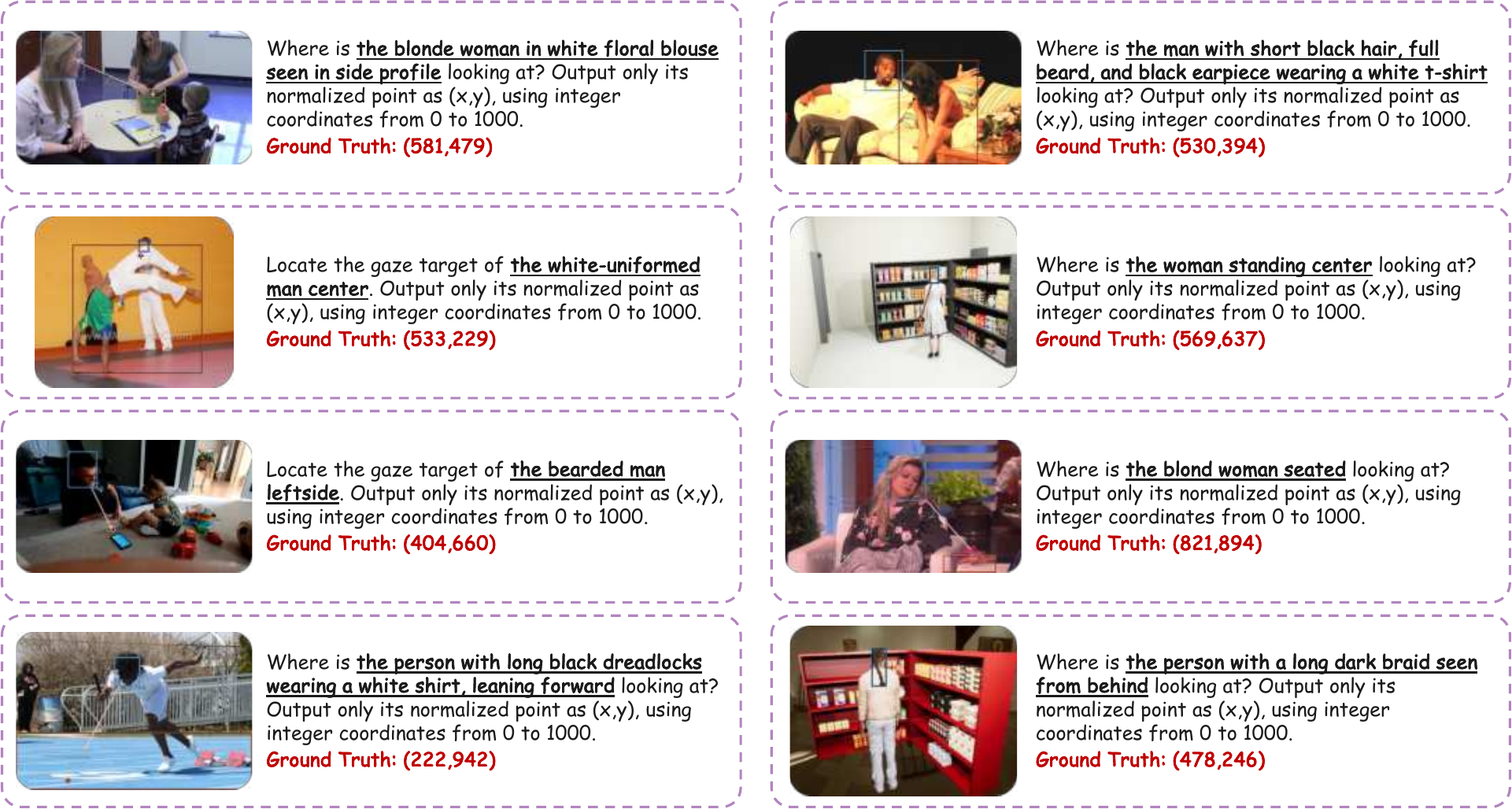}
      \caption{Sample cases of GTL-T2P in HoliGaze-Corpus.}
      \label{fig:corpus_gtl_t2p}
\end{figure}

\begin{figure}[p]
  \centering
      \includegraphics[width=1.0\textwidth]{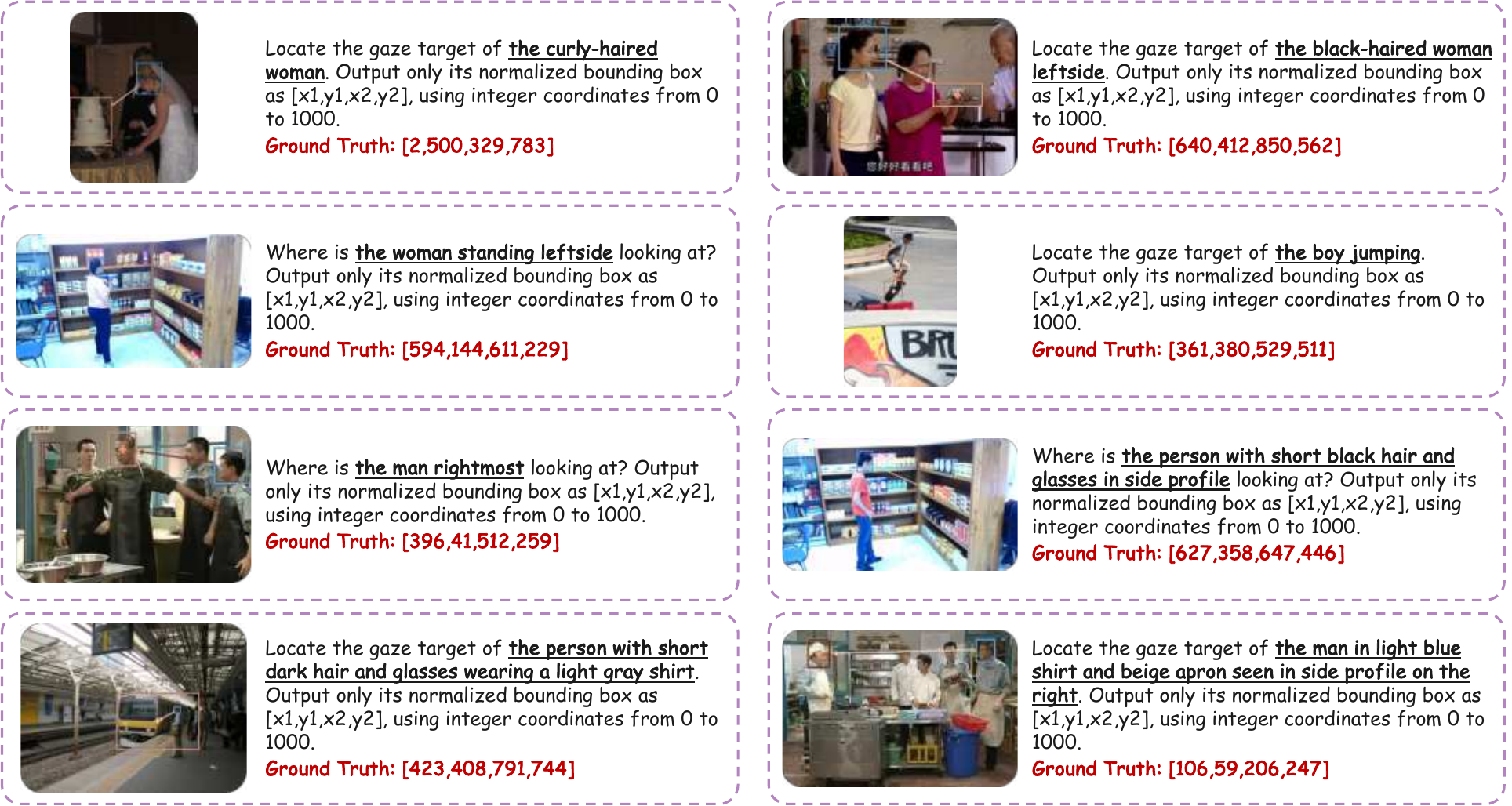}
      \caption{Sample cases of GTL-T2B in HoliGaze-Corpus.}
      \label{fig:corpus_gtl_t2b}
\end{figure}

\begin{figure}[p]
  \centering
      \includegraphics[width=1.0\textwidth]{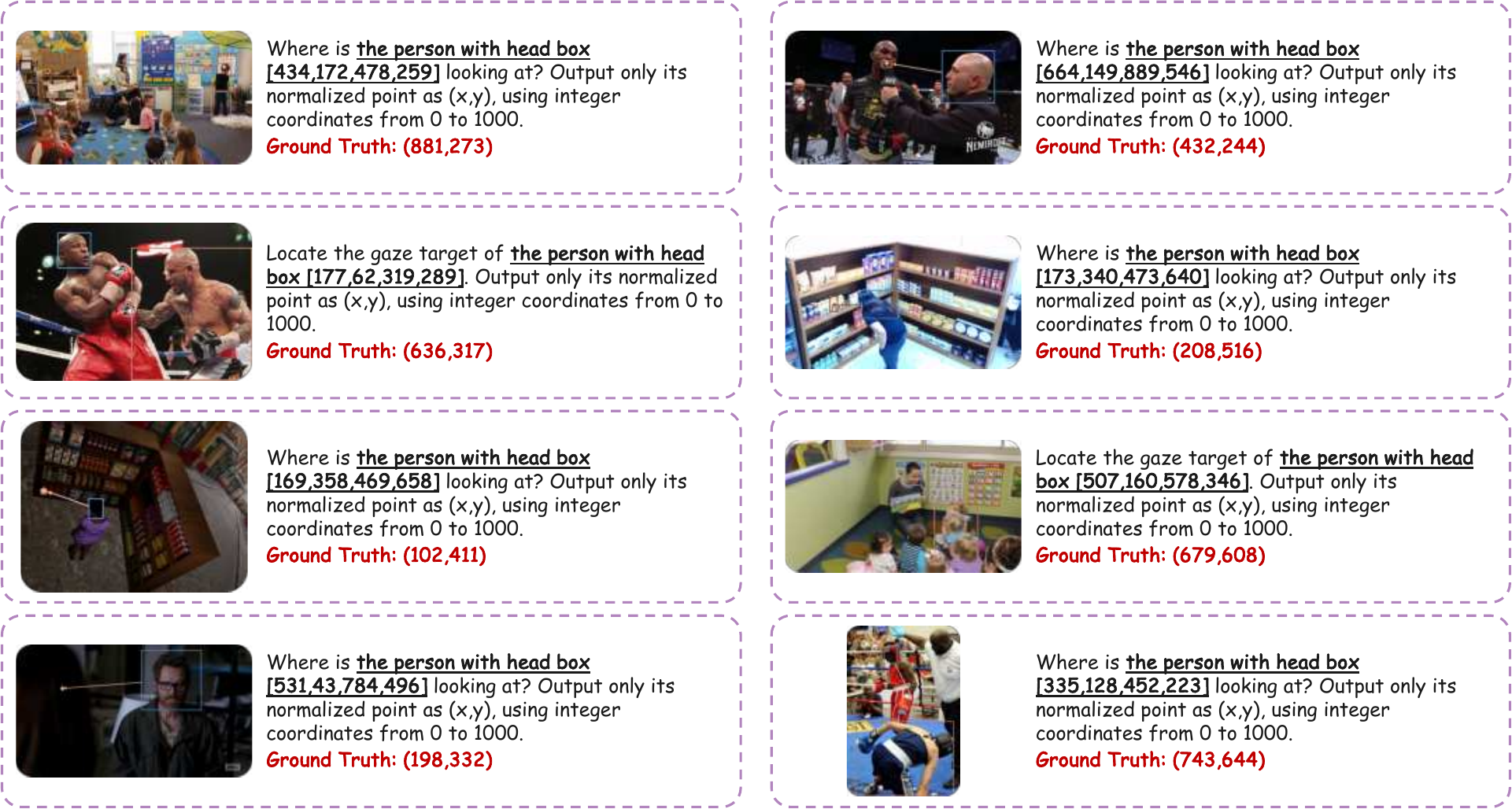}
      \caption{Sample cases of GTL-B2P in HoliGaze-Corpus.}
      \label{fig:corpus_gtl_b2p}
\end{figure}

\begin{figure}[p]
  \centering
      \includegraphics[width=0.92\textwidth]{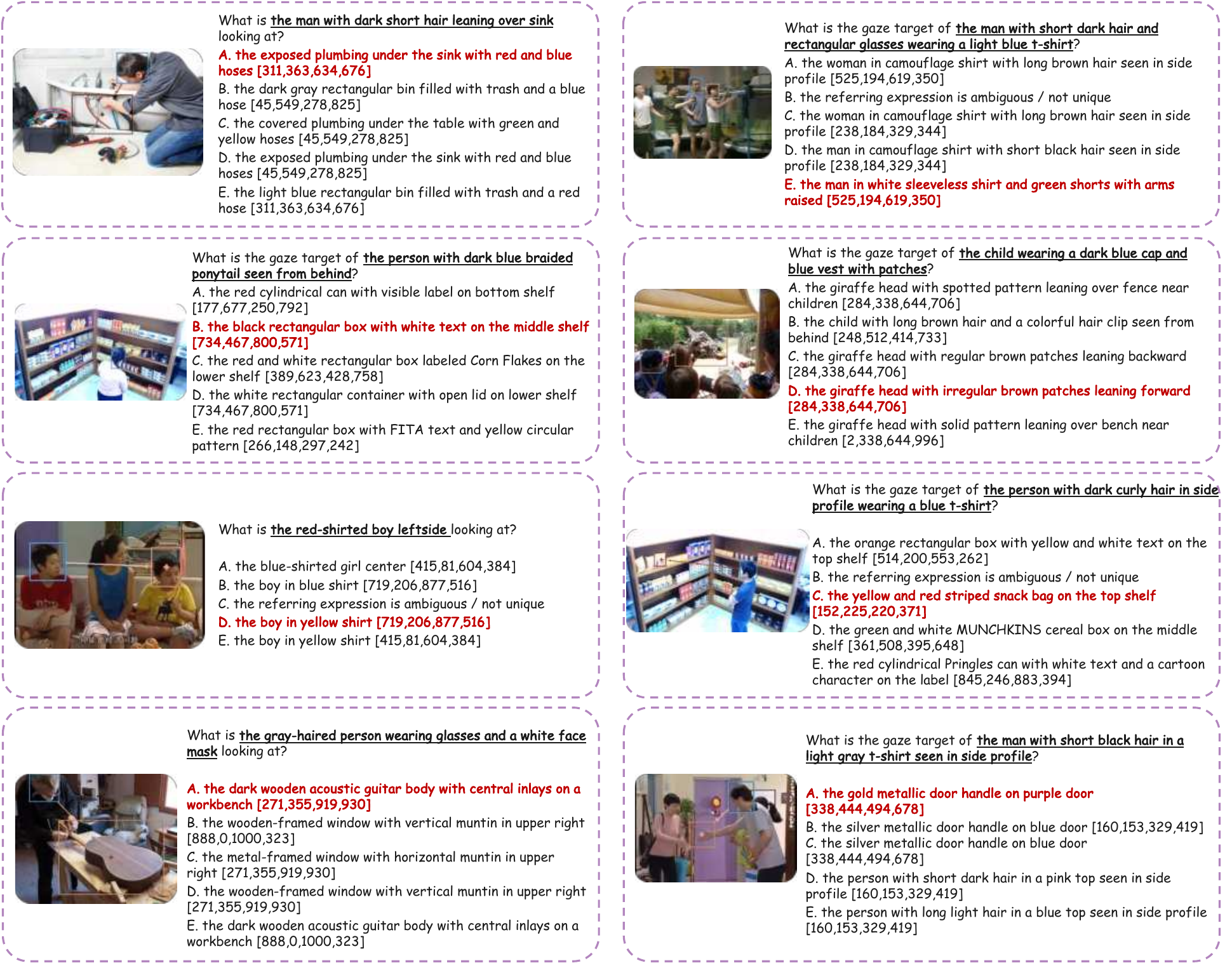}
      \caption{Sample cases of GTD-MCQ in HoliGaze-Corpus.}
      \label{fig:corpus_gtd_mcq}
\end{figure}

\begin{figure}[p]
  \centering
      \includegraphics[width=1.0\textwidth]{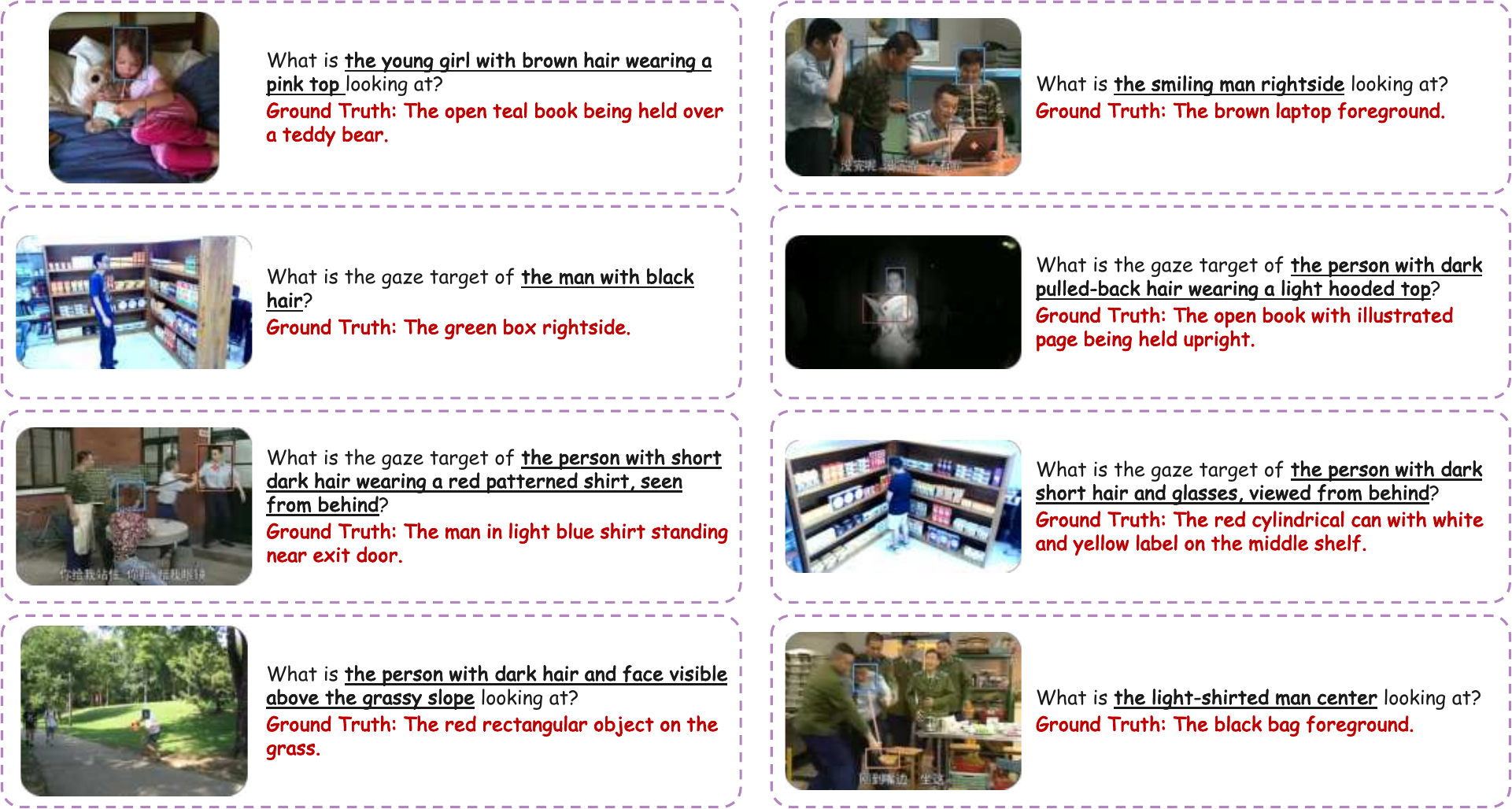}
      \caption{Sample cases of GTD-OpenQA in HoliGaze-Corpus.}
      \label{fig:corpus_gtd_openqa}
\end{figure}

\begin{figure}[p]
  \centering
      \includegraphics[width=1.0\textwidth]{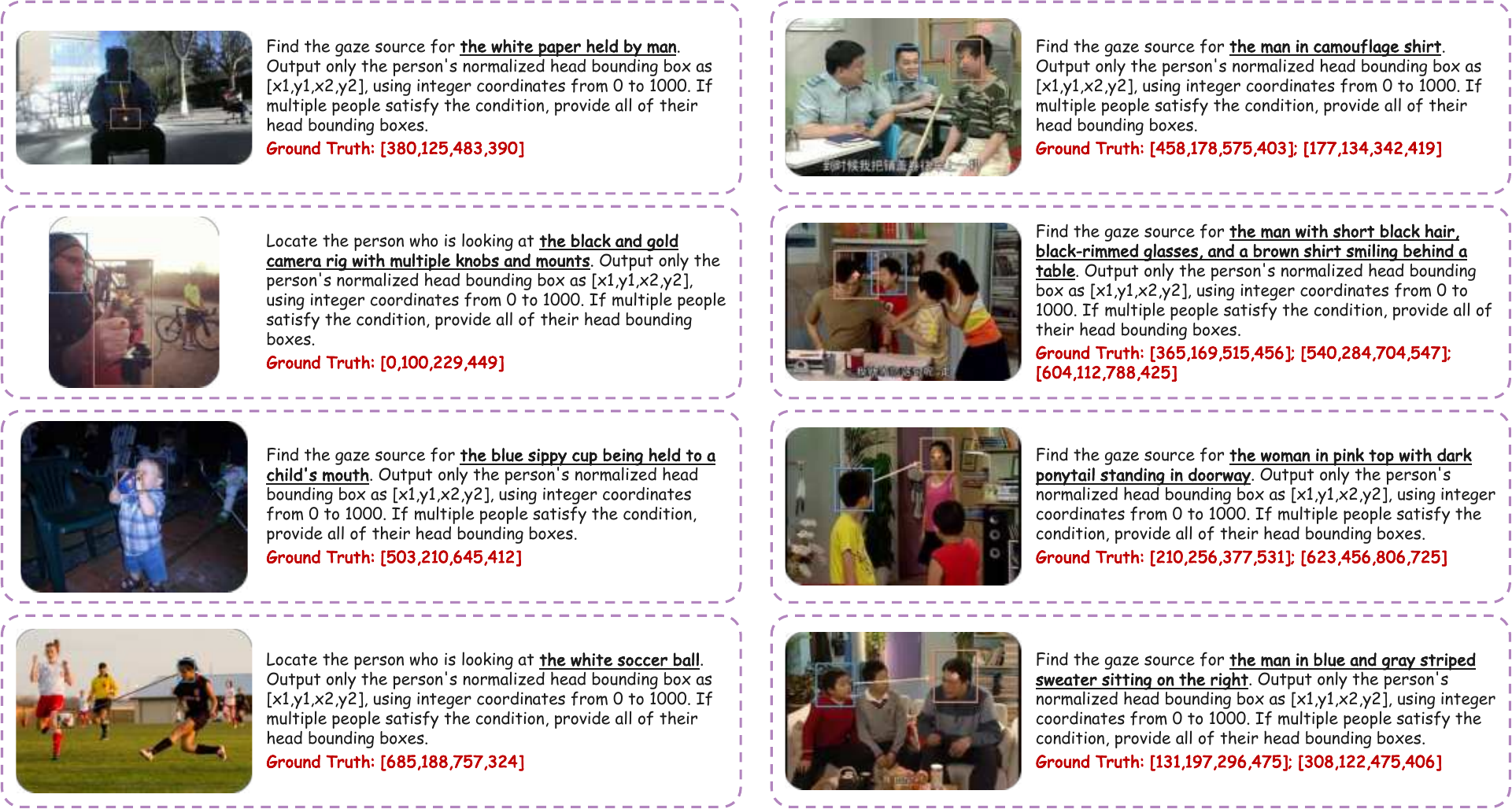}
      \caption{Sample cases of GSL in HoliGaze-Corpus.}
      \label{fig:corpus_gsl}
\end{figure}

\begin{figure}[p]
\centering
    \includegraphics[width=1.0\textwidth]{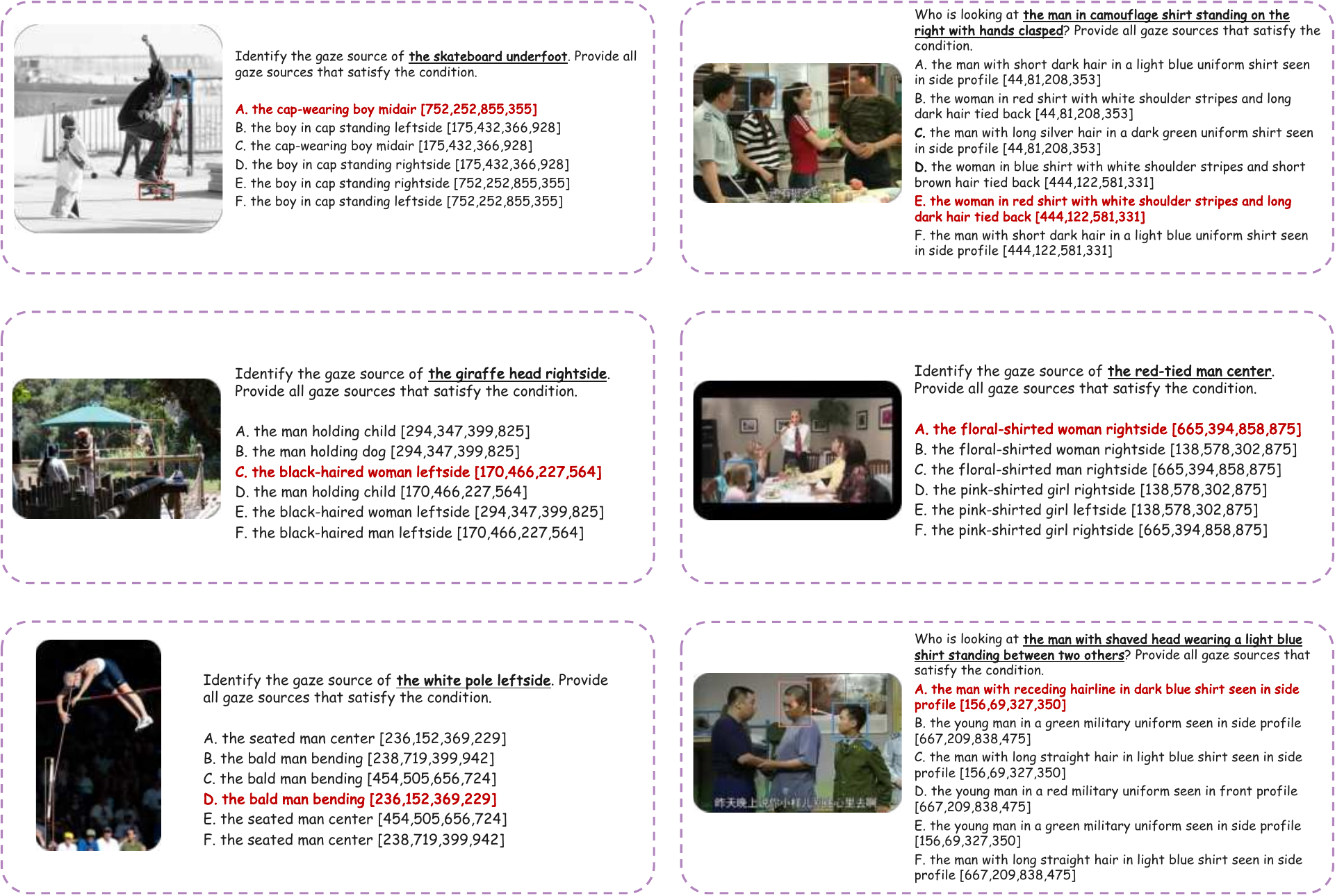}
    \caption{Sample cases of GSD-MCQ in HoliGaze-Corpus.}
    \label{fig:corpus_gsd_mcq}
\end{figure}

\begin{figure}[p]
  \centering
      \includegraphics[width=1.0\textwidth]{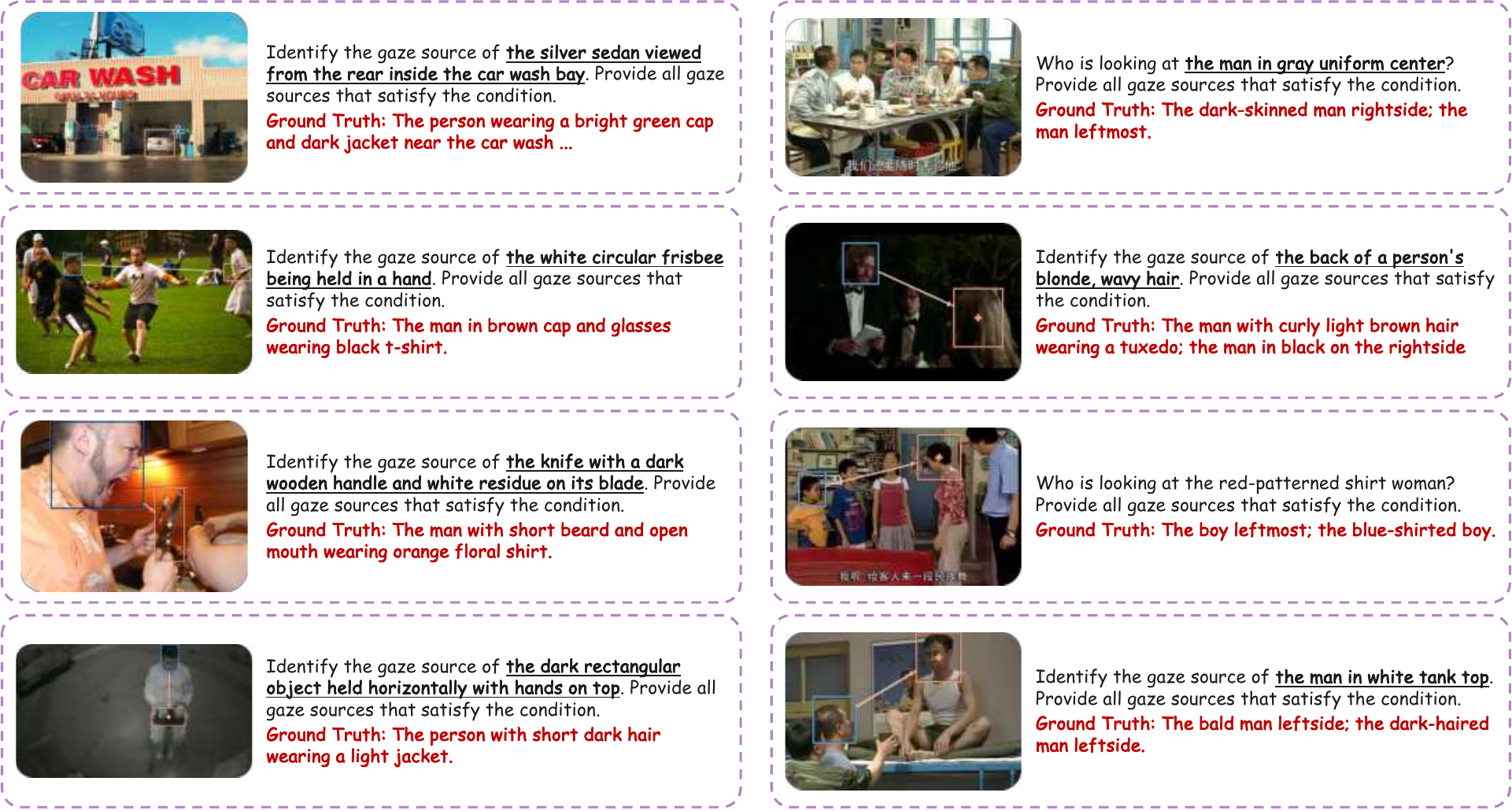}
      \caption{Sample cases of GSD-OpenQA in HoliGaze-Corpus.}
      \label{fig:corpus_gsd_openqa}
\end{figure}

\begin{figure}[p]
  \centering
      \includegraphics[width=1.0\textwidth]{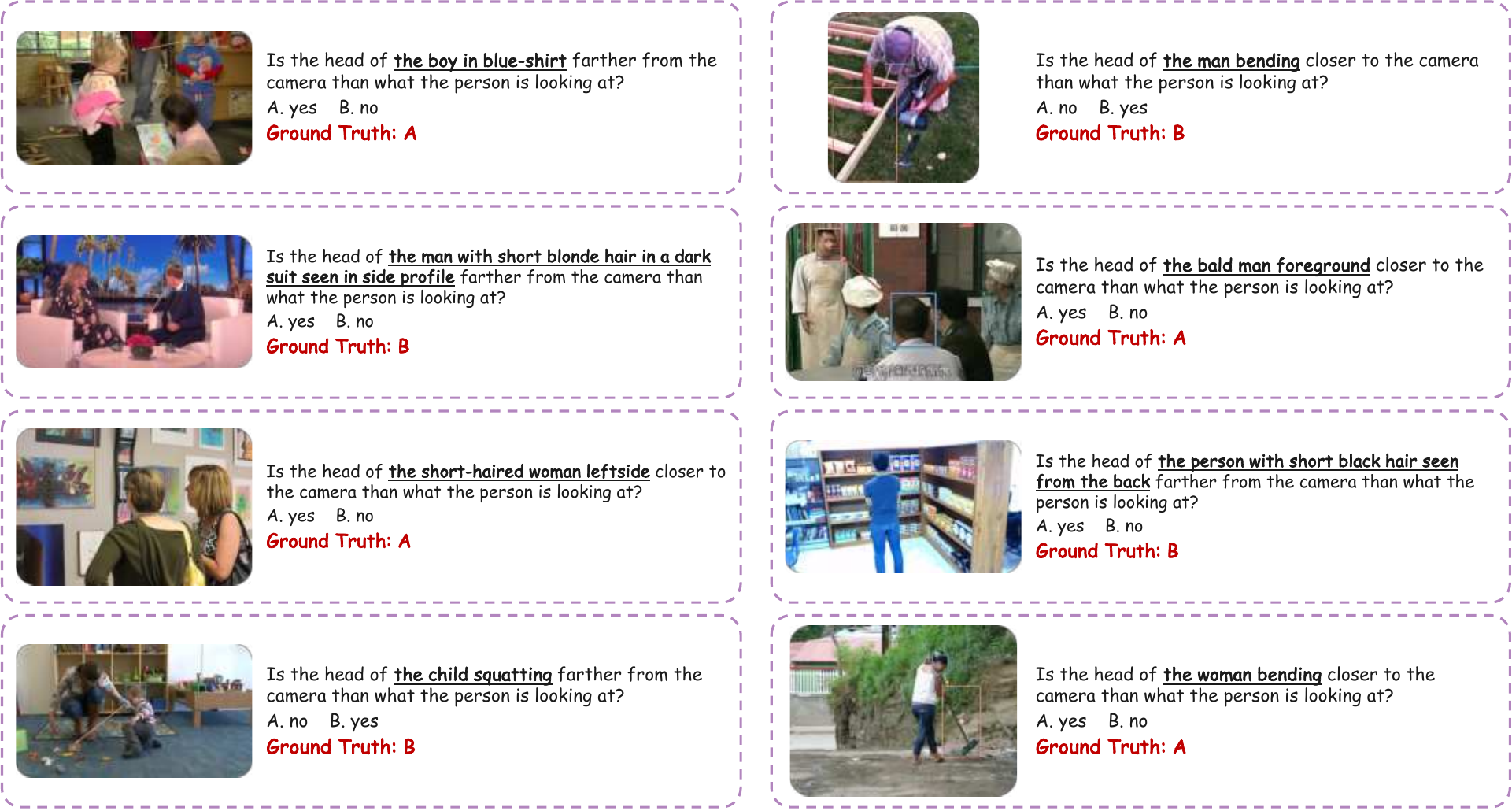}
      \caption{Sample cases of GDO in HoliGaze-Corpus.}
      \label{fig:corpus_gdo}
\end{figure}

\begin{figure}[p]
  \centering
      \includegraphics[width=1.0\textwidth]{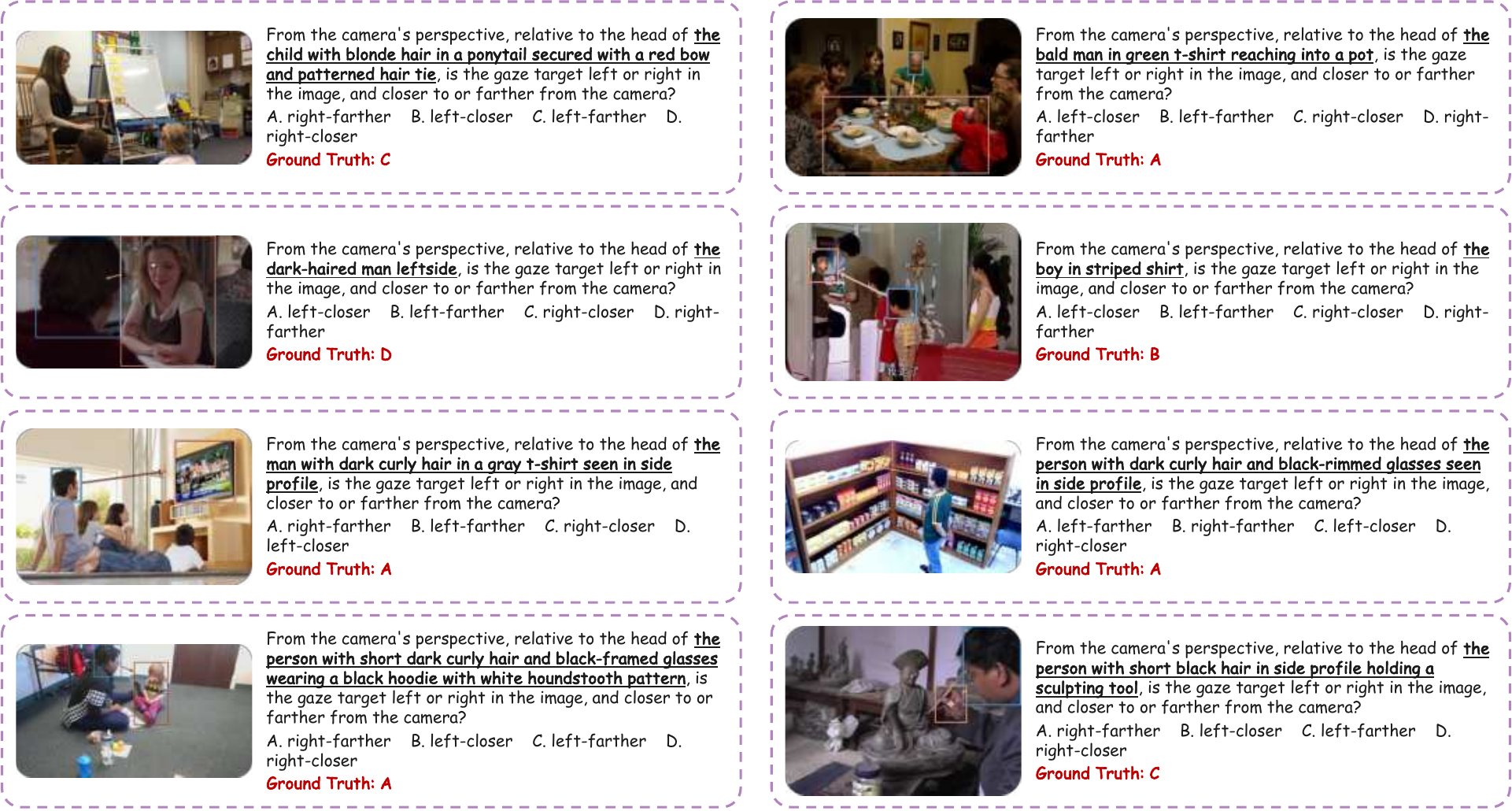}
      \caption{Sample cases of GDR in HoliGaze-Corpus.}
      \label{fig:corpus_gdr}
\end{figure}

\begin{figure}[p]
  \centering
      \includegraphics[width=1.0\textwidth]{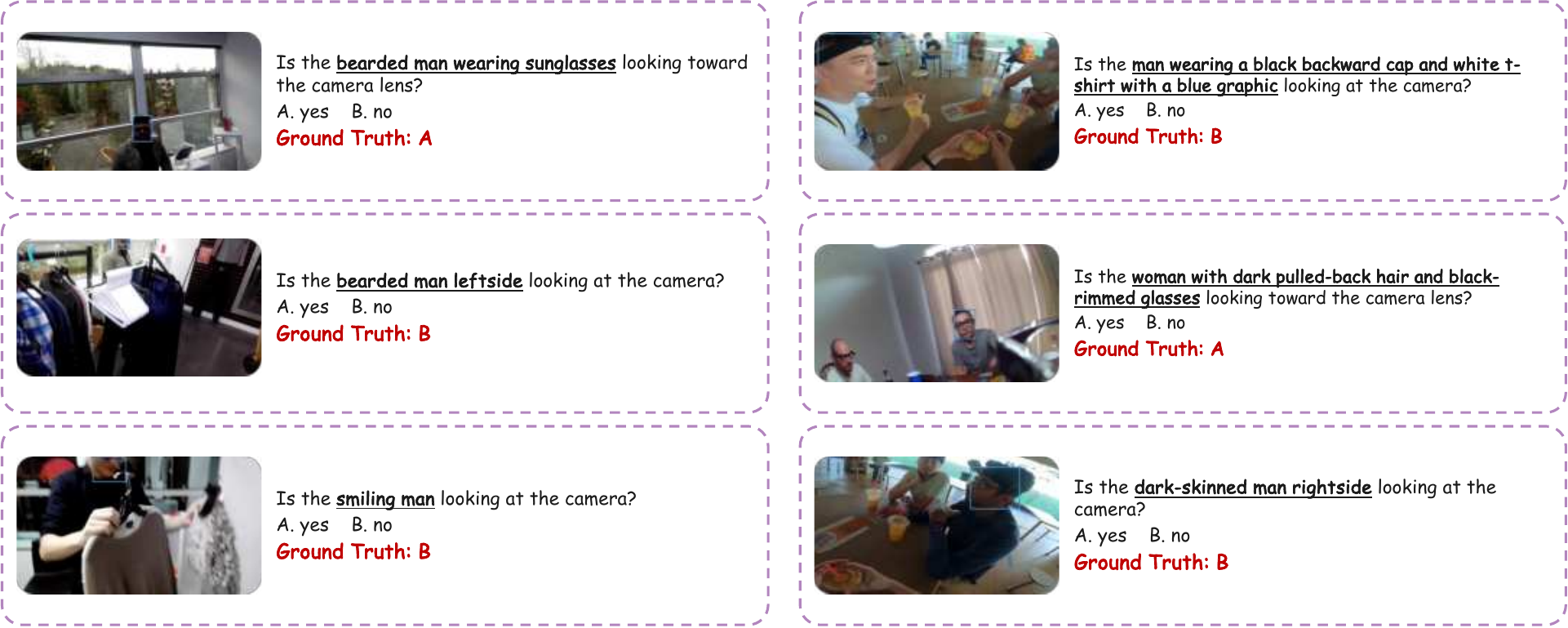}
      \caption{Sample cases of GCD in HoliGaze-Corpus.}
      \label{fig:corpus_gcd}
\end{figure}

\begin{figure}[p]
  \centering
      \includegraphics[width=1.0\textwidth]{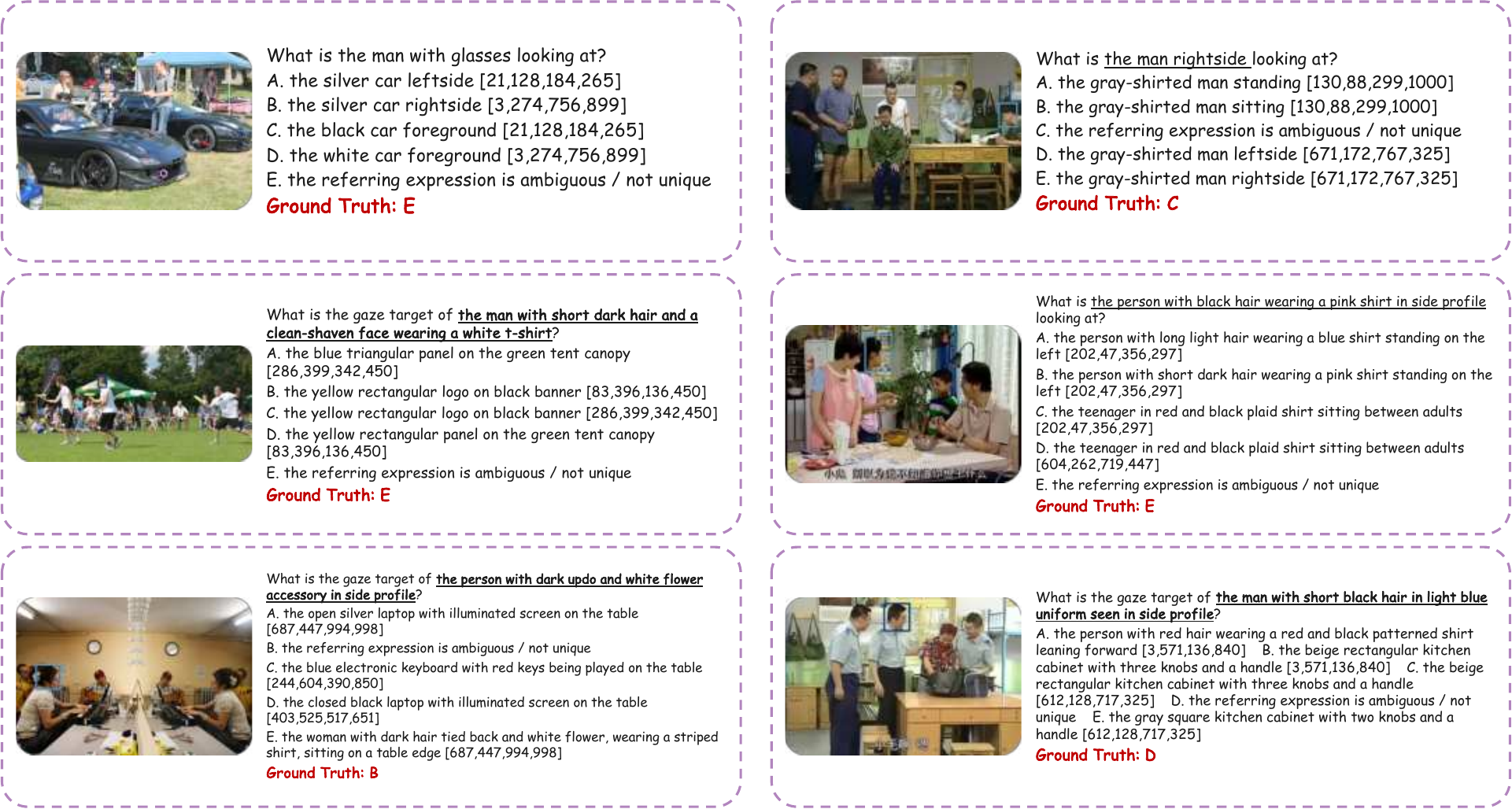}
      \caption{Sample cases of ARR-MCQ in HoliGaze-Corpus.}
      \label{fig:corpus_arr_mcq}
\end{figure}

\begin{figure}[p]
  \centering
      \includegraphics[width=1.0\textwidth]{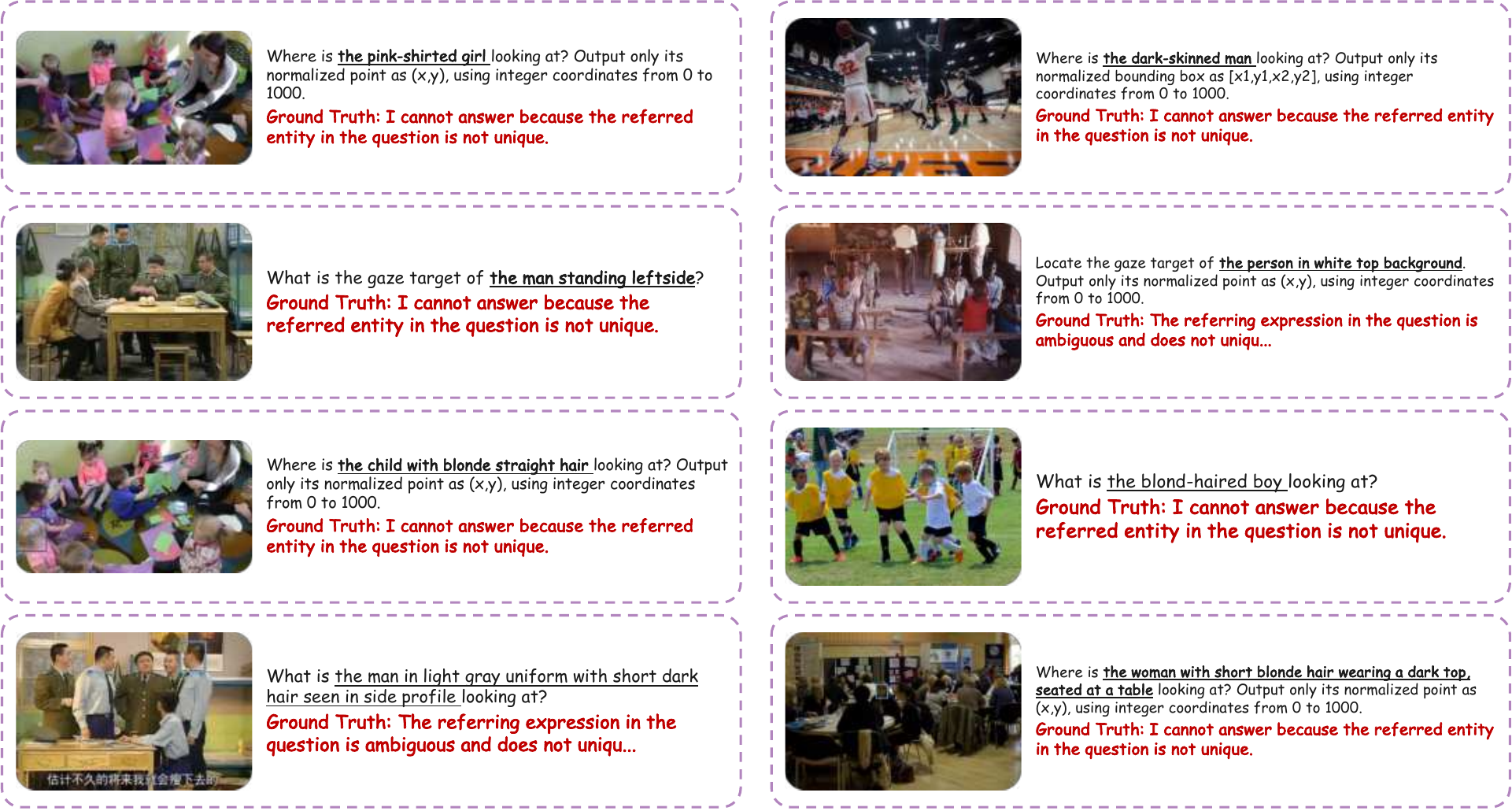}
      \caption{Sample cases of ARR-OpenQA in HoliGaze-Corpus.}
      \label{fig:corpus_arr_openqa}
      \vspace{-2mm}
\end{figure}

\begin{figure}[p]
\centering
    \includegraphics[width=1.0\textwidth]{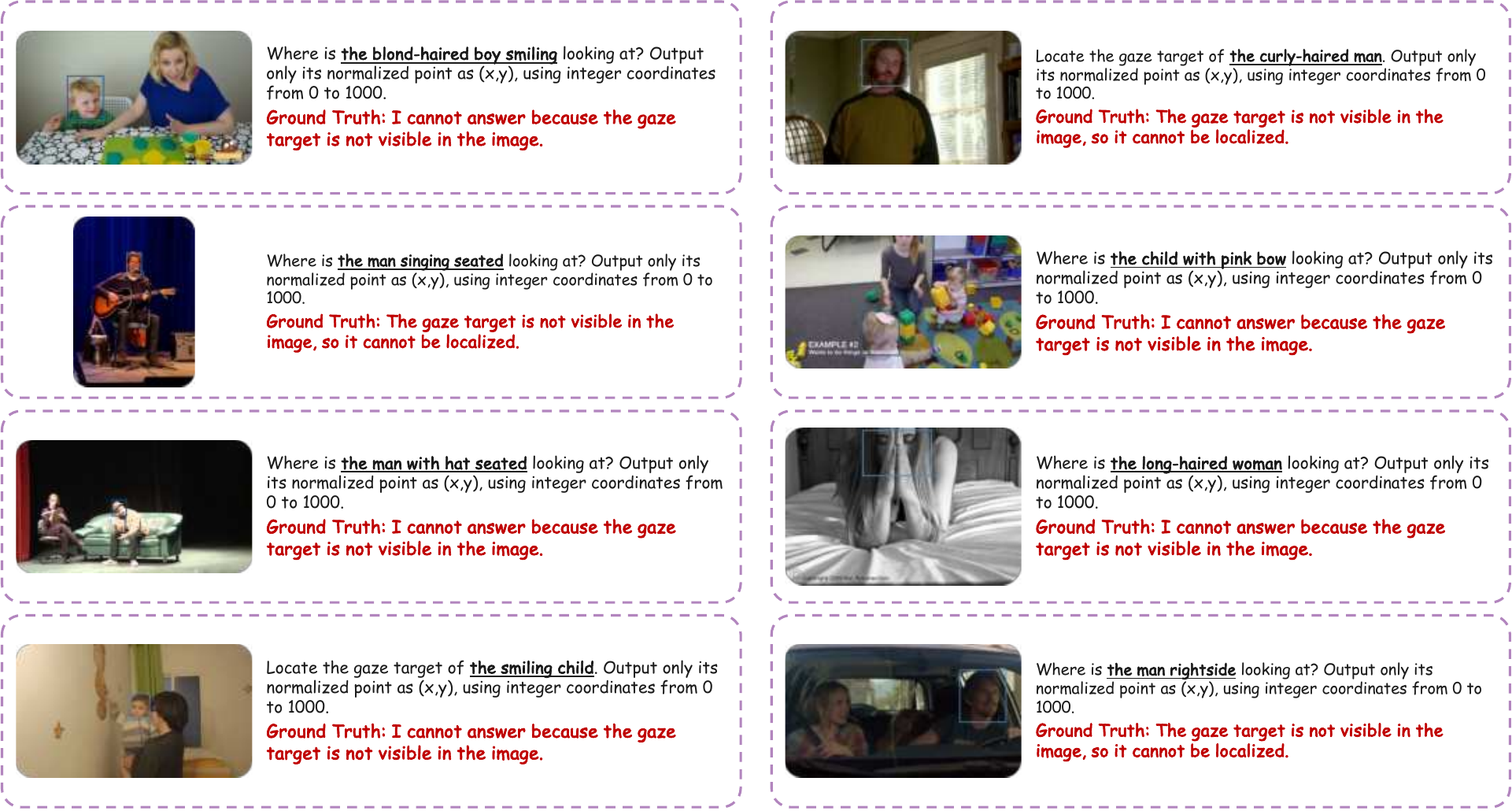}
    \caption{Sample cases of PV in HoliGaze-Corpus.}
    \label{fig:corpus_pv}
\end{figure}

\paragraph{Details of HoliGaze-Corpus and Source-Dataset Composition.}
Figure~\ref{fig:corpus_statistics} provides a comprehensive overview of the statistics for HoliGaze-Corpus, detailing the distribution of different task types and answer lengths. The corpus contains a balanced mix of localization and reasoning tasks, providing a robust foundation for training holistic gaze following models. Furthermore, Table~\ref{tab:app_source} summarizes the composition of HoliGaze from various source datasets. To construct a diverse and comprehensive benchmark, we aggregate data from multiple domains, including everyday scenes (GazeFollow, VideoAttentionTarget, VideoCoAtt), human-object interactions (GazeHOI), child-centric environments (ChildPlay), social interactions (Ego4D-Social), and retail environments (GOO). Moreover, Figures~\ref{fig:corpus_gtl_t2p} to \ref{fig:corpus_pv} further present diverse data samples across the 9 task formats in the HoliGaze-Corpus. These examples illustrate the high-quality annotations, varied geometric distributions, and rich semantic context provided by our scalable data engine, which collectively empower the holistic capabilities of LookAnything.

\begin{table}[!ht]
\centering
\caption{\textbf{Composition of data sources in HoliGaze.}}
\label{tab:app_source}
\setlength{\tabcolsep}{8pt}
\renewcommand{\arraystretch}{1.08}
\small
\begin{tabular}{lrrrrrr}
\toprule
Source Dataset & HoliGaze-Corpus & \% & HoliGaze-Bench & \% & Overall & \% \\
\midrule
GazeFollow & $170{,}299$ & $56.20$ & $940$ & $11.75$ & $171{,}239$ & $55.06$ \\
GazeHOI & $42{,}497$ & $14.03$ & $2{,}404$ & $30.05$ & $44{,}901$ & $14.44$ \\
VideoAttentionTarget & $30{,}094$ & $9.93$ & $540$ & $6.75$ & $30{,}634$ & $9.85$ \\
VideoCoAtt & $26{,}960$ & $8.90$ & $1{,}696$ & $21.20$ & $28{,}656$ & $9.21$ \\
ChildPlay & $21{,}765$ & $7.18$ & $460$ & $5.75$ & $22{,}225$ & $7.15$ \\
GOO-Syn & $6{,}385$ & $2.11$ & $440$ & $5.50$ & $6{,}825$ & $2.20$ \\
GOO-Real & $0$ & $0.00$ & $1{,}070$ & $13.38$ & $1{,}070$ & $0.34$ \\
Ego4D-Social & $5{,}000$ & $1.65$ & $450$ & $5.62$ & $5{,}450$ & $1.75$ \\
\midrule
\textbf{Total} & $\mathbf{303{,}000}$ & $100$ & $\mathbf{8{,}000}$ & $100$ & $\mathbf{311{,}000}$ & $100$ \\
\bottomrule
\end{tabular}
\end{table}

\section{Additional Dataset Construction Details}
\label{sec:appendix_data_engine}

In this section, we provide further details on the construction process of the HoliGaze dataset.

\subsection{Data Collection}
\label{subsec:app_data_collection}
We collect static images and video keyframes from eight source datasets, introduced respectively as follows:
\begin{itemize}
    \item \textbf{GazeFollow}~\citep{recasens2015they}: As a pioneering large-scale dataset for gaze following in the wild, GazeFollow comprises images of diverse unconstrained scenes, providing annotations of head bounding boxes and the corresponding target points the individuals are looking at.
    \item \textbf{VideoAttentionTarget (VAT)}~\citep{chong2020detecting}: VAT extends gaze target detection to the temporal domain, offering a comprehensive dataset of videos that depict complex, everyday social interactions. It provides annotations for attended visual targets, including robust handling of out-of-frame gaze instances.
    \item \textbf{ChildPlay}~\citep{tafasca2023childplay}: Specifically focusing on child-centric environments, this dataset captures the unique gaze behaviors of children engaged in unscripted play and social interactions in naturalistic settings.
    \item \textbf{GOO-Real \& GOO-Syn}~\citep{tomas2021goo}: Tailored for studying human attention in retail environments, these twin datasets formally introduce the task of gaze object prediction. GOO-Real captures shoppers in real-world grocery aisles, while GOO-Syn provides a large-scale synthetic counterpart. Both offer dense, object-level bounding box annotations for gaze targets.
    \item \textbf{GazeHOI}~\citep{tafasca2024toward}: Proposed for semantic gaze target detection, GazeHOI provides rich annotations linking human gaze to specific objects they are interacting with. It facilitates the joint reasoning of visual attention and physical actions in human-object interaction scenarios.
    \item \textbf{VideoCoAtt}~\citep{fan2018inferring}: Emphasizing shared attention dynamics, VideoCoAtt features video sequences where multiple individuals in social scenes simultaneously look at a common target, enabling the study of multi-person co-gaze.
    \item \textbf{Ego4D-Social}~\citep{grauman2022ego4d}: Extracted from the massive Ego4D egocentric video dataset, this subset focuses on first-person social interactions. It provides critical annotations for mutual gaze and camera contact, framing visual attention from the camera wearer's perspective.
\end{itemize}
In summary, these datasets collectively cover a highly rich and diverse array of scenarios, ranging from unconstrained everyday scenes to specialized retail aisles. By integrating these disparate sources, we repurpose them to support a more comprehensive suite of holistic gaze-centric tasks.

\paragraph{Schema Unification.} To harmonize the diverse data formats, we map the heterogeneous annotations from all source datasets into a standardized, shared schema. This unified schema encompasses an RGB image, a source head region (represented by a bounding box), a gaze target location (represented as a 2D point, a bounding box, or an out-of-frame indicator), a monocular depth map, semantic referring expressions for both the source and target, as well as flags indicating referential ambiguity and depth polarity. Since each native dataset only contributes a subset of these fields, the missing annotations are populated at scale through an automated completion pipeline, as detailed in \S\ref{subsec:app_annotation_completion}.

\subsection{Annotation Completion}
\label{subsec:app_annotation_completion}
We employ several off-the-shelf models to automatically populate the missing fields within our unified schema.

\paragraph{Head Masks.} For datasets already providing head bounding boxes (e.g., GazeFollow, VAT), we directly convert these boxes into binary masks. For datasets providing only a head point (e.g., GOO), we utilize SAM3~\citep{carion2026sam} prompted with the text ``head'' to recover the head regions. We select the highest-scoring mask containing the annotated point. If none match, we fall back to the mask whose tight bounding box contains the point, or the box nearest to the point within a strict distance threshold. The final head mask is then defined as the rasterized rectangle of this selected bounding box.

\paragraph{Scene Depth.} To obtain structural scene priors, Depth Anything v3 (DA3)~\citep{lin2025depth} is employed to estimate a robust monocular depth map for every image. 

\begin{figure}[p]
  \centering
      \includegraphics[width=0.9\textwidth]{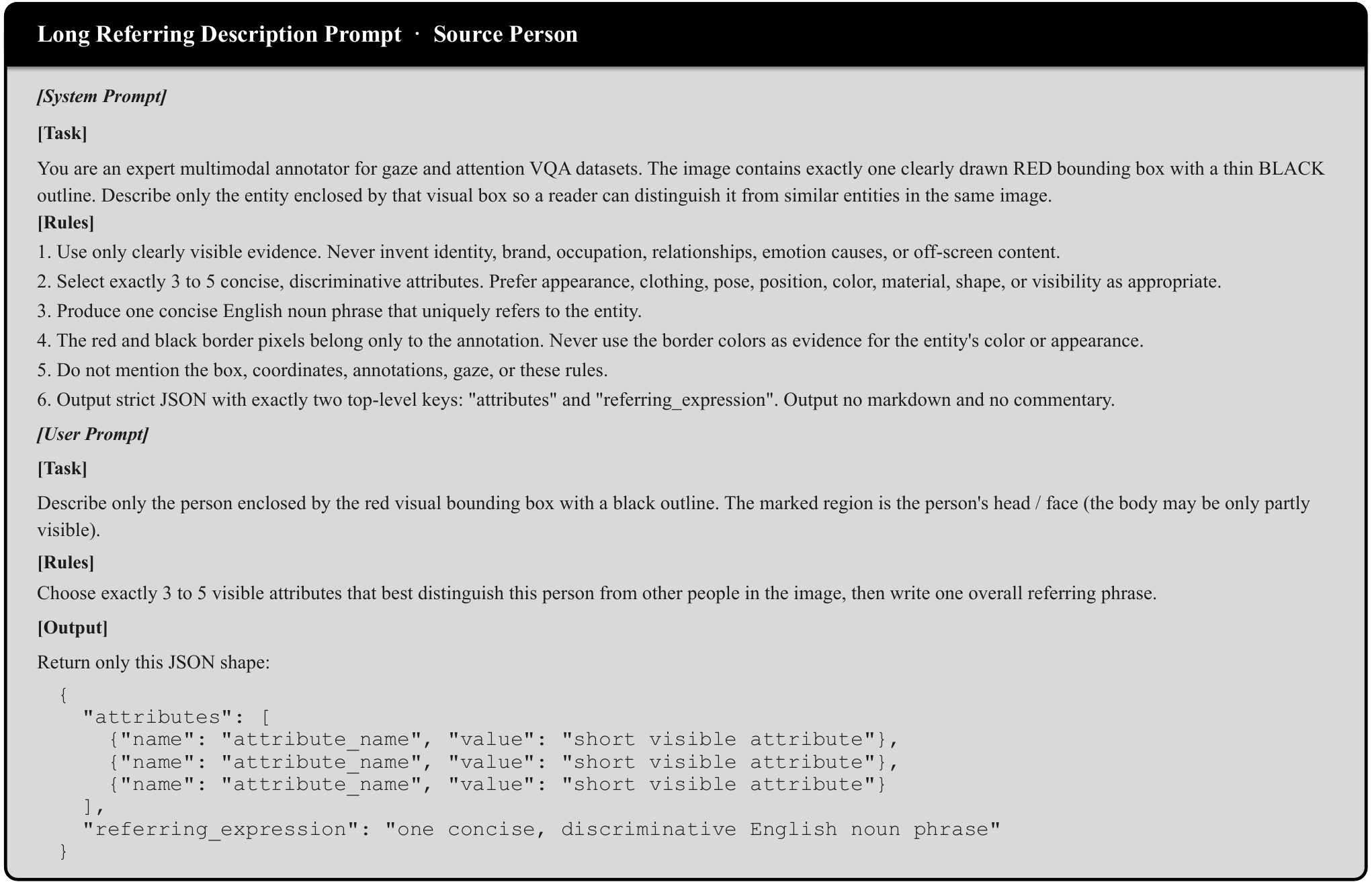}
      \caption{Prompt template for long referring description generation of source person.}
      \label{fig:prompt_long_desc_source_person}
\end{figure}

\begin{figure}[p]
  \centering
      \includegraphics[width=0.9\textwidth]{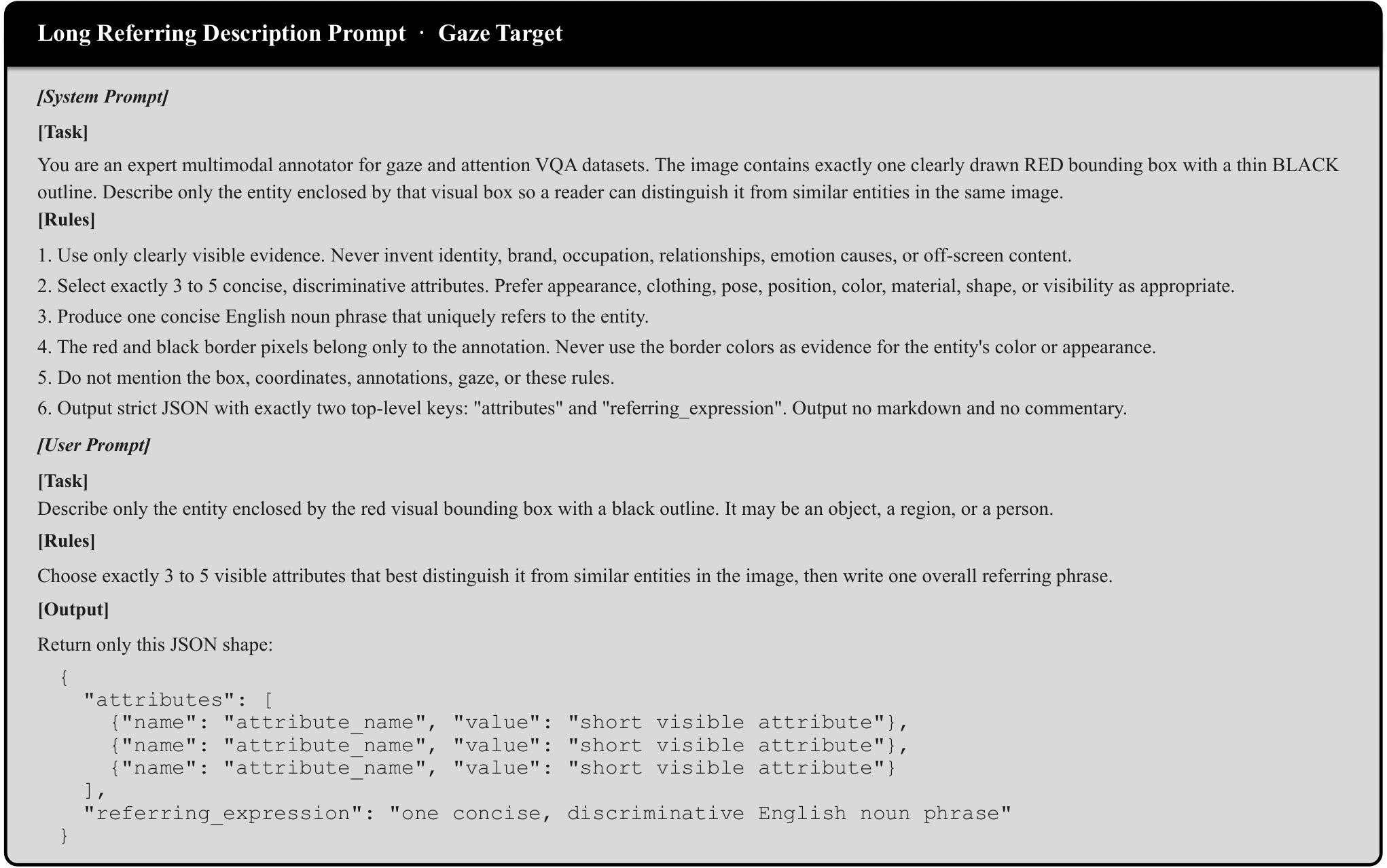}
      \caption{Prompt template for long referring description generation of gaze target.}
      \label{fig:prompt_long_desc_target}
\end{figure}

\begin{figure}[p]
  \centering
      \includegraphics[width=1.0\textwidth]{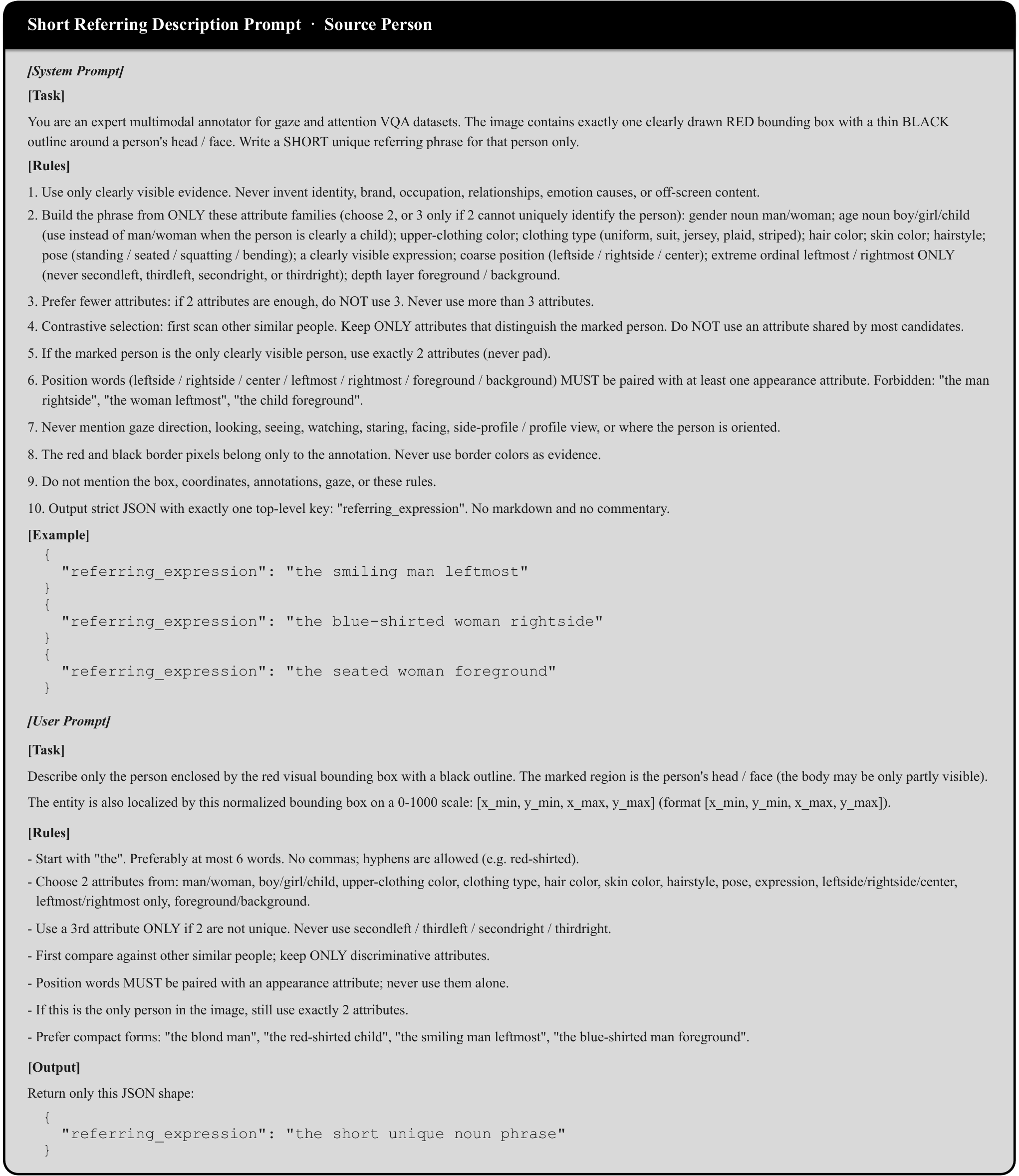}
      \caption{Prompt template for short referring description generation of source person.}
      \label{fig:prompt_short_desc_source_person}
\end{figure}

\begin{figure}[p]
  \centering
      \includegraphics[width=1.0\textwidth]{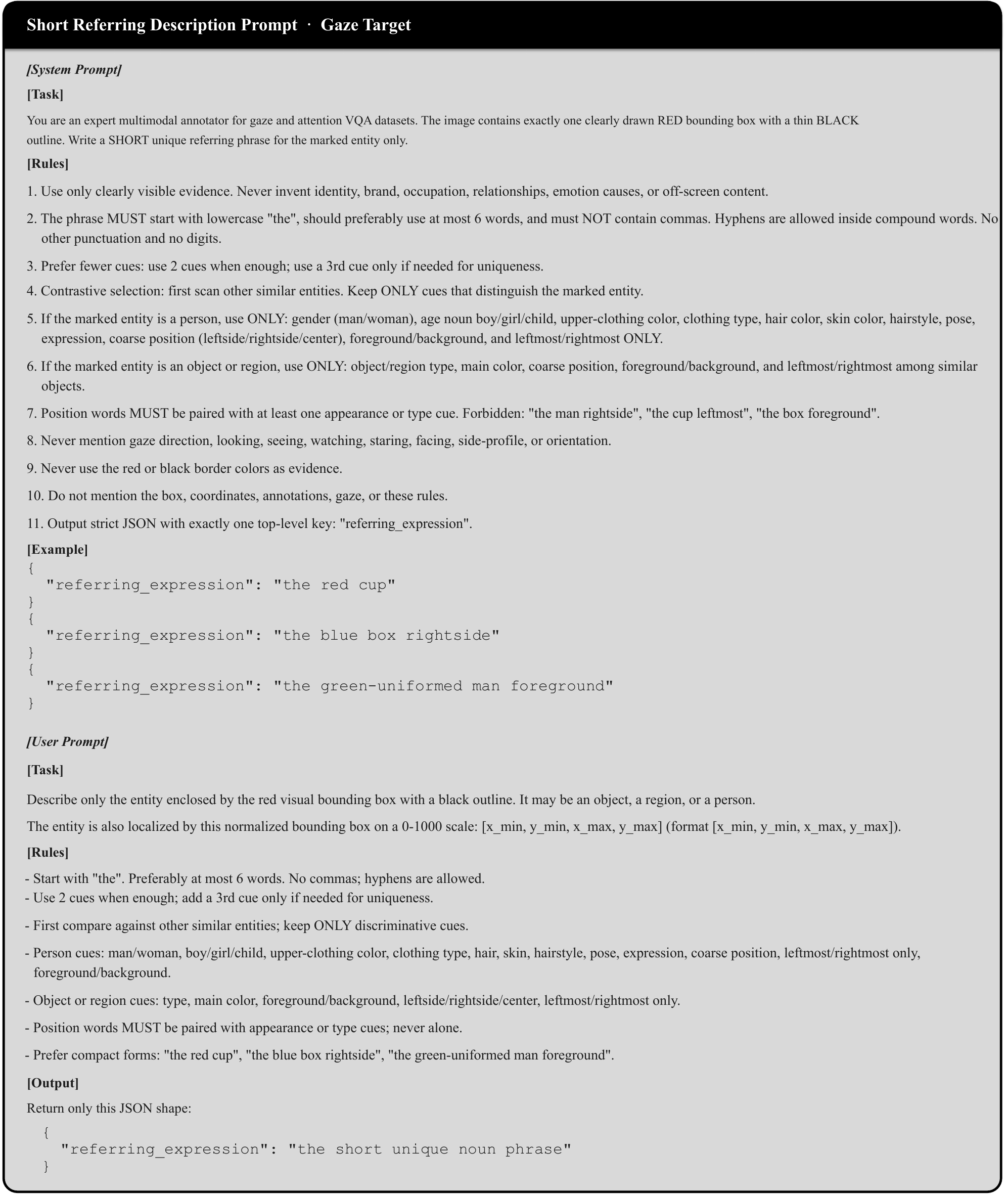}
      \caption{Prompt template for short referring description generation of gaze target.}
      \label{fig:prompt_short_desc_target}
\end{figure}

\paragraph{Referring Expressions.} We utilize the Gemini API~\citep{comanici2025gemini} to generate referring expressions for both the source person and gaze target from visual box prompts. Crucially, the image shown to the model is modified to carry a single red bounding box with a thin black outline to explicitly highlight the referred entity. The model writes one long expression and one short expression for each valid entity. The prompt templates for generating long and short referring descriptions of gaze sources and targets are depicted in Figure~\ref{fig:prompt_long_desc_source_person}, Figure~\ref{fig:prompt_long_desc_target}, Figure~\ref{fig:prompt_short_desc_source_person}, and Figure~\ref{fig:prompt_short_desc_target}, respectively.

\begin{figure}[p]
  \centering
      \includegraphics[width=1.0\textwidth]{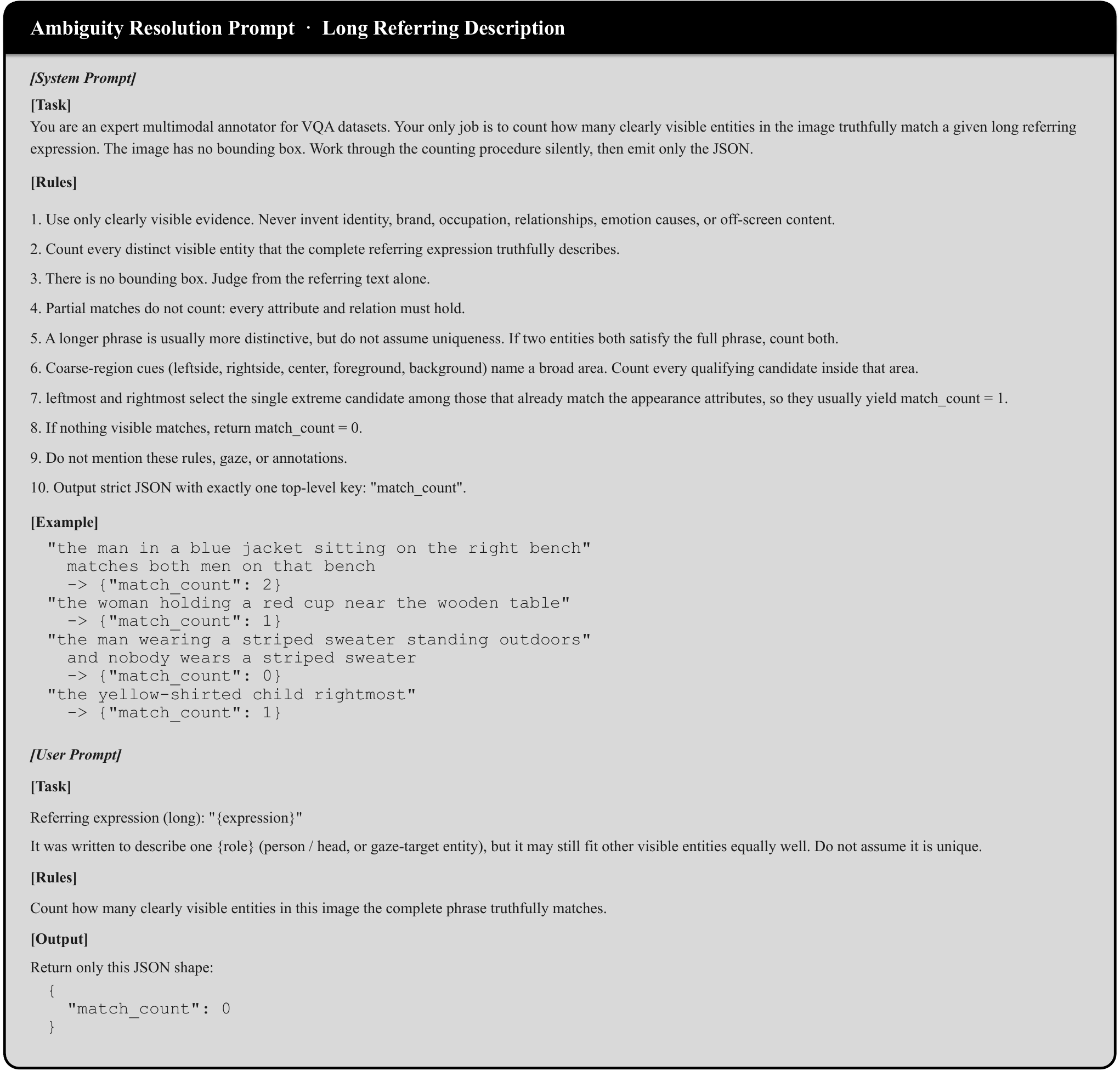}
      \caption{Prompt template for ambiguity resolution of long referring description.}
      \label{fig:prompt_arr_long}
\end{figure}

\begin{figure}[p]
  \centering
      \includegraphics[width=1.0\textwidth]{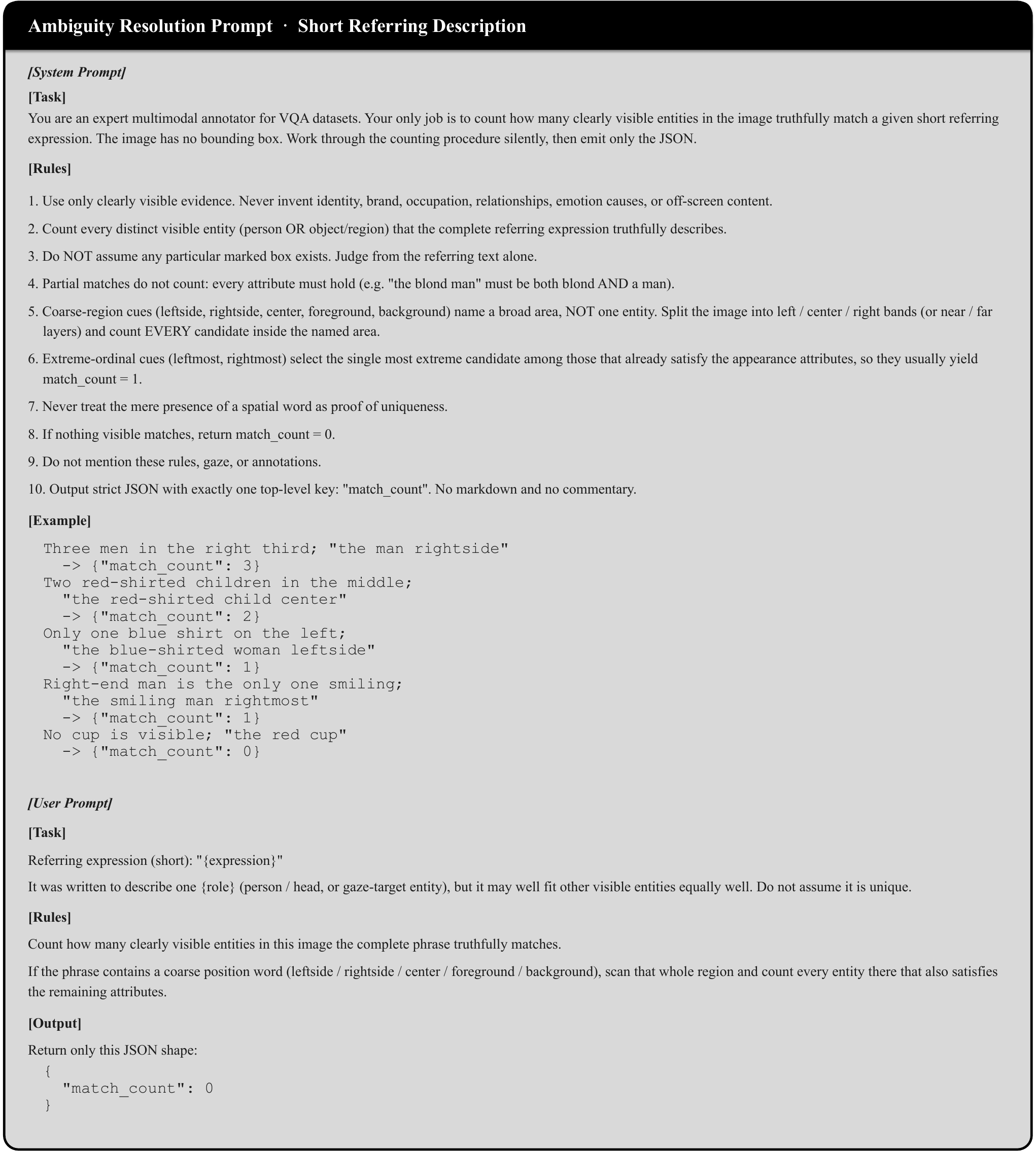}
      \caption{Prompt template for ambiguity resolution of short referring description.}
      \label{fig:prompt_arr_short}
\end{figure}

\paragraph{Ambiguity Check.} To construct ambiguity labels, we further employ Gemini to count the number of entities matching each referring expression in the image, using the prompts detailed in Figure~\ref{fig:prompt_arr_long} and Figure~\ref{fig:prompt_arr_short}. Unique matches (i.e., count of 1) proceed as standard samples, whereas non-unique matches (i.e., count $\ge 2$) are designated for Ambiguous Reference Recognition (ARR) tasks.

\begin{figure}[p]
  \centering
      \includegraphics[width=0.78\textwidth]{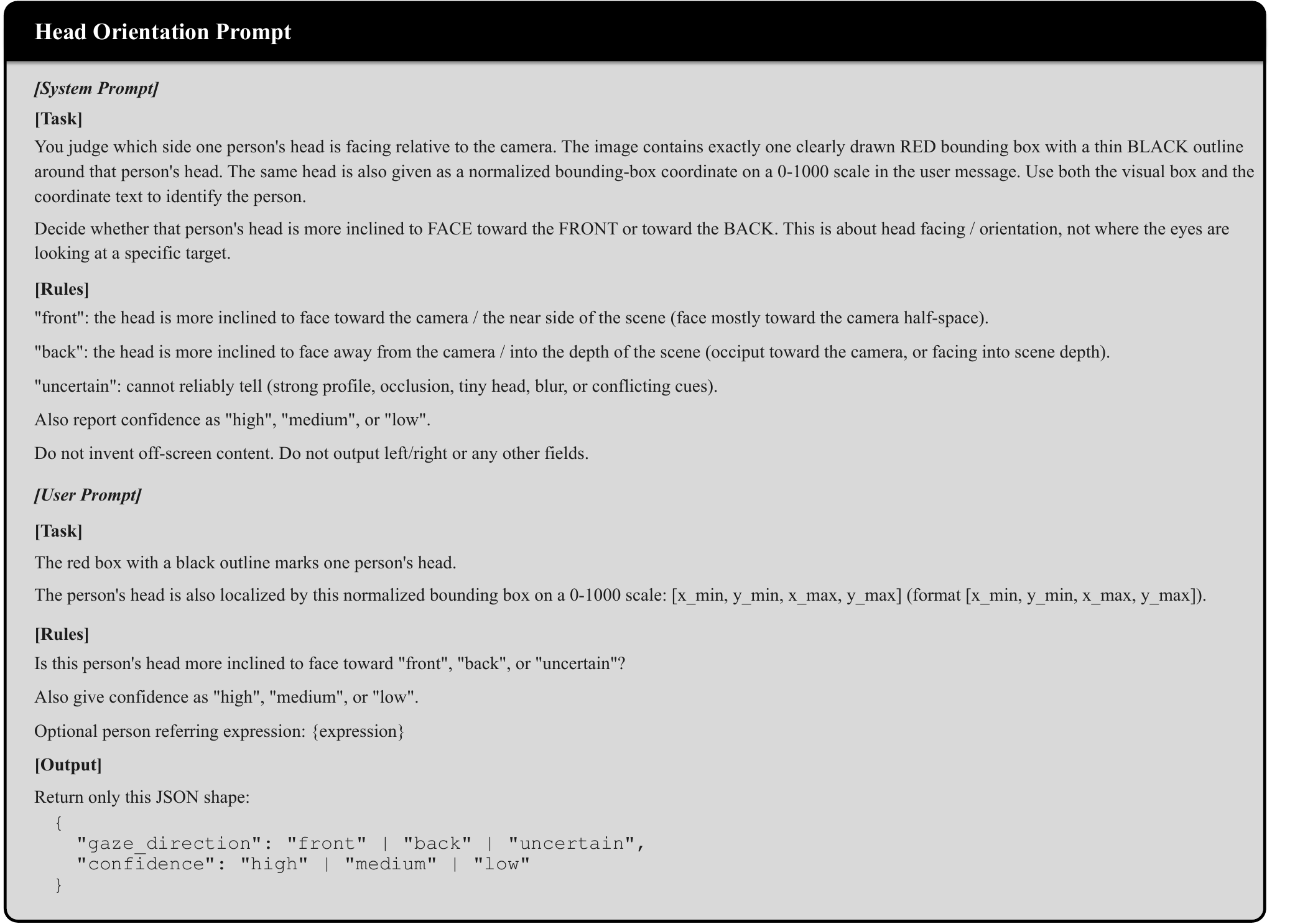}
      \caption{Prompt template for head orientation prediction.}
      \label{fig:prompt_face_orientation}
\end{figure}

\paragraph{Head-Orientation Judgment.} To support gaze depth and direction questions, we prompt Gemini to predict the source person's head orientation relative to the camera following the template in Figure~\ref{fig:prompt_face_orientation}. To ensure consistency, predictions are strictly constrained to \texttt{front}, \texttt{back}, or \texttt{uncertain}.

\paragraph{Relative-Contrast Depth Filter.} We use the DA3 depth maps to compute the relative depth contrast between the source head and gaze target. Let $d_s$ and $d_t$ be the median depth values within a $5\times5$ window around the source and target point, respectively. We define $f$ as the 5th percentile of positive depths in the image (with a floor of $10^{-3}$). The relative contrast is calculated as $r = |d_t - d_s| / \max(\min(|d_t|, |d_s|), f)$. Sample pairs are discarded if $r \le 0.10$; otherwise, they are assigned a depth difference polarity based on the sign of $d_t - d_s$.

\paragraph{Depth Consensus Filtering.} To guarantee high-quality supervision for the Gaze Depth Ordering (GDO) and Gaze Direction Recognition (GDR) tasks, we apply a strict consensus filter. Samples are retained only if the head orientation predicted by Gemini logically aligns with the DA3 depth polarity. For instance, if the head is judged as ``back'' (facing away), the DA3 depth map must indicate the target is farther than the source head ($d_t > d_s$) while passing the contrast threshold. This consensus robustly prevents closer/farther labels from being inverted due to monocular scale ambiguity.

\subsection{\texorpdfstring{Q$\&$A}{Q\&A} Construction}
\label{subsec:app_qa_construction}

We curate task-specific templates to instantiate the unified schema into both openQA and MCQ formats. 

\paragraph{Canonical Templates.} To formulate diverse gaze-centric tasks into a standard visual question answering (VQA) framework, we design canonical templates for each task, as detailed in Table~\ref{tab:app_templates}. These templates utilize semantic referring expressions to naturally ground the gaze sources and targets. For the GTL-B2P task, the template incorporates head bounding box to represent the gaze source person.

\begin{table}[!htbp]
\centering
\caption{\textbf{Canonical HoliGaze question and answer templates.} \texttt{\{person\}} and \texttt{\{target\}} are filled by referring expressions. The GTL-B2P task utilizes the head bounding box to represent the gaze source person.}
\label{tab:app_templates}
\setlength{\tabcolsep}{4pt}
\renewcommand{\arraystretch}{1.10}
\small
\begin{tabular}{p{0.11\textwidth}p{0.48\textwidth}p{0.35\textwidth}}
\toprule
Task & Question templates & Answer grammar \\
\midrule
GTD & What is \{person\} looking at? / What is the gaze target of \{person\}? & \{target\}. \\
GTL-T2P & Where is \{person\} looking at? Output only its normalized point as (x,y), using integer coordinates from 0 to 1000. & (x,y) \\
GTL-T2B & Where is \{person\} looking at? Output only its normalized bounding box as [x1,y1,x2,y2], using integer coordinates from 0 to 1000. & [x1,y1,x2,y2] \\
GTL-B2P & Locate the gaze target of the person with head box [x1,y1,x2,y2]. Output only its normalized point as (x,y), \ldots & (x,y) \\
GSD & Who is looking at \{target\}? Provide all gaze sources that satisfy the condition. & \{persons\}. \\
GSL & Locate the person who is looking at \{target\}. / Find the gaze source for \{target\}. Output only the person's normalized head bounding box as [x1,y1,x2,y2], \ldots If multiple people satisfy the condition, provide all of their head bounding boxes. & [x1,y1,x2,y2]; \ldots \\
GDO & Is the head of \{person\} farther from / closer to the camera than what the person is looking at? & yes / no \\
GDR & From the camera's perspective, relative to the head of \{person\}, is the gaze target left or right in the image, and closer to or farther from the camera? & left-closer / left-farther / right-closer / right-farther \\
GCD & Is \{person\} looking toward the camera lens? / Is \{person\} looking at the camera? & yes / no \\
ARR & Host GTD or GTL question with a non-unique referring expression. & I cannot answer because the referred entity in the question is not unique. \\
PV & Host GTL-T2P question whose target lies outside the frame. & I cannot answer because the gaze target is not visible in the image. \\
\bottomrule
\end{tabular}
\end{table}

\paragraph{OpenQA Generation.} The openQA format encompasses GTL, GTD, GSL, and GSD, alongside the refusal tasks ARR and PV. For standard localization and description tasks, the canonical templates directly prompt the model to generate the corresponding spatial coordinates (points or boxes) or natural language semantic descriptions.

\paragraph{Ambiguity and Presence Verification (Refusal).} To evaluate the model's ability to refuse ill-posed queries, ambiguous expressions identified during the Ambiguity Check are explicitly retained as ARR positives. Similarly, instances where the gaze target is known to be out-of-frame are utilized to formulate Presence Verification (PV) questions. For both ARR and PV, the model is required to output standardized refusal templates (see Table~\ref{tab:app_templates}) rather than hallucinating an arbitrary location or description.

\begin{figure}[p]
  \centering
      \includegraphics[width=0.78\textwidth]{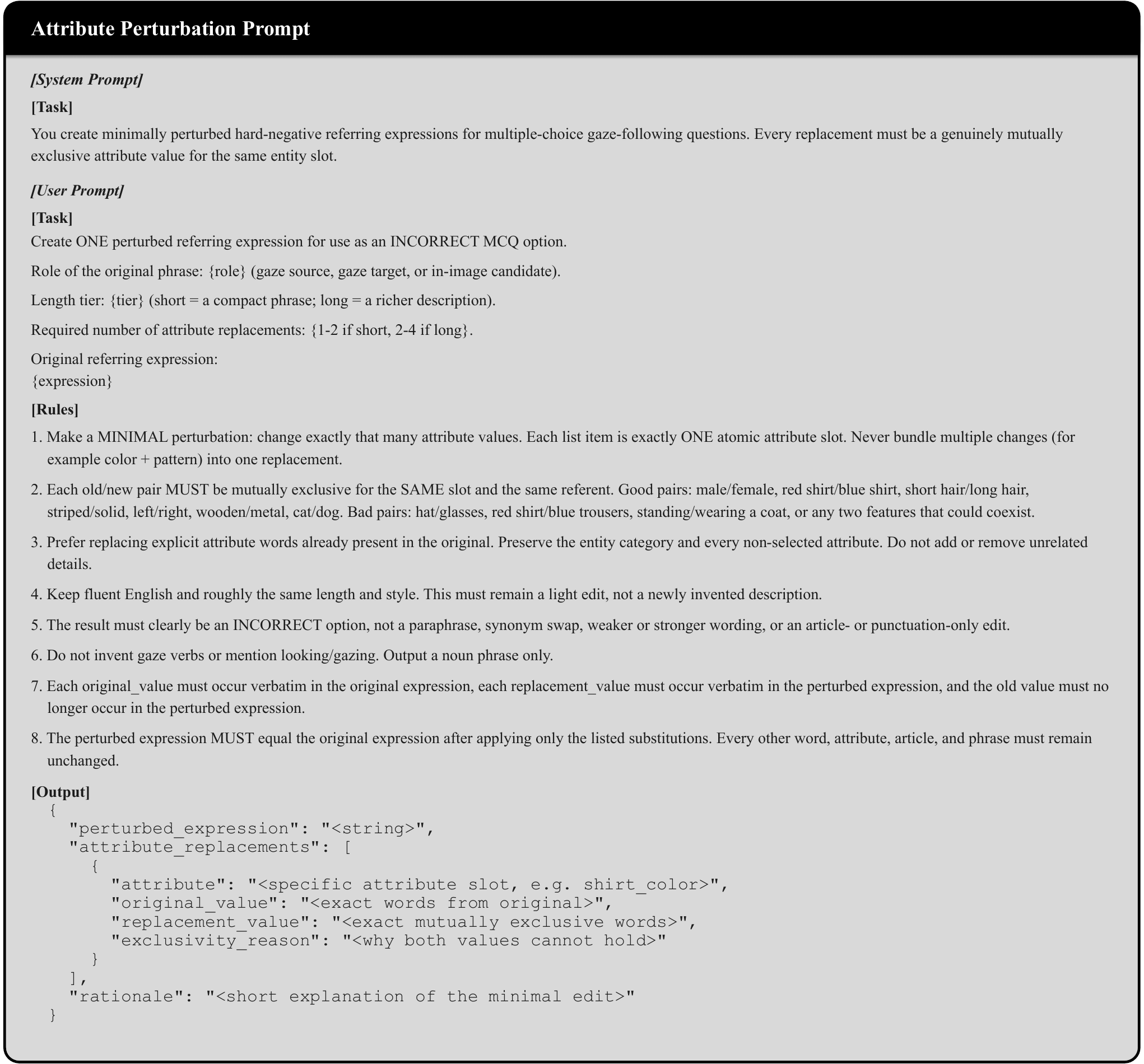}
      \caption{Prompt template for attribute perturbation of referring description.}
      \label{fig:prompt_attribute_perturbation}
\end{figure}

\paragraph{MCQ Distractors.} GTD-MCQ provides 5 single-choice options, each pairing a description with a bounding box; an option is deemed correct only if both elements match. GSD-MCQ provides 6 options and allows multi-select (1 to 3) to naturally accommodate co-gaze scenarios. Distractors for both tasks are drawn from four distinct sources to ensure rigorous evaluation: 
(i)~\textbf{Expansion}: Other in-image entities are proposed by SAM3 and subsequently re-described by Gemini~\citep{comanici2025gemini}. 
(ii)~\textbf{Attribute Perturbation}: Qwen3-32B~\citep{yang2025qwen3} rewrites the gold description into hard negatives by altering 1 to 4 specific attributes (e.g., changing ``red shirt'' to ``blue shirt'') while preserving the rest of the sentence, utilizing the prompt illustrated in Figure~\ref{fig:prompt_attribute_perturbation}. 
(iii)~\textbf{Permutation}: We systematically swap bounding boxes across options, pairing a correct description with a wrong location (e.g., a translated copy of the ground-truth box or another entity's box), and vice versa. 
(iv)~\textbf{Refusal Templates}: ARR refusal templates are included as distractors to test whether the model incorrectly rejects a perfectly valid and answerable query.

\paragraph{Data Splits.} To strictly prevent data leakage, we adhere to each source dataset's native split to curate \textbf{HoliGaze-Corpus} and \textbf{HoliGaze-Bench}. The former is sampled exclusively from the training sets, while the latter is drawn from the test sets, except for Ego4D-Social, which uses its official validation split.

\subsection{Human Verification}
\label{subsec:app_human_verification}
To establish a rigorous evaluation standard, the HoliGaze-Bench undergoes comprehensive instance-level human verification. We engage five trained annotators to cross-check every sample in the benchmark. To ensure objectivity, the annotators first evaluate the same samples independently; any inconsistencies in their assessments are then resolved through centralized discussion to reach a consensus.

During this review, the annotators meticulously verify seven dimensions of correctness: (1) the fundamental validity of the gaze relationship (i.e., whether the source person is genuinely looking at the designated target); (2) the accuracy and visual faithfulness of the referring descriptions; (3) the strict uniqueness of the references within the image; (4) the logical validity and mutual exclusivity of the MCQ distractors; (5) the spatial consistency and tightness of the annotated points and bounding boxes; (6) the correctness of the depth polarity for GDO and GDR tasks; and (7) the accurate identification of out-of-frame targets for PV tasks.

Any sample failing these checks is either manually corrected or discarded. Ultimately, the annotators modified 56 out of the 8,000 instances (a modification rate of $0.70\%$). This low intervention rate indicates that our automated data engine already enforces robust quality control, while the subsequent human review further guarantees a reliable evaluation benchmark.

\section{Details of Out-of-Distribution Evaluation}
\label{sec:appendix_ood}
In this section, we detail the out-of-distribution (OOD) evaluation settings introduced in Section~4.5 of the main text.

\subsection{Evaluation on Unseen Datasets}
To assess the robust generalization of gaze-following models, we evaluate them on two OOD datasets, introduced respectively as follows:
\begin{itemize}
    \item \textbf{GFIE}~\citep{hu2023gfie}: This dataset features diverse and complex indoor daily life scenarios, capturing natural human activities and gaze behaviors in egocentric environments.
    \item \textbf{4D-OR}~\citep{chen2026they}: Tailored for clinical scene analysis, this dataset captures surgical activities in authentic operating rooms. It presents highly challenging, crowded, and multi-person interactive environments.
\end{itemize}

\begin{figure}[!ht]
\centering
    \includegraphics[width=1.0\textwidth]{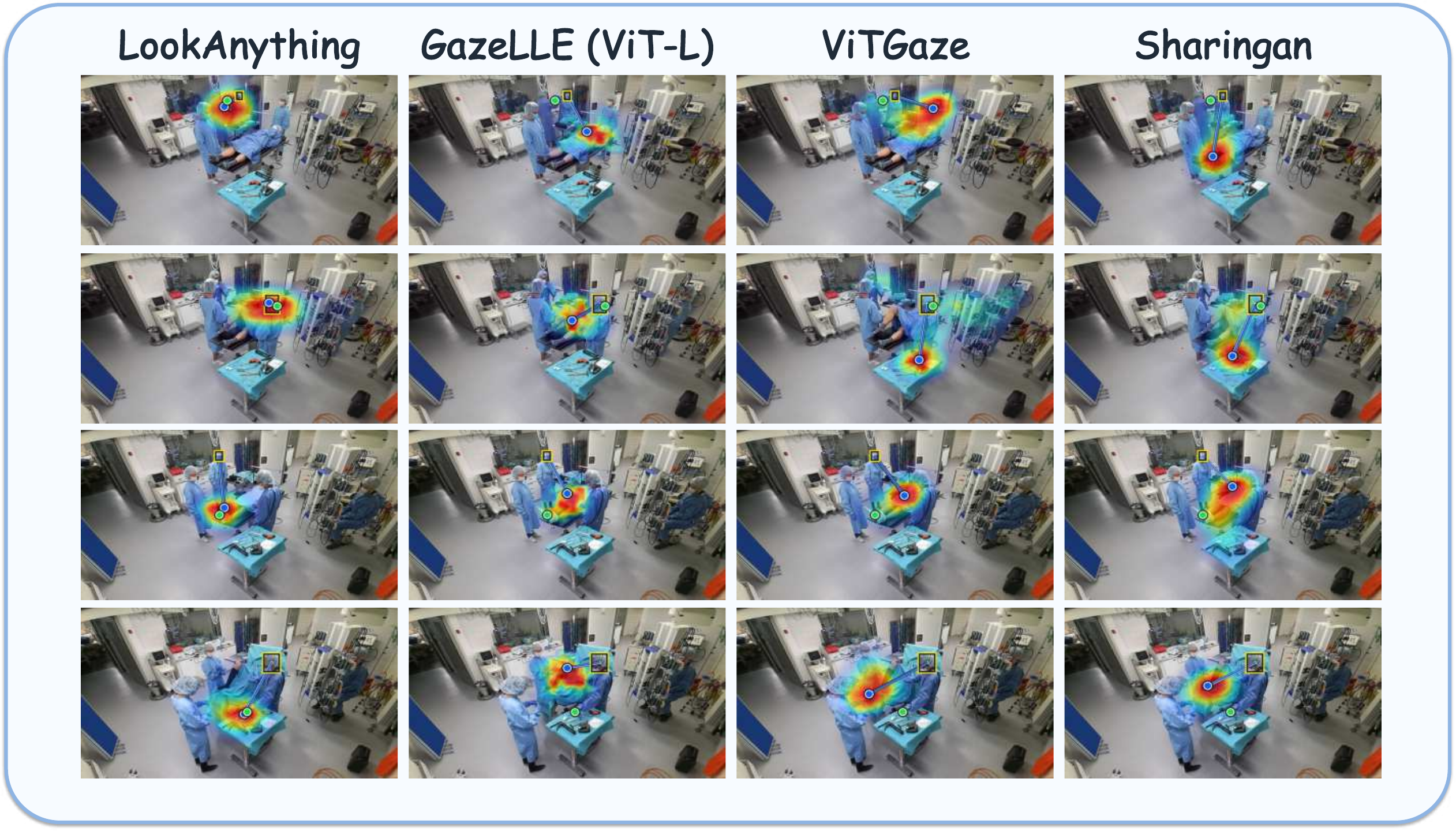}
    \caption{\textbf{Qualitative OOD evaluation results on 4D-OR.} The predicted gaze target points are marked in blue, and the ground-truth  points are marked in green.}
    \label{fig:qual_ood_4dor}
\end{figure}

\begin{figure}[!ht]
\centering
    \includegraphics[width=1.0\textwidth]{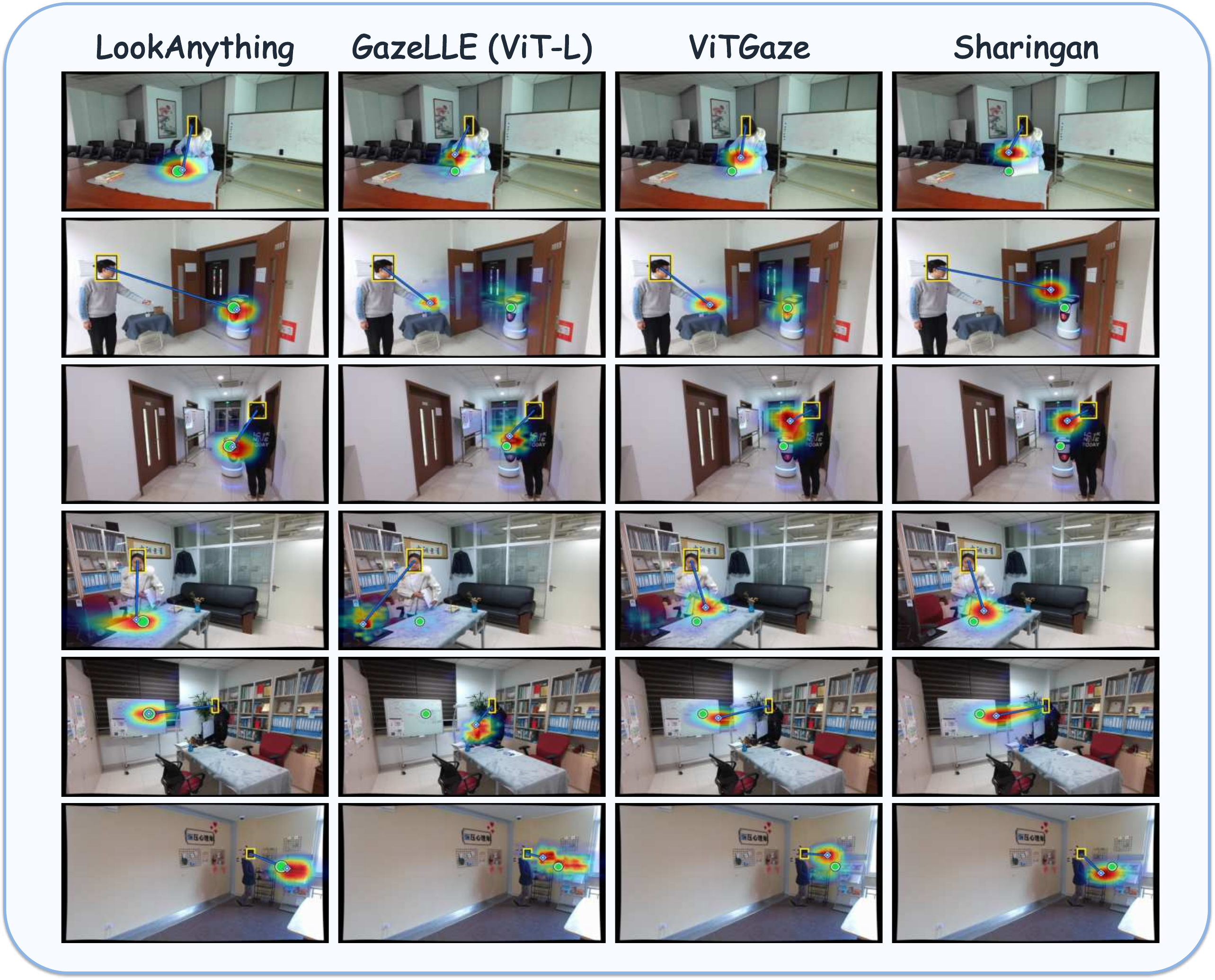}
    \caption{\textbf{Qualitative OOD evaluation results on GFIE.} The predicted gaze target points are marked in blue, and the ground-truth points are marked in green.}
    \label{fig:qual_ood_gfie}
\end{figure}

Crucially, the image domains and scene layouts of both GFIE and 4D-OR remain strictly unseen during the training of LookAnything, Gaze-LLE, and other baseline models. This strict domain separation ensures a rigorous evaluation of cross-domain transferability. In addition to the quantitative results reported in the main text, we provide qualitative visualizations on these unseen datasets in Figures~\ref{fig:qual_ood_4dor} and \ref{fig:qual_ood_gfie}. LookAnything accurately infers the gaze targets in these completely unfamiliar scenes, demonstrating strong zero-shot generalization.

\subsection{Real-World Robotic Deployment}
To validate LookAnything's applicability in dynamic physical environments, we deploy the model on a Unitree G1 robot across diverse everyday scenarios, including shopping aisles, storage shelves, and desktop workspaces.

The deployment involved a total of four subjects. All subjects, as well as any bystanders who incidentally entered the camera frame during recording, provided signed informed consent.

We designed two camera perspective settings to evaluate the model's robustness:
\begin{itemize}
    \item \textbf{Static Perspective:} The robot remains stationary while subjects move and interact freely within the camera's fixed field of view.
    \item \textbf{Dynamic Following:} A human operator teleoperates the robot to physically follow the moving subjects. This setting specifically challenges LookAnything's ability to maintain continuous and stable gaze following despite shifting camera perspectives and ego-motion.
\end{itemize}

For comprehensive qualitative results and continuous tracking demonstrations, please refer to the supplementary real-world demo video.

\section{Additional Experimental Results}
\label{sec:appendix_results}

\subsection{Effect of Multi-Task Joint Training}
\label{subsec:app_multitask_ablation}
To investigate the impact of multi-task joint training on the conventional gaze target detection capability, we conduct an ablation study where LookAnything is trained exclusively using conventional gaze target detection samples (i.e., the GTL-B2P task format). 

\begin{table}[!htbp]
\centering
\caption{\textbf{Ablation study on multi-task joint training.} ``GTL-B2P Only'' trains LookAnything exclusively on conventional gaze target detection, while ``Multi-Task'' denotes joint training on diverse gaze-centric tasks of HoliGaze-Corpus.}
\label{tab:ablation_multitask}
\setlength{\tabcolsep}{4.5pt}
\renewcommand{\arraystretch}{1.08}
\small
\resizebox{\textwidth}{!}{%
\begin{tabular}{l cc cc ccc cc cc cc}
\toprule
\multirow{2}{*}{\textbf{Setting}} & \multicolumn{2}{c}{\textbf{Avg.}} & \multicolumn{2}{c}{\textbf{VAT}} & \multicolumn{3}{c}{\textbf{GazeFollow}} & \multicolumn{2}{c}{\textbf{ChildPlay}} & \multicolumn{2}{c}{\textbf{GazeHOI}} & \multicolumn{2}{c}{\textbf{GOO-Real}} \\
\cmidrule(lr){2-3} \cmidrule(lr){4-5} \cmidrule(lr){6-8} \cmidrule(lr){9-10} \cmidrule(lr){11-12} \cmidrule(lr){13-14}
& L2$\downarrow$ & AUC$\uparrow$ & L2$\downarrow$ & AUC$\uparrow$ & L2$\downarrow$ & Min L2$\downarrow$ & AUC$\uparrow$ & L2$\downarrow$ & AUC$\uparrow$ & L2$\downarrow$ & AUC$\uparrow$ & L2$\downarrow$ & AUC$\uparrow$ \\
\midrule
GTL-B2P Only & 0.111 & 0.939 & 0.100 & 0.937 & 0.092 & 0.051 & \textbf{0.945} & \textbf{0.093} & 0.954 & 0.134 & 0.937 & 0.138 & 0.924 \\
\rowcolor{cvprblue!15} Multi-Task (Final) & \textbf{0.109} & \textbf{0.942} & \textbf{0.098} & \textbf{0.941} & \textbf{0.091} & \textbf{0.048} & \textbf{0.945} & 0.094 & \textbf{0.956} & \textbf{0.130} & \textbf{0.940} & \textbf{0.132} & \textbf{0.927} \\
\bottomrule
\end{tabular}%
}
\end{table}

As shown in Table~\ref{tab:ablation_multitask}, the model trained solely on the single GTL-B2P task already achieves strong performance across the five native benchmarks. However, when trained jointly across all holistic dimensions (our final multi-task setting), LookAnything yields further performance improvements in almost all metrics. Specifically, multi-task training reduces the $L_2$ distance and increases the AUC score across the board, with notable gains on challenging datasets like GazeHOI and GOO-Real.

This consistent improvement confirms that our task unification does not compromise, but rather enhances, fundamental localization accuracy. By jointly optimizing diverse tasks, such as gaze target description, source localization, and depth reasoning, the model acquires a more comprehensive understanding of human gaze behaviors and scene contexts.

\subsection{Blind Baseline for MCQ Tasks}
\label{subsec:app_blind_baseline}
To investigate the potential influence of dataset biases and spurious language shortcuts in the multiple-choice question (MCQ) tasks, we train a blind baseline on the HoliGaze-Corpus. This blind model shares the exact architecture as LookAnything but receives only the textual questions and options during training, with the visual image input completely masked. During evaluation, the images are provided normally. This setup rigorously tests whether the model can artificially inflate its performance by exploiting statistical shortcuts in the textual training distribution, thereby validating the rationality of our benchmark design.

\begin{wraptable}[8]{r}{0.52\textwidth}
\vspace{-6pt}
\centering
\caption{\textbf{Ablation study using a blind baseline on MCQ tasks.} The blind model uses the same architecture as LookAnything but is trained with no visual inputs.}
\label{tab:blind_baseline}
\setlength{\tabcolsep}{4pt}
\renewcommand{\arraystretch}{1.08}
\footnotesize
\resizebox{\linewidth}{!}{%
\begin{tabular}{l ccccc}
\toprule
\textbf{Model} & \textbf{GTD} & \textbf{GSD} & \textbf{GDO} & \textbf{GDR} & \textbf{GCD} \\
\midrule
Qwen3-VL-2B & 0.394 & 0.483 & 0.550 & 0.237 & 0.660 \\
Blind Baseline & 0.406 & 0.505 & 0.494 & 0.269 & 0.593 \\
\rowcolor{cvprblue!15} LookAnything & \textbf{0.764} & \textbf{0.865} & \textbf{0.924} & \textbf{0.863} & \textbf{0.847} \\
\bottomrule
\end{tabular}%
}
\vspace{-4pt}
\end{wraptable}
As reported in Table~\ref{tab:blind_baseline}, the blind baseline does not appear to exploit trivial shortcuts. On spatial reasoning tasks, it achieves performance roughly equivalent to random chance, scoring 0.494 on the binary GDO task and 0.269 on the four-way GDR task. For description-oriented tasks (GTD and GSD), its performance (0.406 and 0.505, respectively) closely mirrors that of the zero-shot base model, Qwen3-VL-2B (0.394 and 0.483). This indicates that mere exposure to the text-only training data provides no unfair advantage. In contrast, the full LookAnything model which is trained jointly with comprehensive visual grounding exhibits massive performance gains (e.g., boosting GDR to 0.863 and GDO to 0.924).

These results confirm that the HoliGaze dataset is free from trivial linguistic shortcuts. Accurate holistic gaze following strictly requires deep visual-spatial reasoning rather than language guessing, verifying the high quality and rationality of the constructed MCQ samples.

\subsection{Scaling Analysis}

As a foundation model for holistic gaze following, LookAnything is expected to benefit from larger training data volumes. To examine this, we perform a scaling analysis by evaluating model checkpoints at different pretraining stages, mapped to the absolute volume of training data seen by the model (ranging from 38K to 227K samples).

\begin{figure}[!ht]
\centering
    \includegraphics[width=1.0\textwidth]{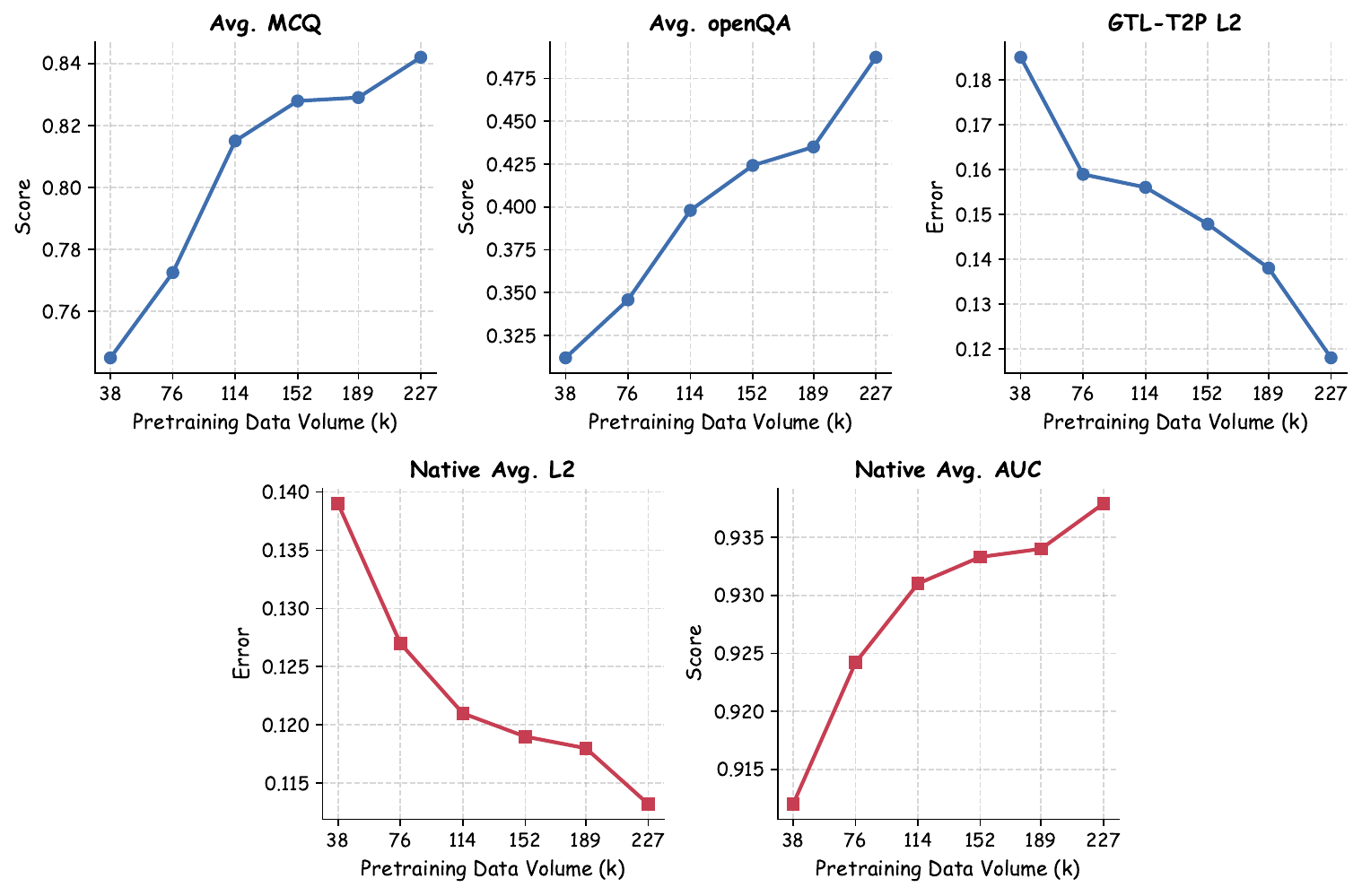}
    \caption{\textbf{Effect of pretraining data volume on LookAnything's performance.} The top row illustrates the holistic gaze following metrics on HoliGaze-Bench, while the bottom row demonstrates the average performance across the 5 native gaze target detection benchmarks.}
    \label{fig:scaling_law}
    \vspace{-2mm}
\end{figure}

As shown in Figure~\ref{fig:scaling_law}, model performance improves consistently with increased training data volume across both the holistic capabilities and conventional tasks. On HoliGaze-Bench (top row), the multiple-choice accuracy (Avg. MCQ), generation quality for open-ended queries (Avg. openQA), and spatial localization error (GTL-T2P $L_2$) all show steady gains as the data size scales up. A similar trend is observed on the conventional native benchmarks (bottom row), where the average $L_2$ error continuously decreases and the Area Under Curve (Avg. AUC) progressively rises. 

Notably, within the observed data regime, the performance trajectories maintain clear, unbroken momentum without reaching a plateau. These results strongly support our core motivation of building a foundation model for holistic gaze following. They demonstrate that scaling up diverse, heterogeneous gaze data fundamentally enhances the model's representational capacity and downstream generalizability, while also suggesting ample room for further gains with even larger data volumes.

\subsection{Inference Profiling}
\label{subsec:app_inference_profiling}
We profile the inference efficiency of LookAnything under different output formats and generation pathways. A key advantage of our architecture is its flexibility. The model can selectively output spatial predictions autoregressively via the language backbone or directly regress them in a single forward pass via the lightweight auxiliary branches. Furthermore, it supports diverse output formats including points, bounding boxes, and heatmaps, allowing it to seamlessly adapt to varying downstream application requirements.

To quantify this efficiency, we benchmark the inference latency, computational overhead (FLOPs), and peak GPU memory consumption across different settings. All profiling tests are conducted on a single NVIDIA RTX 5090 GPU using BF16 precision. For the language generation pathway, KV cache is enabled. We run 10 independent inference iterations for each setting and report the median metrics.

Table~\ref{tab:inference_profiling} summarizes the profiling results. The auxiliary branches offer significant efficiency advantages, producing spatial coordinates in a single forward pass with ultra-low latency (e.g., 111.30 ms for bounding boxes) and reduced computational overhead. This makes them highly suited for real-time embodied control. Conversely, while the autoregressive language branch inherently requires more time (e.g., 439.56 ms for bounding boxes), it provides the critical advantages of fine-grained spatial reasoning, unified textual formatting, and contextual comprehension. Together, these complementary pathways empower LookAnything with versatile deployment capabilities tailored to either high-speed execution or comprehensive understanding.

\begin{table}[!htbp]
\centering
\caption{\textbf{Inference profiling of LookAnything.} We compare the efficiency of generating spatial predictions via the auxiliary branches versus the language backbone. Metrics are the medians over 10 runs on a single RTX 5090 (BF16).}
\label{tab:inference_profiling}
\setlength{\tabcolsep}{8pt}
\renewcommand{\arraystretch}{1.10}
\small
\begin{tabular}{ll ccc}
\toprule
\textbf{Generation Pathway} & \textbf{Output Format} & \textbf{Latency (ms)} & \textbf{FLOPs (T)} & \textbf{Peak Memory (MiB)} \\
\midrule
\multirow{3}{*}{Auxiliary Branch} & Heatmap & 117.58 & 1.811 & 4329.6 \\
& Point & 116.88 & 1.805 & 4329.6 \\
& Bounding Box & 111.30 & 1.824 & 4329.6 \\
\midrule
\multirow{2}{*}{Language Branch} & Point & 295.73 & 1.836 & 4344.2 \\
& Bounding Box & 439.56 & 1.881 & 4345.2 \\
\bottomrule
\end{tabular}
\end{table}

\subsection{Visualization of Auxiliary Predictions}

\begin{figure}[!ht]
  \centering
      \includegraphics[width=1.0\textwidth]{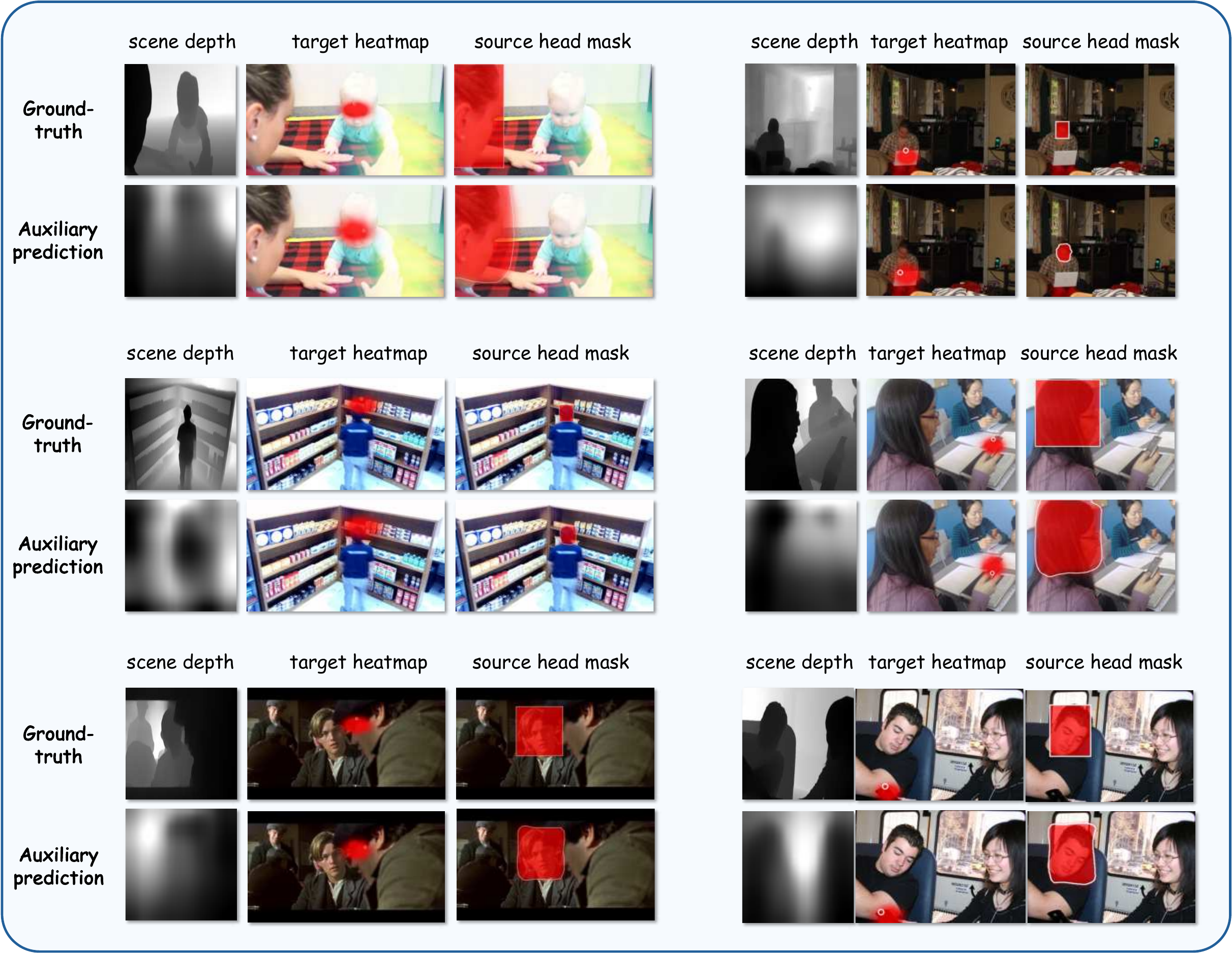}
      \caption{Visualizations of auxiliary predictions generated by the structurally disentangled queries.}
      \label{fig:aux_pred_vis}
      \vspace{-2mm}
\end{figure}

As shown in Figure~\ref{fig:aux_pred_vis}, we visualize the auxiliary predictions generated by the structurally disentangled queries in LookAnything, including the scene depth maps, the source head masks, and the gaze target spatial heatmaps. Although the auxiliary decoders adopt a lightweight MAE-style architecture with patch-level reconstruction, LookAnything is able to reconstruct semantically meaningful geometric and physical representations of the entire scene. 

The effectiveness of these predictions highlights the value of our auxiliary supervision design. Empowered by the large-scale multi-task pretraining on the extensive HoliGaze-Corpus, the gaze-prior queries are driven to implicitly integrate global visual context, such as background structures, spatial layout, and object appearances, with localized gaze behaviors. As a result, the model yields coherent and accurate predictions for the critical physical priors, successfully internalizing them without relying on explicit external inputs like cropped head images or pre-computed depth maps.

Although the predicted depth maps are relatively coarse due to the inherent patch-level resolution of the auxiliary decoders~\citep{he2022masked}, the source head masks remain remarkably accurate and well-localized. Together, they provide essential geometric guidance for downstream tasks. In particular, the language backbone fundamentally benefits from anticipating the scene depth and grounding the human head, which helps resolve spatial ambiguities, refine precise target localization, and improve overall holistic gaze understanding.

\subsection{Qualitative Visualization Results}

\begin{figure}[p]
\centering
    \includegraphics[width=1.0\textwidth]{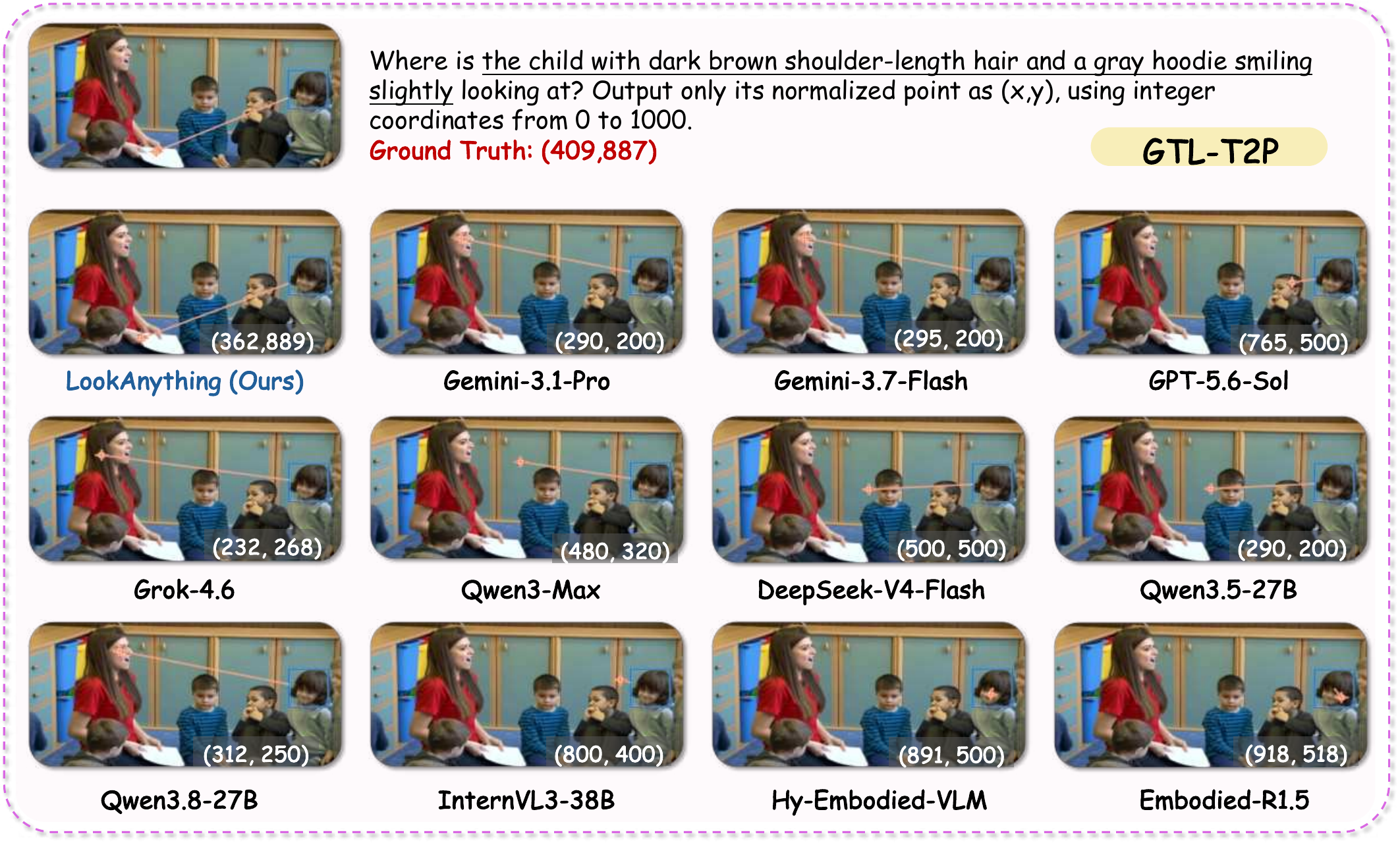}
    \caption{Qualitative Cases of GTL-T2P.}
    \label{fig:qual_gtlt2p_1}
\end{figure}

\begin{figure}[p]
\centering
  \includegraphics[width=1.0\textwidth]{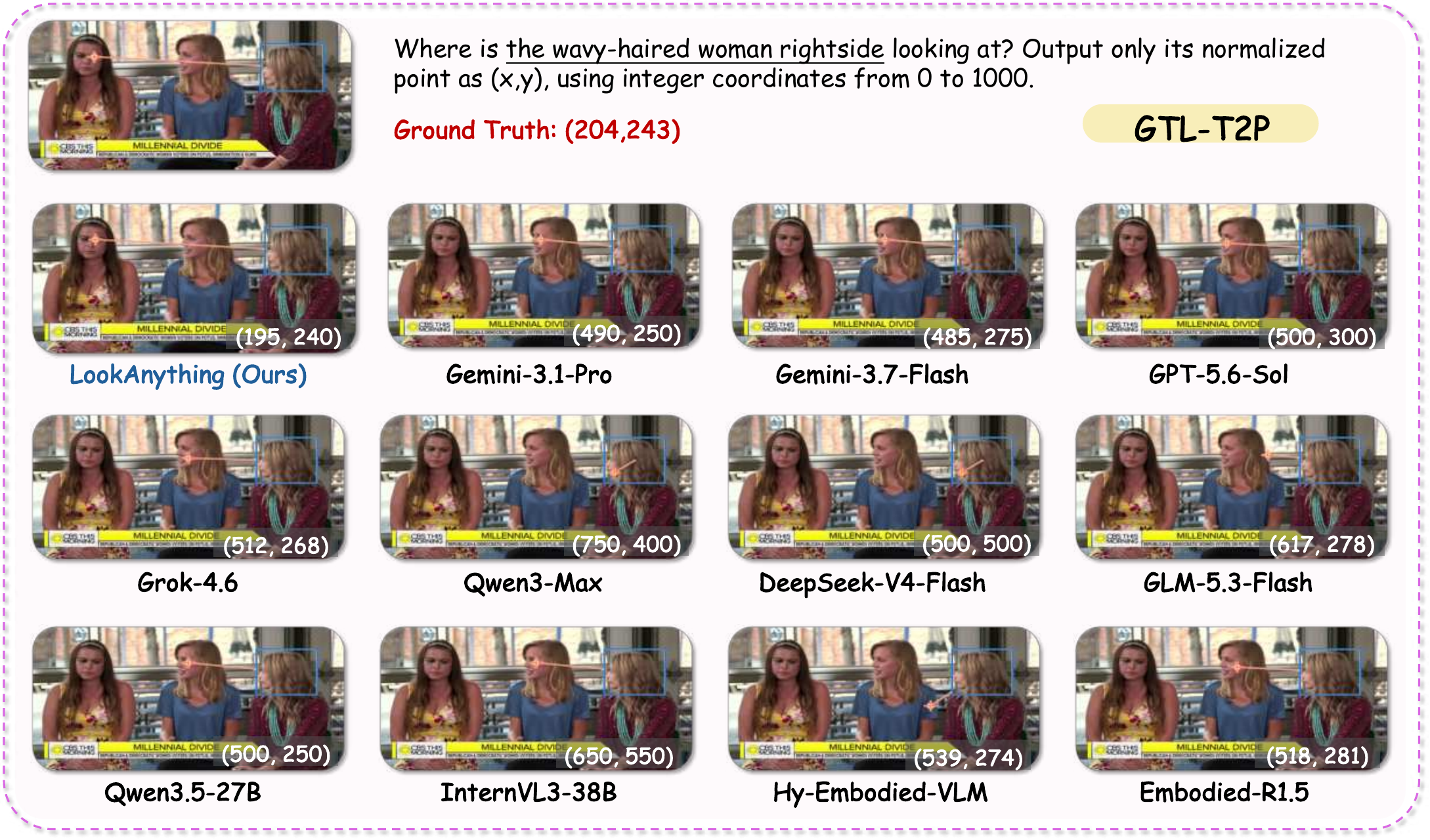}
  \caption{Qualitative Cases of GTL-T2P.}
  \label{fig:qual_gtlt2p_2}
\end{figure}

\begin{figure}[p]
\centering
  \includegraphics[width=0.9\textwidth]{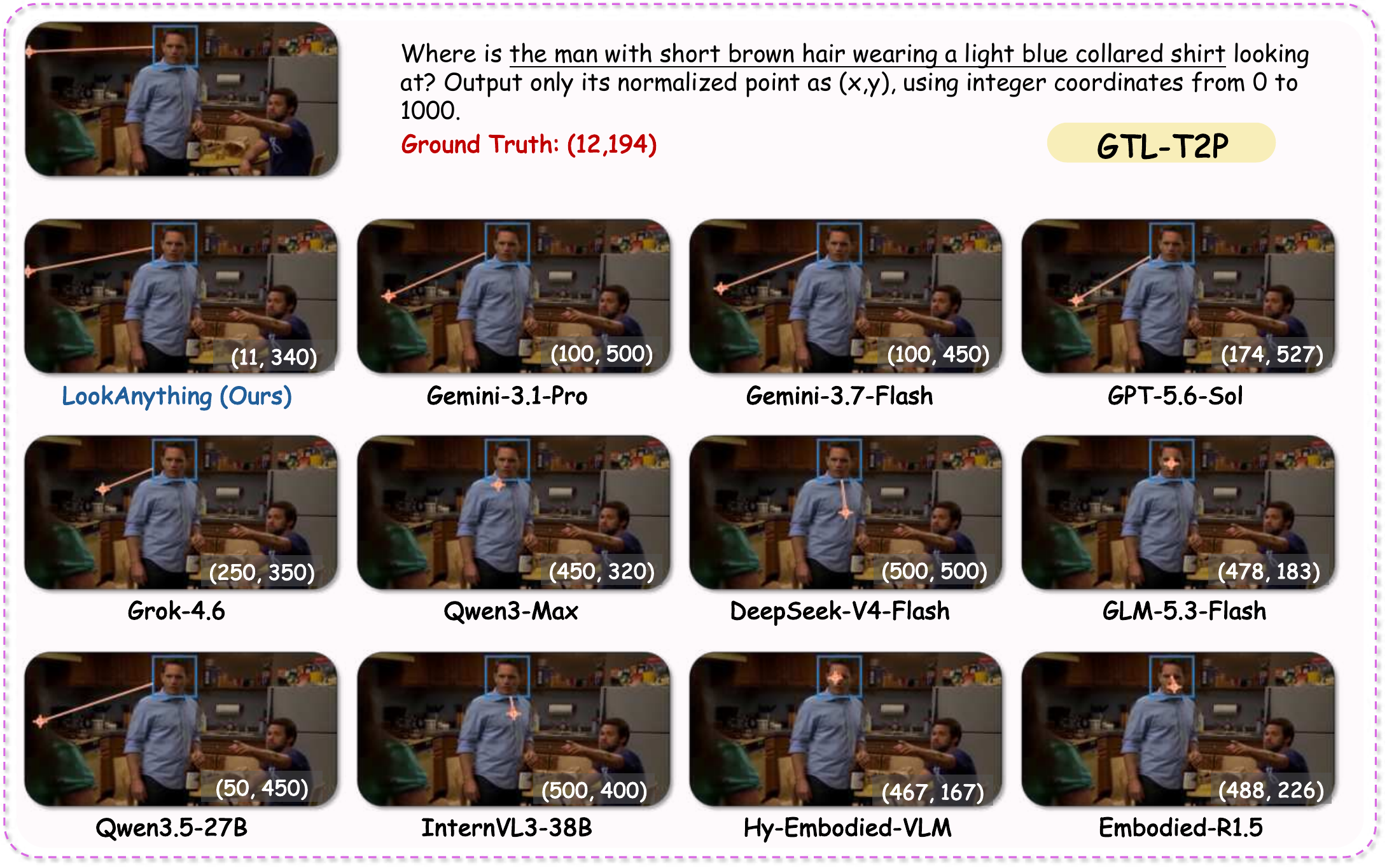}
  \caption{Qualitative Cases of GTL-T2P.}
  \label{fig:qual_gtlt2p_3}
\end{figure}

\begin{figure}[p]
\centering
    \includegraphics[width=0.9\textwidth]{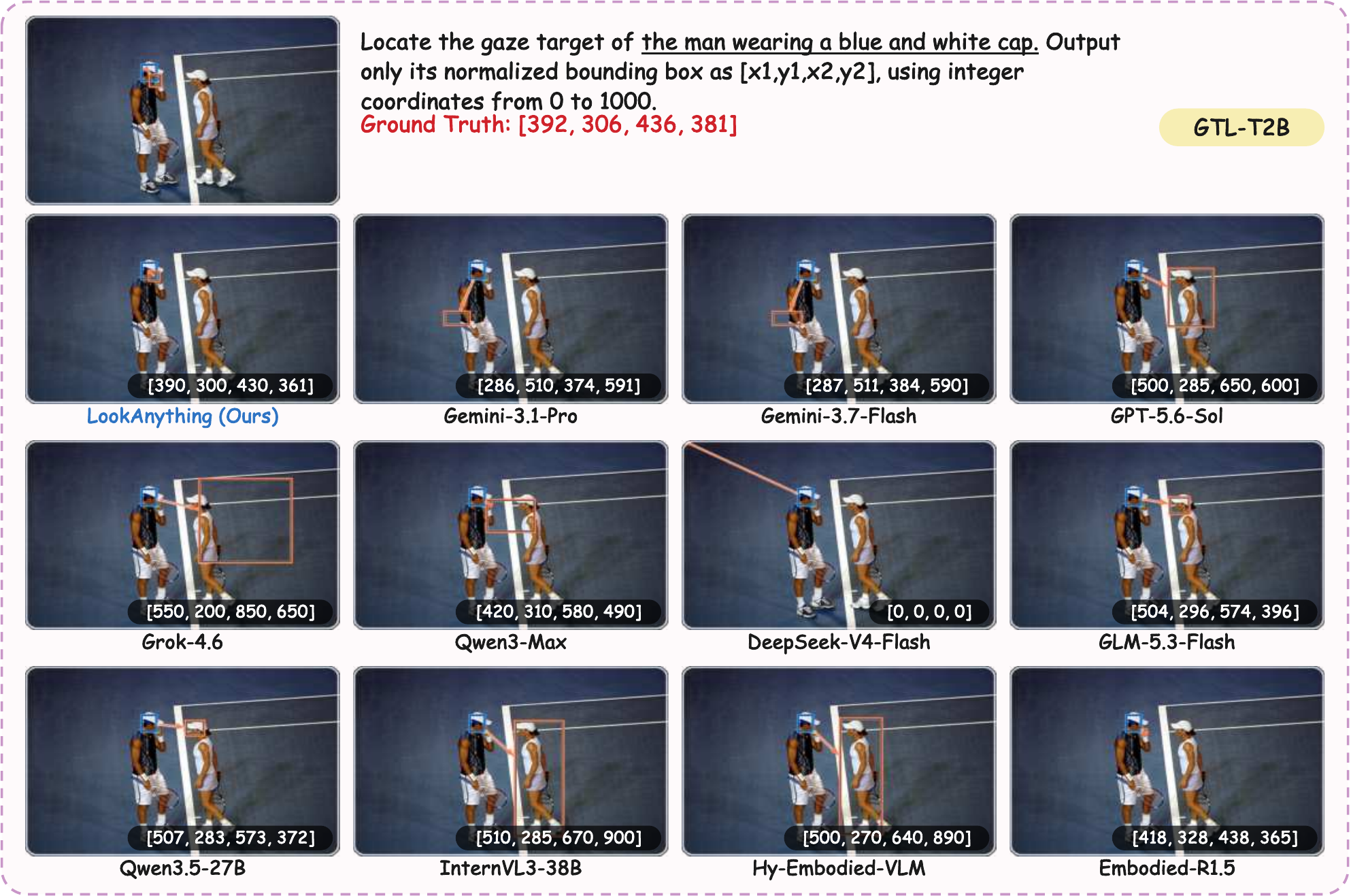}
    \caption{Qualitative Cases of GTL-T2B.}
    \label{fig:qual_gtlt2b_1}
\end{figure}

\begin{figure}[p]
\centering
    \includegraphics[width=1.0\textwidth]{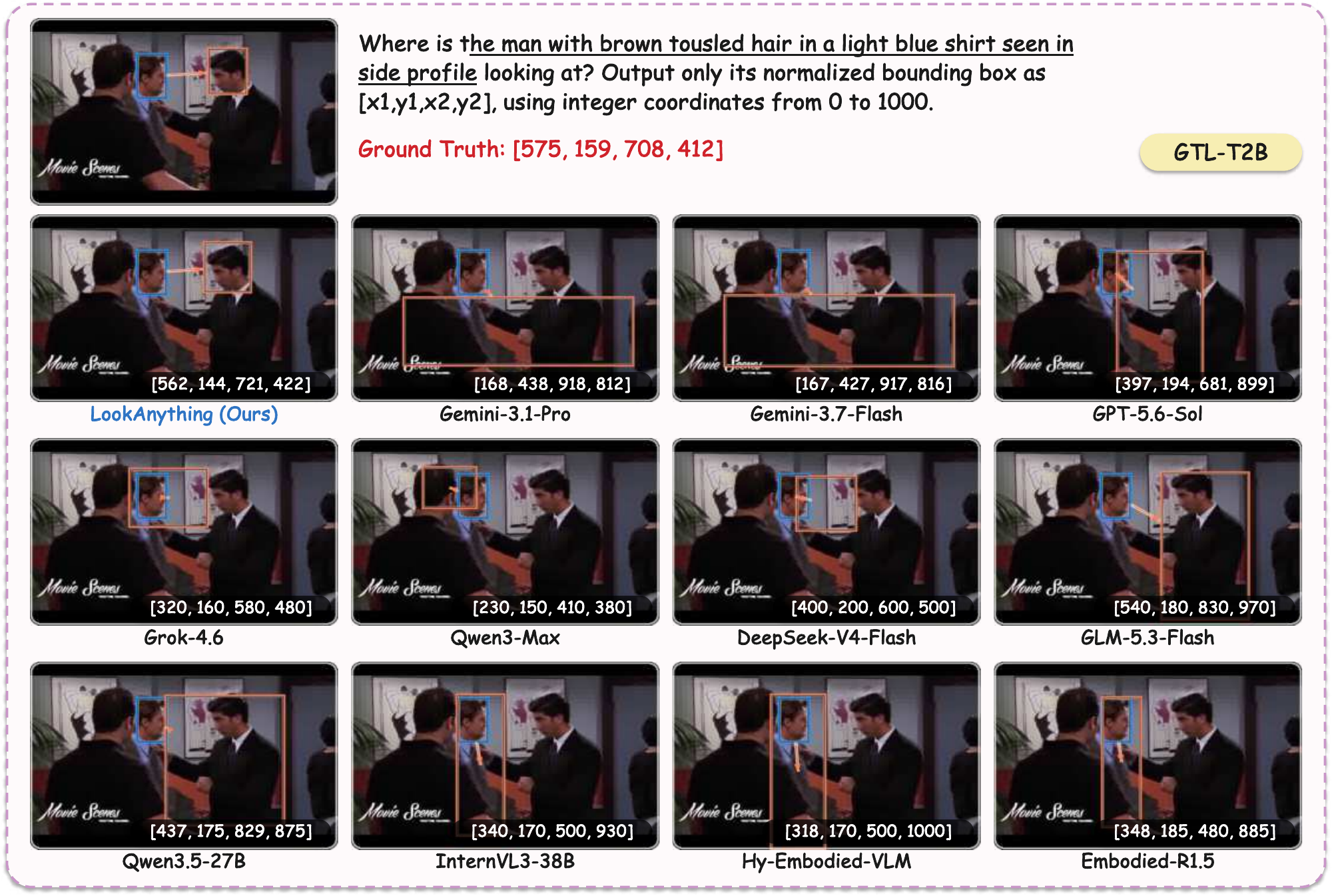}
    \caption{Qualitative Cases of GTL-T2B.}
    \label{fig:qual_gtlt2b_2}
\end{figure}

\begin{figure}[p]
  \centering
      \includegraphics[width=1.0\textwidth]{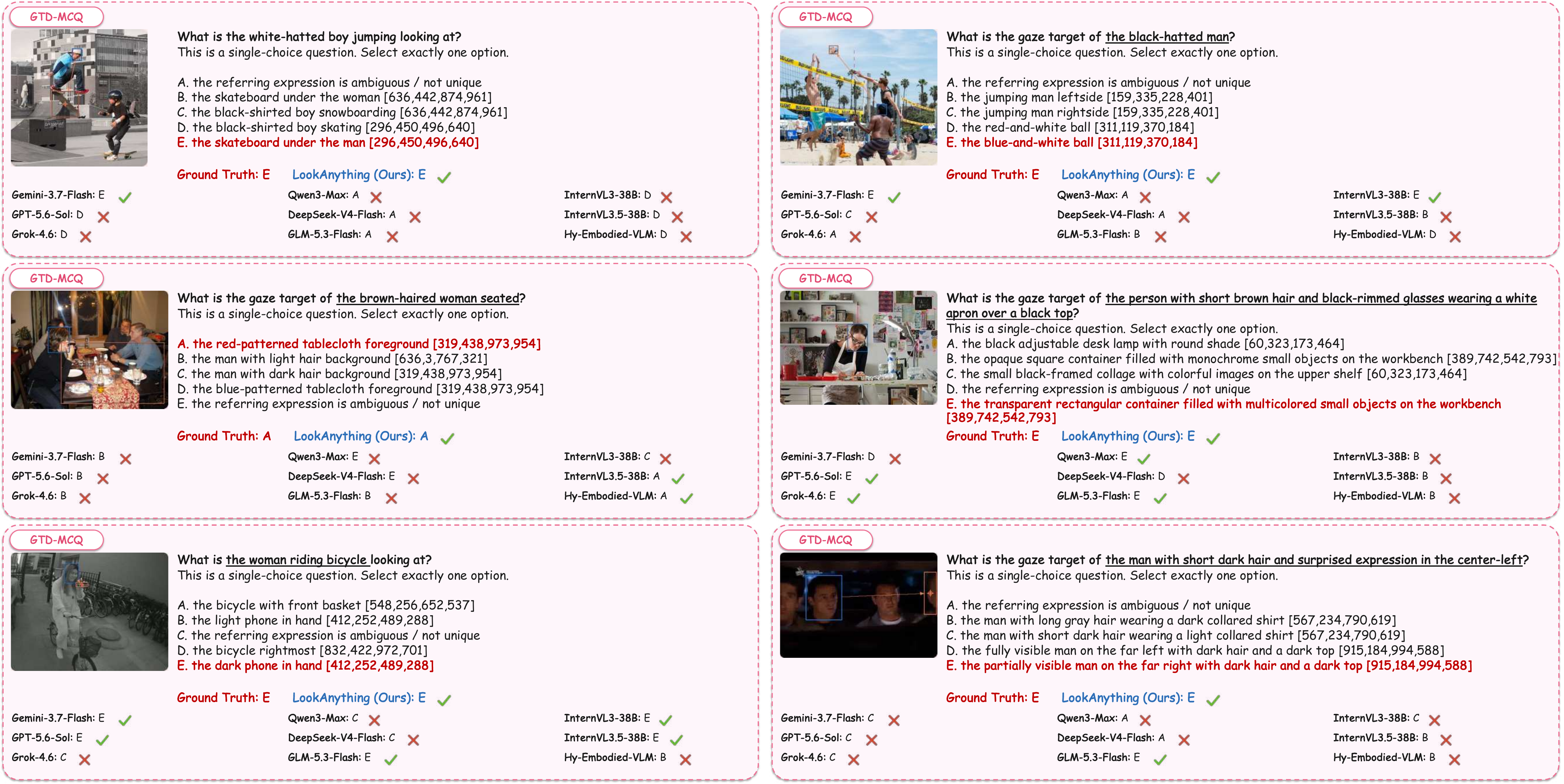}
      \caption{Qualitative Cases of GTD-MCQ.}
      \label{fig:qual_gtd_1}
\end{figure}

\begin{figure}[p]
  \centering
      \includegraphics[width=0.9\textwidth]{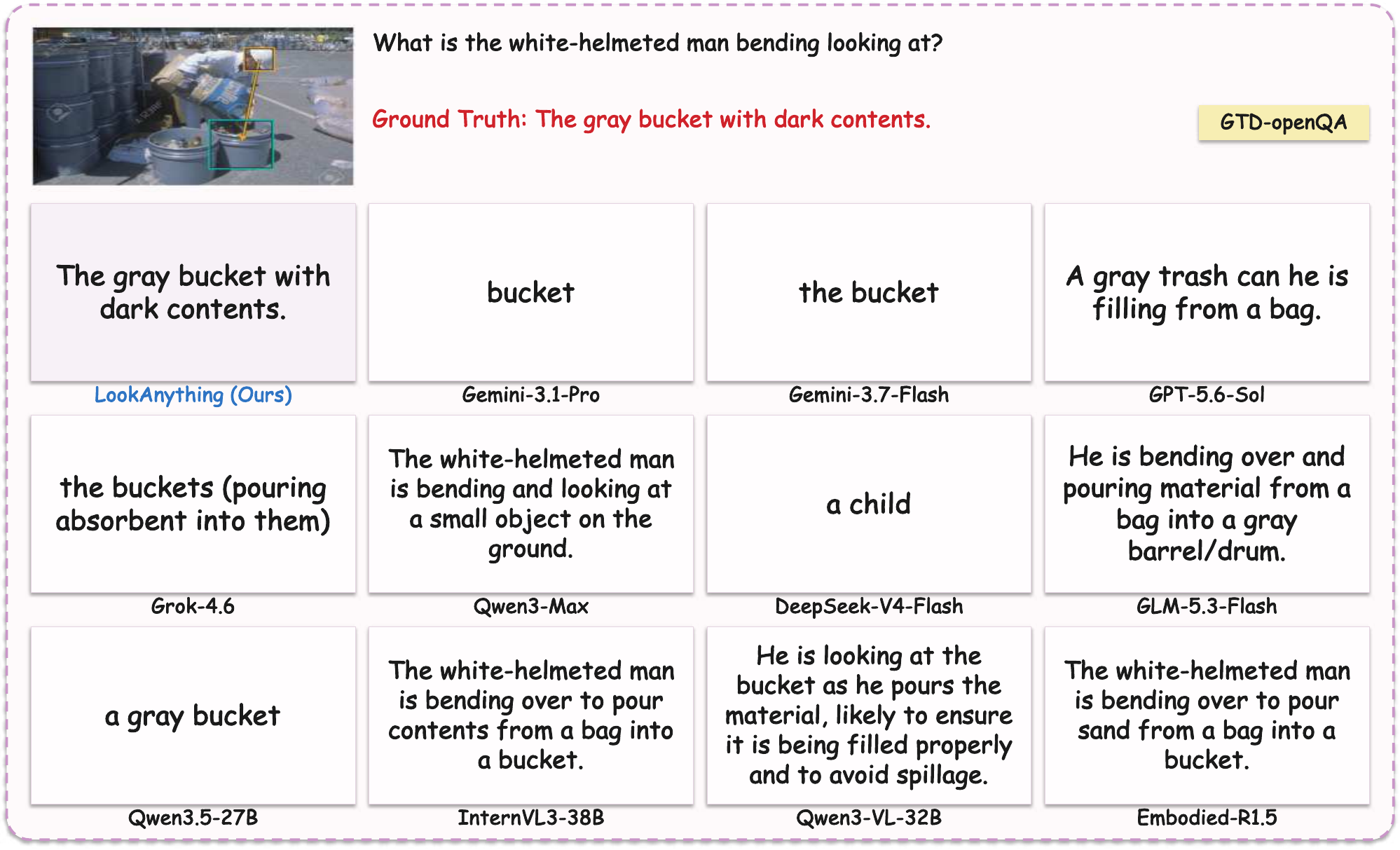}
      \caption{Qualitative Cases of GTD-OpenQA.}
      \label{fig:qual_gtd_2}
\end{figure}

\begin{figure}[p]
  \centering
      \includegraphics[width=0.9\textwidth]{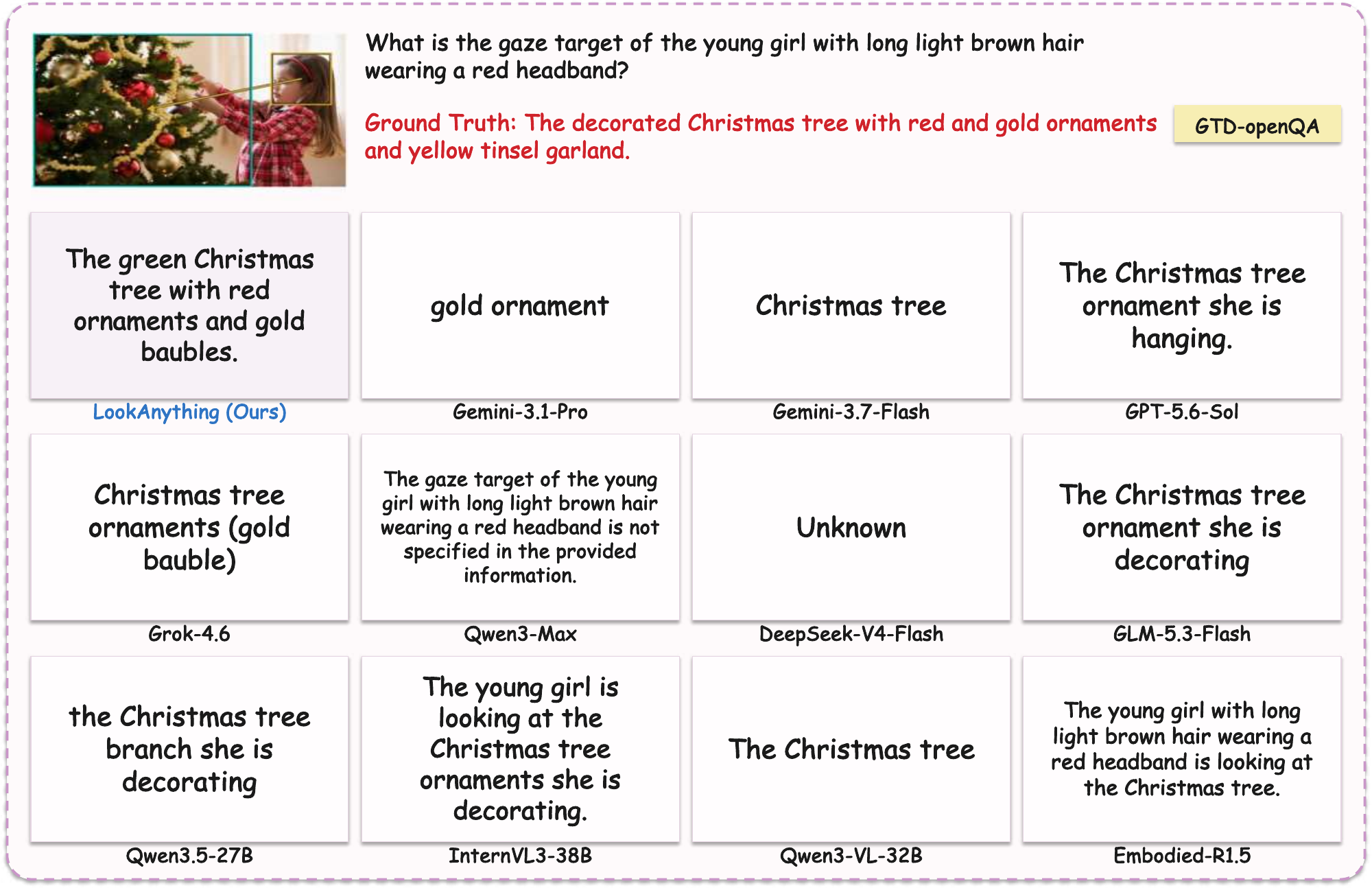}
      \caption{Qualitative Cases of GTD-OpenQA.}
      \label{fig:qual_gtd_3}
\end{figure}

\begin{figure}[p]
  \centering
      \includegraphics[width=1.0\textwidth]{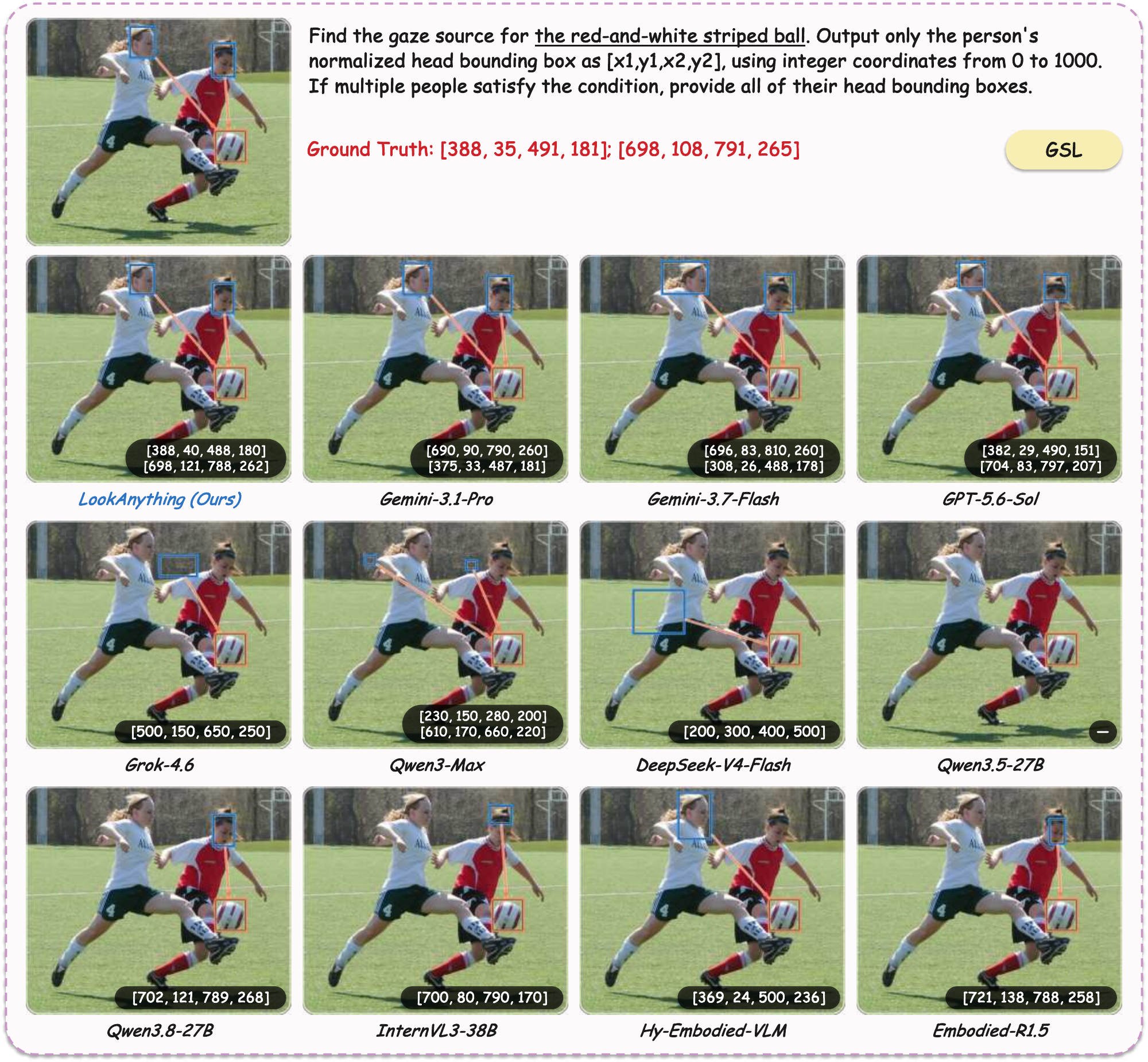}
      \caption{Qualitative Cases of GSL.}
      \label{fig:qual_gsl_2}
\end{figure}

\begin{figure}[p]
  \centering
      \includegraphics[width=1.0\textwidth]{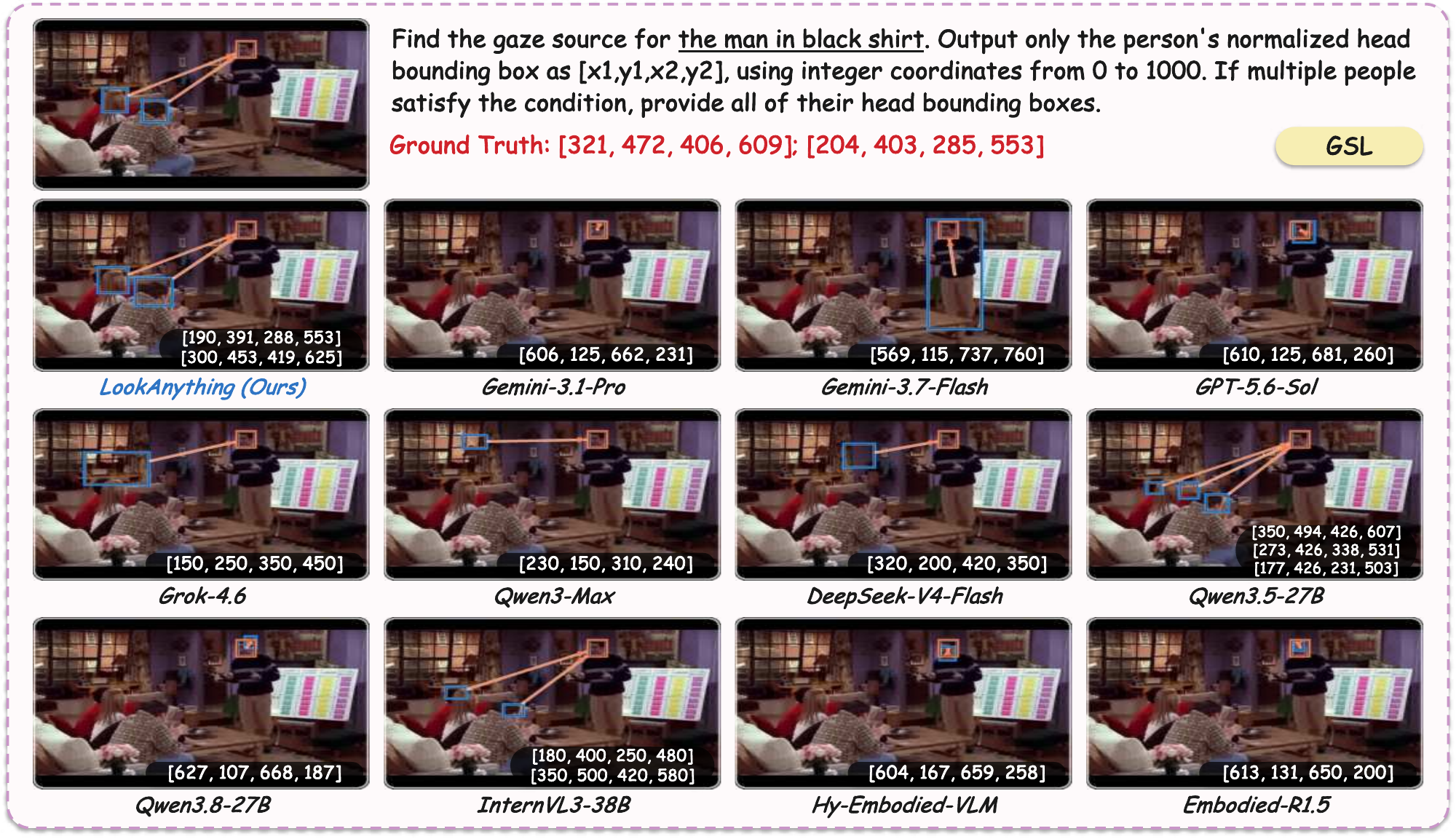}
      \caption{Qualitative Cases of GSL.}
      \label{fig:qual_gsl_3}
\end{figure}

\begin{figure}[p]
  \centering
      \includegraphics[width=1.0\textwidth]{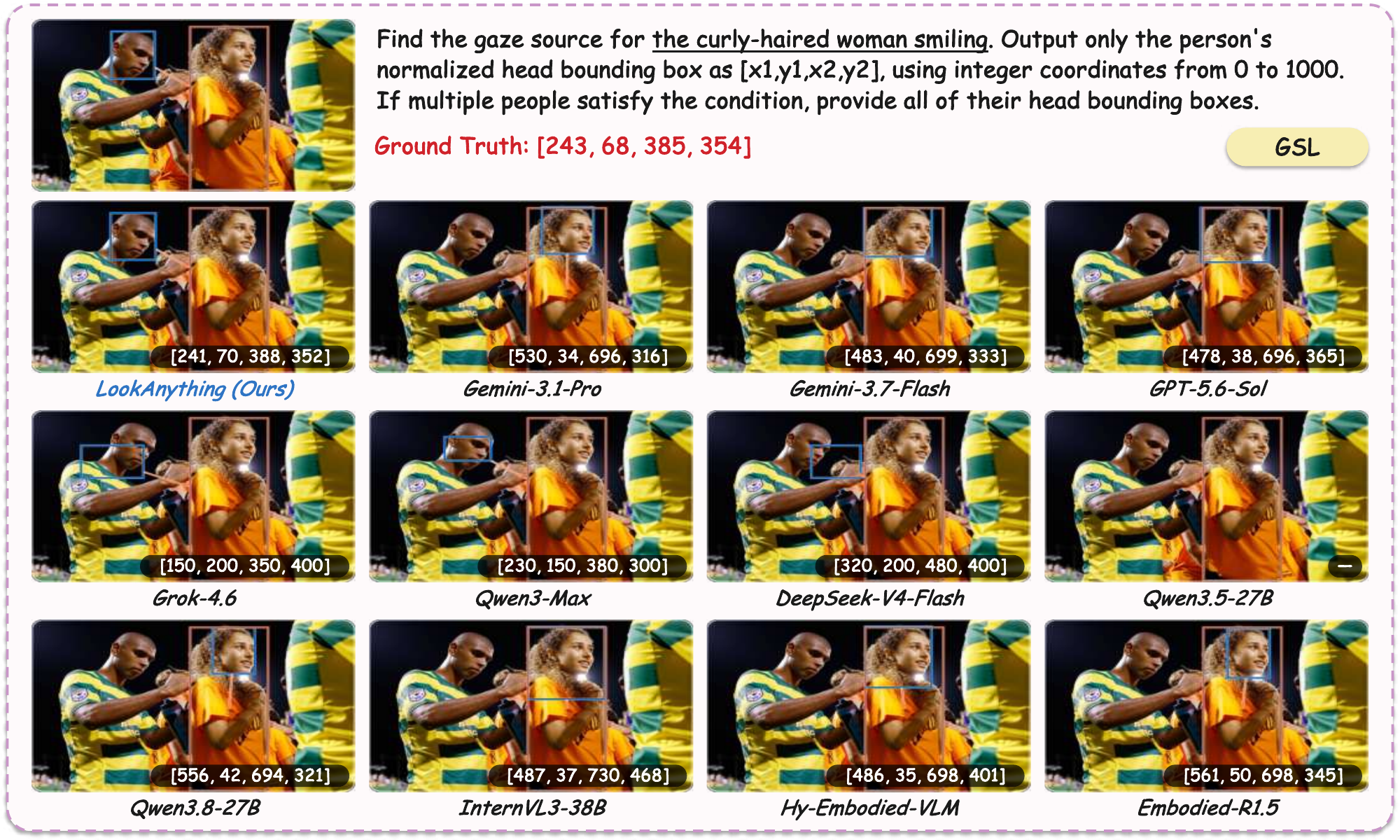}
      \caption{Qualitative Cases of GSL.}
      \label{fig:qual_gsl_4}
\end{figure}

\begin{figure}[p]
  \centering
      \includegraphics[width=1.0\textwidth]{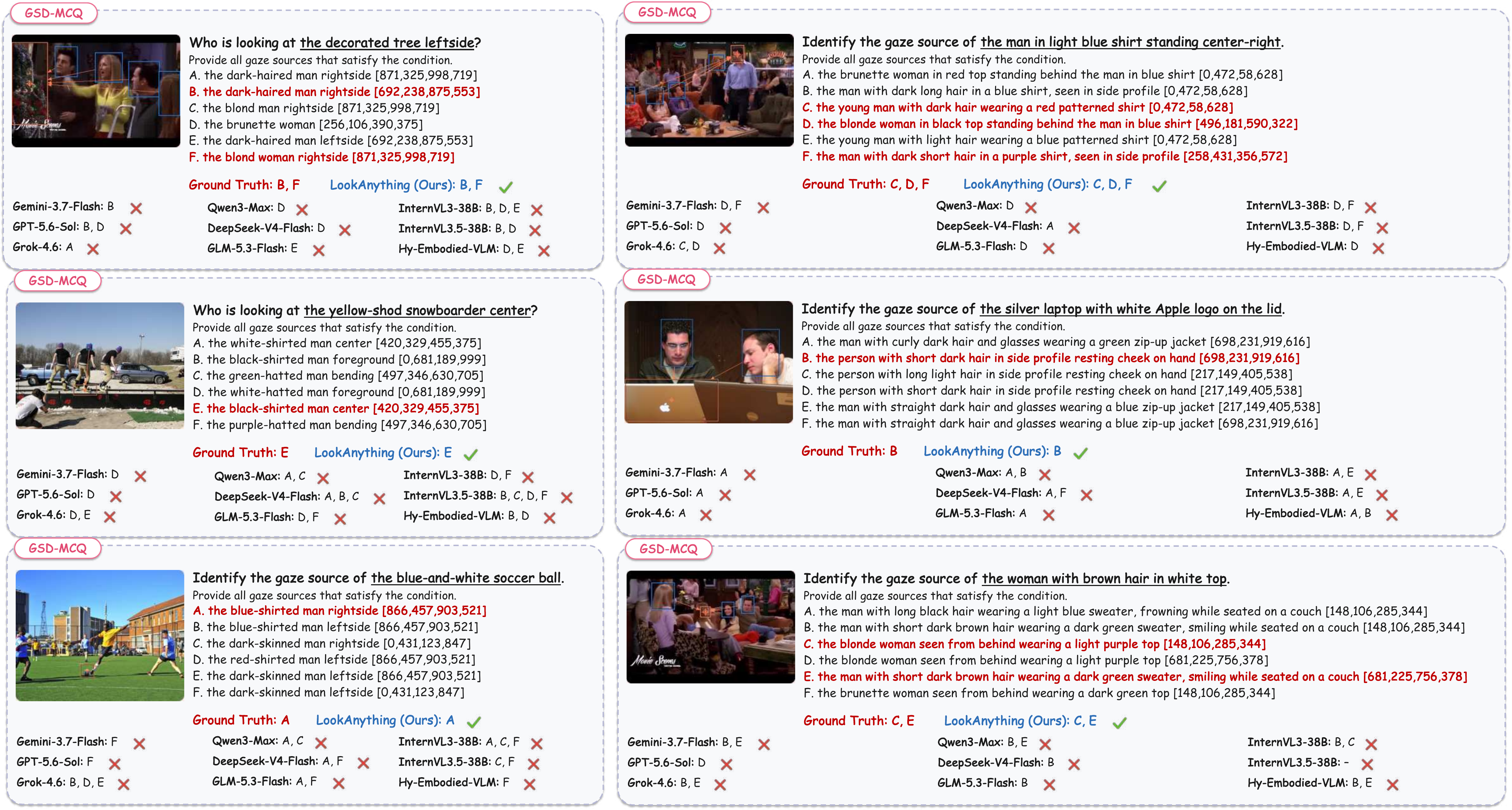}
      \caption{Qualitative Cases of GSD.}
      \label{fig:qual_gsd}
\end{figure}

\begin{figure}[p]
\centering
    \includegraphics[width=1.0\textwidth]{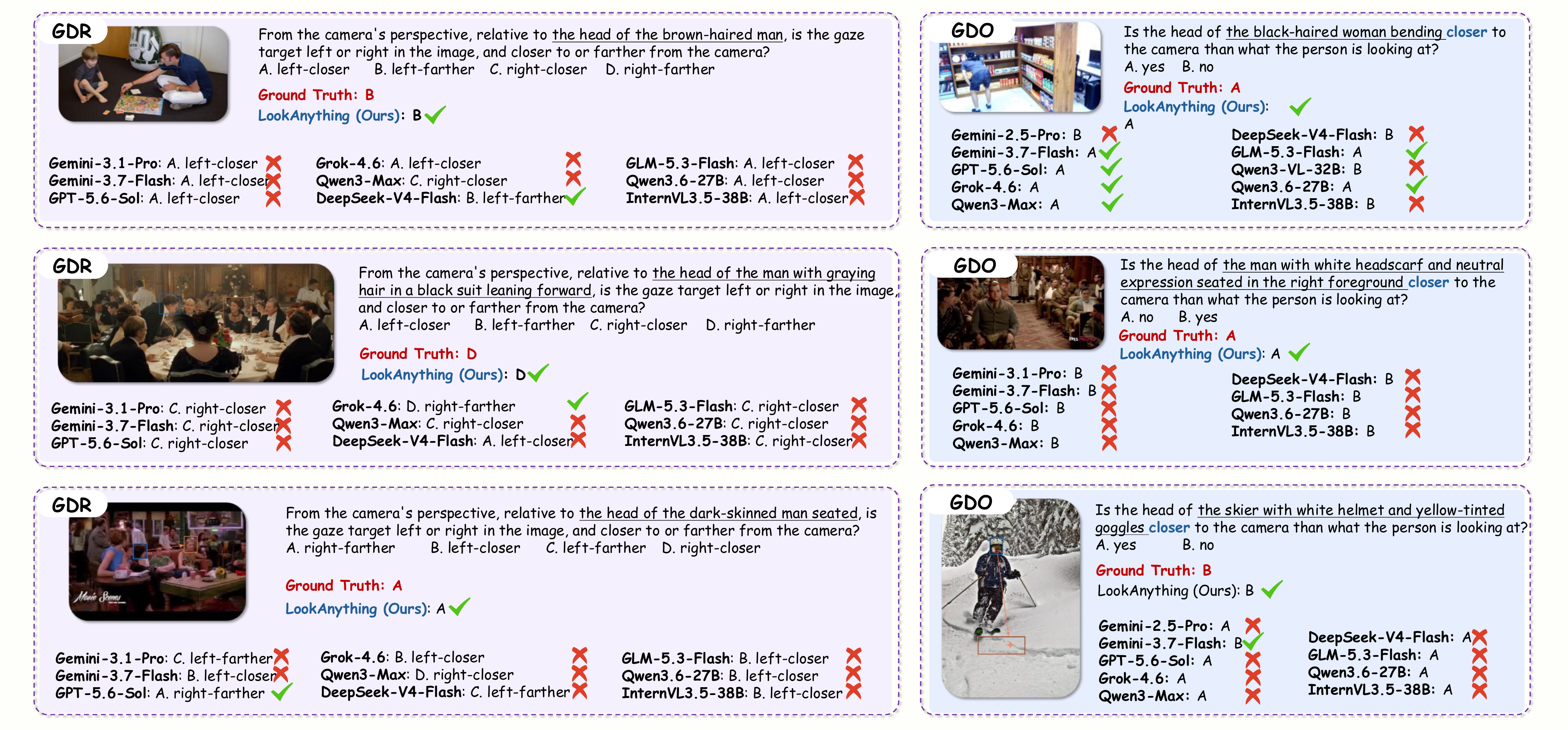}
    \caption{Qualitative Cases of GDO/GDR.}
    \label{fig:qual_gdo_gdr}
\end{figure}

\begin{figure}[p]
\centering
    \includegraphics[width=1.0\textwidth]{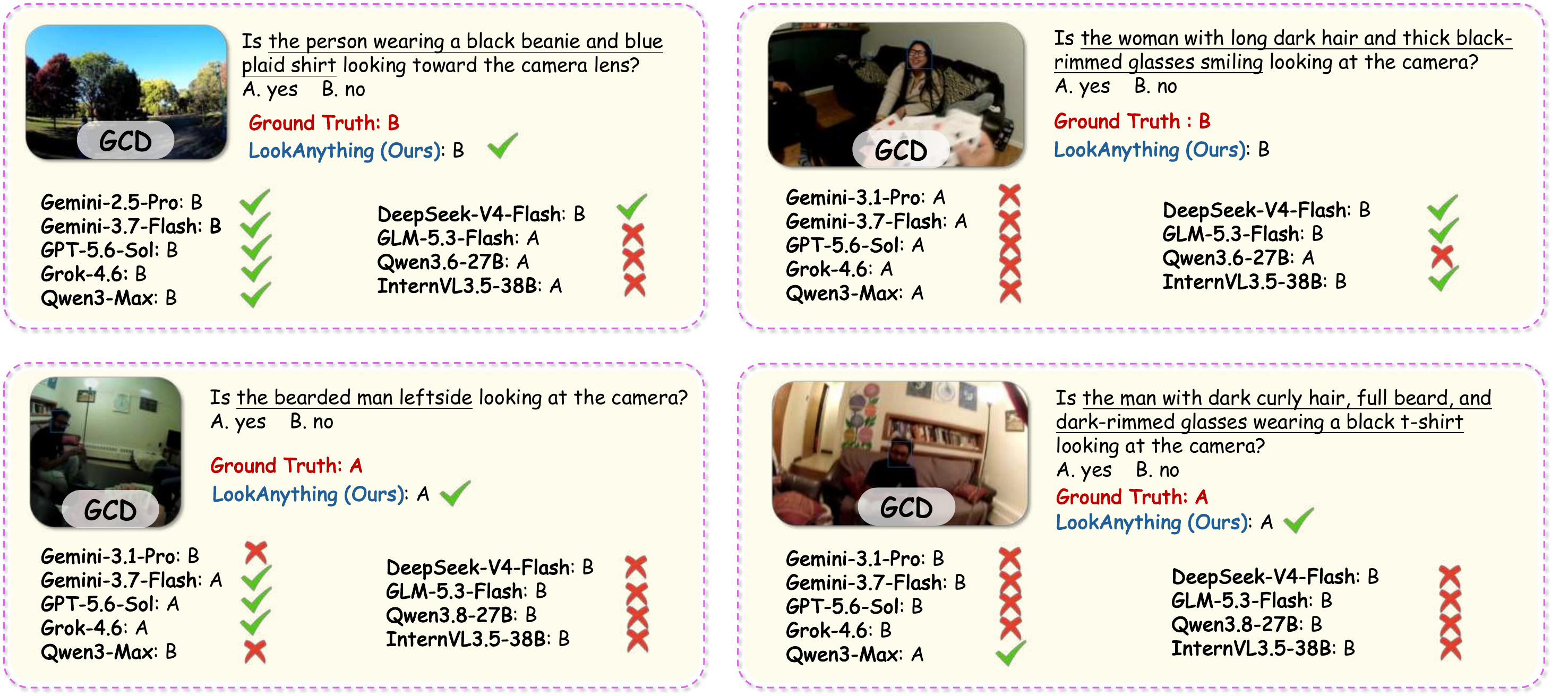}
    \caption{Qualitative Cases of GCD.}
    \label{fig:qual_gcd}
\end{figure}

\begin{figure}[p]
\centering
    \includegraphics[width=1.0\textwidth]{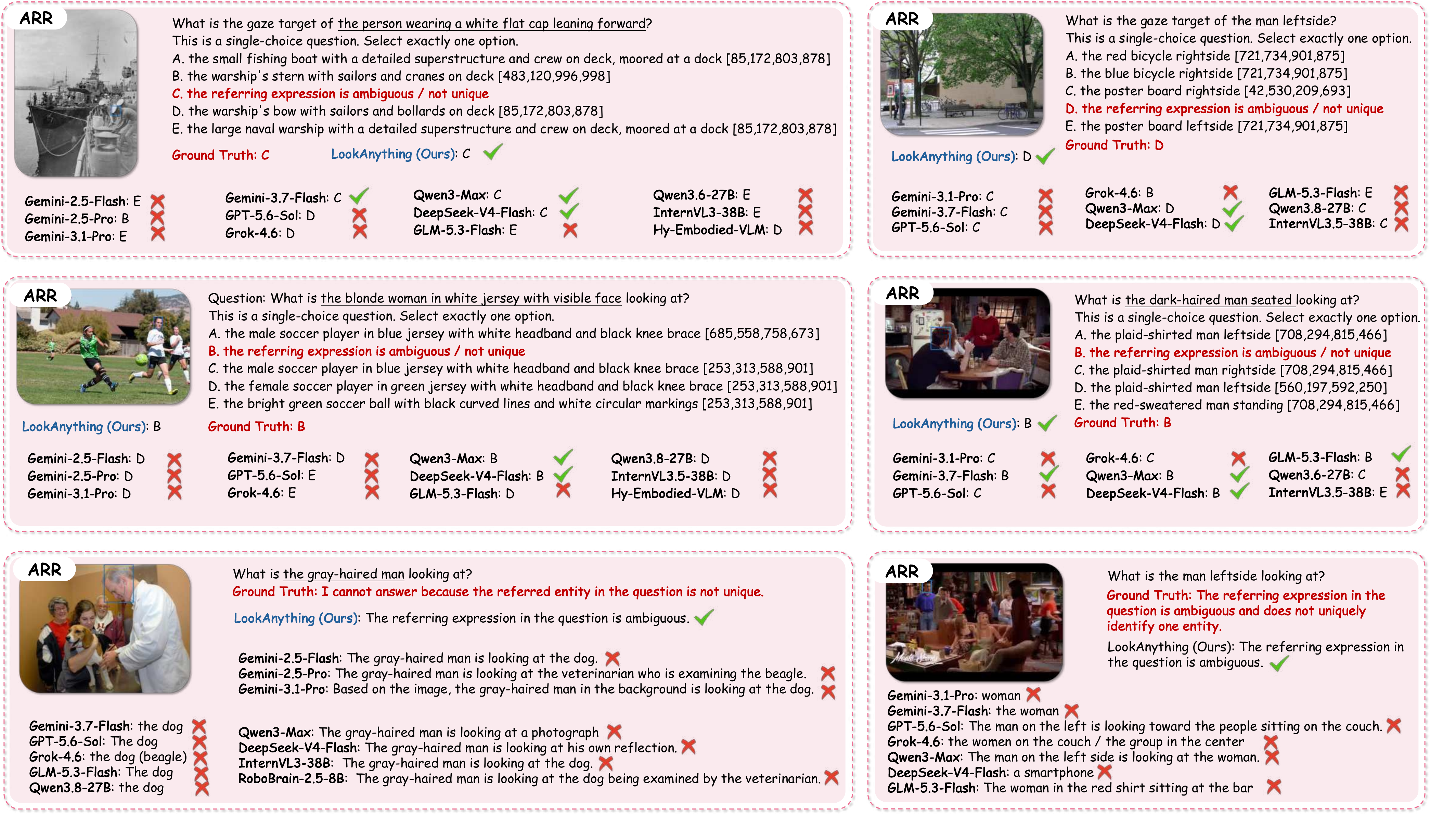}
    \caption{Qualitative Cases of ARR.}
    \label{fig:qual_arr}
\end{figure}

\begin{figure}[p]
  \centering
      \includegraphics[width=1.0\textwidth]{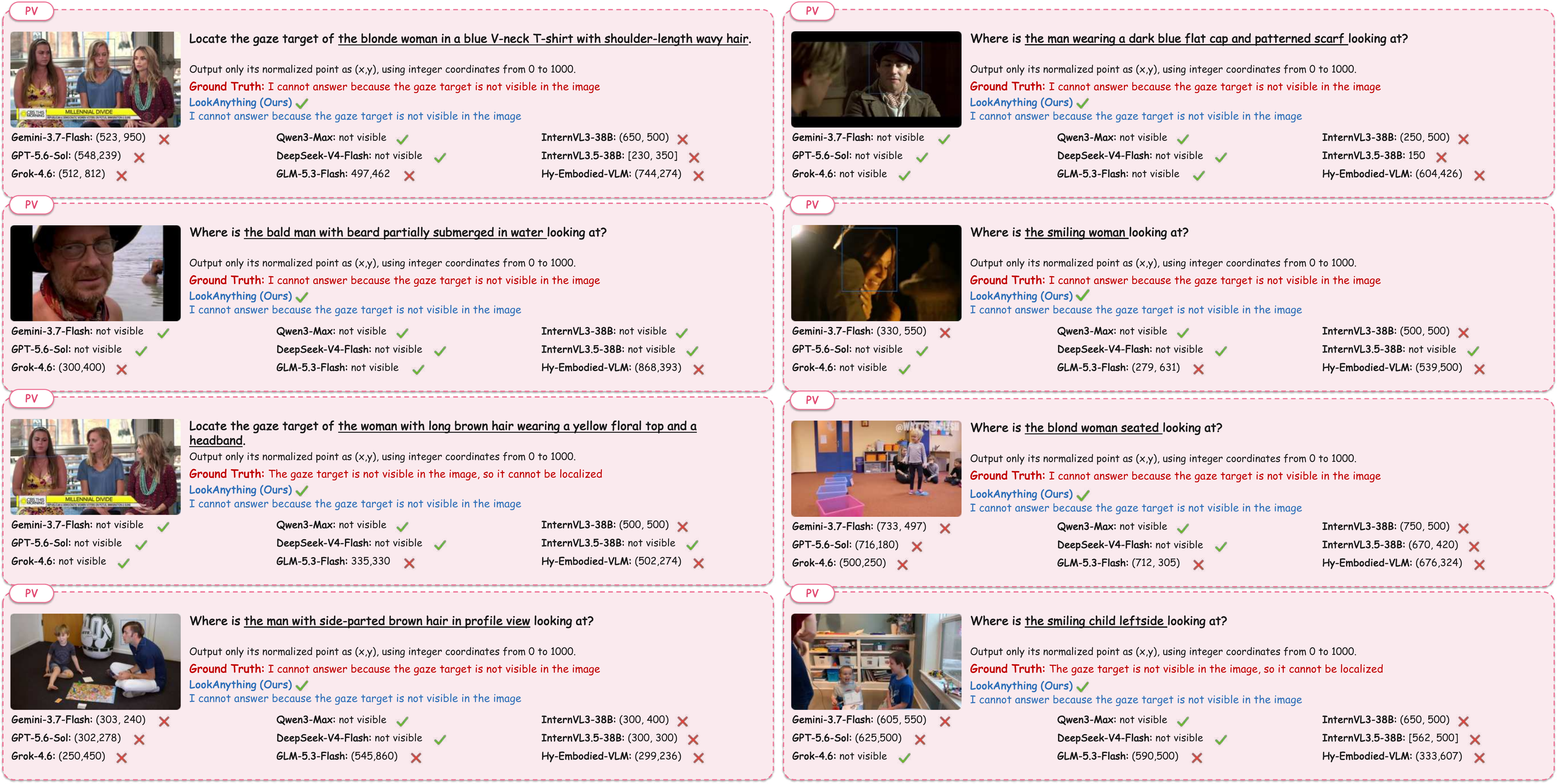}
      \caption{Qualitative Cases of PV.}
      \label{fig:qual_pv}
\end{figure}

In this section, we present additional qualitative visualization results on HoliGaze-Bench across diverse gaze-centric tasks. Figures~\ref{fig:qual_gtlt2p_1} to \ref{fig:qual_pv} showcase the model's capabilities across the 9 defined tasks. 

Specifically, Figures~\ref{fig:qual_gtlt2p_1} to \ref{fig:qual_gtlt2b_2} demonstrate the precise grounding capabilities in Gaze Target Localization (GTL) across both point and box formats, handling complex spatial relationships and distant targets. Figures~\ref{fig:qual_gtd_1} to \ref{fig:qual_gtd_3} visualize Gaze Target Description (GTD), where the model accurately attributes semantic meaning to the localized targets. Conversely, Figures~\ref{fig:qual_gsl_2} to \ref{fig:qual_gsd} illustrate Gaze Source Localization (GSL) and Description (GSD), highlighting the model's ability to trace back from objects to the attending human sources, even in multi-person shared attention scenarios. 

Furthermore, Figure~\ref{fig:qual_gdo_gdr} demonstrates strong geometric reasoning in the Gaze Depth Orientation (GDO) and Gaze Direction Recognition (GDR) tasks, where the model successfully discerns fine-grained 3D spatial arrangements and camera-relative directions. Figure~\ref{fig:qual_gcd} shows accurate Gaze Contact Detection (GCD), which is crucial for identifying social intent. Finally, Figures~\ref{fig:qual_arr} and \ref{fig:qual_pv} highlight the model's robustness in answering ill-posed queries via Ambiguous Reference Recognition (ARR) and Presence Verification (PV), appropriately issuing refusals instead of hallucinating answers. 

Overall, as observed in these diverse examples, LookAnything effectively and robustly processes complex multimodal queries, grounding visual entities precisely while providing accurate geometric reasoning and semantic interpretations.


\clearpage
\setlength{\bibsep}{5pt}
\bibliographystyle{unsrtnat}
\bibliography{main}

@article{recasens2015they,
  title={Where are they looking?},
  author={Recasens, Adria and Khosla, Aditya and Vondrick, Carl and Torralba, Antonio},
  journal={Advances in neural information processing systems},
  volume={28},
  year={2015}
}

@inproceedings{lian2018believe,
  title={Believe it or not, we know what you are looking at!},
  author={Lian, Dongze and Yu, Zehao and Gao, Shenghua},
  booktitle={Asian Conference on Computer Vision},
  pages={35--50},
  year={2018},
  organization={Springer}
}

@inproceedings{fan2018inferring,
  title={Inferring shared attention in social scene videos},
  author={Fan, Lifeng and Chen, Yixin and Wei, Ping and Wang, Wenguan and Zhu, Song-Chun},
  booktitle={2018 IEEE/CVF Conference on Computer Vision and Pattern Recognition},
  pages={6460--6468},
  year={2018},
  organization={IEEE}
}

@inproceedings{saran2018human,
  title={Human gaze following for human-robot interaction},
  author={Saran, Akanksha and Majumdar, Srinjoy and Short, Elaine Schaertl and Thomaz, Andrea and Niekum, Scott},
  booktitle={2018 IEEE/RSJ International Conference on Intelligent Robots and Systems (IROS)},
  pages={8615--8621},
  year={2018},
  organization={IEEE}
}

@inproceedings{chong2020detecting,
  title={Detecting attended visual targets in video},
  author={Chong, Eunji and Wang, Yongxin and Ruiz, Nataniel and Rehg, James M},
  booktitle={2020 IEEE/CVF Conference on Computer Vision and Pattern Recognition (CVPR)},
  pages={5395--5405},
  year={2020},
  organization={IEEE}
}

@inproceedings{fang2021dual,
  title={Dual attention guided gaze target detection in the wild},
  author={Fang, Yi and Tang, Jiapeng and Shen, Wang and Shen, Wei and Gu, Xiao and Song, Li and Zhai, Guangtao},
  booktitle={Proceedings of the IEEE/CVF conference on computer vision and pattern recognition},
  pages={11390--11399},
  year={2021}
}

@inproceedings{tomas2021goo,
  title={Goo: A dataset for gaze object prediction in retail environments},
  author={Tomas, Henri and Reyes, Marcus and Dionido, Raimarc and Ty, Mark and Mirando, Jonric and Casimiro, Joel and Atienza, Rowel and Guinto, Richard},
  booktitle={2021 IEEE/CVF Conference on Computer Vision and Pattern Recognition Workshops (CVPRW)},
  pages={3119--3127},
  year={2021},
  organization={IEEE}
}

@inproceedings{gupta2022modular,
  title={A modular multimodal architecture for gaze target prediction: Application to privacy-sensitive settings},
  author={Gupta, Anshul and Tafasca, Samy and Odobez, Jean-Marc},
  booktitle={2022 IEEE/CVF Conference on Computer Vision and Pattern Recognition Workshops (CVPRW)},
  pages={5037--5046},
  year={2022},
  organization={IEEE}
}

@inproceedings{tu2022end,
  title={End-to-end human-gaze-target detection with transformers},
  author={Tu, Danyang and Min, Xiongkuo and Duan, Huiyu and Guo, Guodong and Zhai, Guangtao and Shen, Wei},
  booktitle={2022 IEEE/CVF Conference on Computer Vision and Pattern Recognition (CVPR)},
  pages={2192--2200},
  year={2022},
  organization={IEEE}
}

@inproceedings{tonini2022multimodal,
  title={Multimodal across domains gaze target detection},
  author={Tonini, Francesco and Beyan, Cigdem and Ricci, Elisa},
  booktitle={Proceedings of the 2022 International Conference on Multimodal Interaction},
  pages={420--431},
  year={2022}
}

@inproceedings{grauman2022ego4d,
  title={Ego4d: Around the world in 3,000 hours of egocentric video},
  author={Grauman, Kristen and Westbury, Andrew and Byrne, Eugene and Chavis, Zachary and Furnari, Antonino and Girdhar, Rohit and Hamburger, Jackson and Jiang, Hao and Liu, Miao and Liu, Xingyu and others},
  booktitle={Proceedings of the IEEE/CVF conference on computer vision and pattern recognition},
  pages={18995--19012},
  year={2022}
}

@inproceedings{kasahara2022look,
  title={Look both ways: Self-supervising driver gaze estimation and road scene saliency},
  author={Kasahara, Isaac and Stent, Simon and Park, Hyun Soo},
  booktitle={European Conference on Computer Vision},
  pages={126--142},
  year={2022},
  organization={Springer}
}

@inproceedings{miao2023patch,
  title={Patch-level gaze distribution prediction for gaze following},
  author={Miao, Qiaomu and Hoai, Minh and Samaras, Dimitris},
  booktitle={2023 IEEE/CVF Winter Conference on Applications of Computer Vision (WACV)},
  pages={880--889},
  year={2023},
  organization={IEEE}
}

@inproceedings{tafasca2023childplay,
  title={Childplay: A new benchmark for understanding children’s gaze behaviour},
  author={Tafasca, Samy and Gupta, Anshul and Odobez, Jean-Marc},
  booktitle={2023 IEEE/CVF International Conference on Computer Vision (ICCV)},
  pages={20878--20889},
  year={2023},
  organization={IEEE}
}

@inproceedings{tafasca2024sharingan,
  title={Sharingan: A transformer architecture for multi-person gaze following},
  author={Tafasca, Samy and Gupta, Anshul and Odobez, Jean-Marc},
  booktitle={2024 IEEE/CVF Conference on Computer Vision and Pattern Recognition (CVPR)},
  pages={2008--2017},
  year={2024},
  organization={IEEE}
}

@article{song2024vitgaze,
  title={Vitgaze: gaze following with interaction features in vision transformers},
  author={Song, Yuehao and Wang, Xinggang and Yao, Jingfeng and Liu, Wenyu and Zhang, Jinglin and Xu, Xiangmin},
  journal={Visual Intelligence},
  volume={2},
  number={1},
  pages={31},
  year={2024},
  publisher={Springer}
}

@article{tafasca2024toward,
  title={Toward semantic gaze target detection},
  author={Tafasca, Samy and Gupta, Anshul and Bros, Victor and Odobez, Jean-Marc},
  journal={Advances in neural information processing systems},
  volume={37},
  pages={121422--121448},
  year={2024}
}

@inproceedings{lin2025gazehta,
  title={Gazehta: End-to-end gaze target detection with head-target association},
  author={Lin, Zhi-Yi and Chew, Jouh Yeong and van Gemert, Jan and Zhang, Xucong},
  booktitle={2025 IEEE International Conference Robotics and Automation (ICRA)},
  pages={9447--9454},
  year={2025},
  organization={IEEE}
}

@inproceedings{ryan2025gaze,
  title={Gaze-lle: Gaze target estimation via large-scale learned encoders},
  author={Ryan, Fiona and Bati, Ajay and Lee, Sangmin and Bolya, Daniel and Hoffman, Judy and Rehg, James M},
  booktitle={2025 IEEE/CVF Conference on Computer Vision and Pattern Recognition (CVPR)},
  pages={28874--28884},
  year={2025},
  organization={IEEE}
}

@article{bai2025qwen3,
  title={Qwen3-vl technical report},
  author={Bai, Shuai and Cai, Yuxuan and Chen, Ruizhe and Chen, Keqin and Chen, Xionghui and Cheng, Zesen and Deng, Lianghao and Ding, Wei and Gao, Chang and Ge, Chunjiang and others},
  journal={arXiv preprint arXiv:2511.21631},
  year={2025}
}

@article{comanici2025gemini,
  title={Gemini 2.5: Pushing the frontier with advanced reasoning, multimodality, long context, and next generation agentic capabilities},
  author={Comanici, Gheorghe and Bieber, Eric and Schaekermann, Mike and Pasupat, Ice and Sachdeva, Noveen and Dhillon, Inderjit and Blistein, Marcel and Ram, Ori and Zhang, Dan and Rosen, Evan and others},
  journal={arXiv preprint arXiv:2507.06261},
  year={2025}
}

@article{luo2025visual,
  title={Visual embodied brain: Let multimodal large language models see, think, and control in spaces},
  author={Luo, Gen and Yang, Ganlin and Gong, Ziyang and Chen, Guanzhou and Duan, Haonan and Cui, Erfei and Tong, Ronglei and Hou, Zhi and Zhang, Tianyi and Chen, Zhe and others},
  journal={arXiv preprint arXiv:2506.00123},
  year={2025}
}

@article{team2025robobrain,
  title={Robobrain 2.0 technical report},
  author={Team, BAAI RoboBrain and Cao, Mingyu and Tan, Huajie and Ji, Yuheng and Chen, Xiansheng and Lin, Minglan and Li, Zhiyu and Cao, Zhou and Wang, Pengwei and Zhou, Enshen and others},
  journal={arXiv preprint arXiv:2507.02029},
  year={2025}
}

@article{singh2025openai,
  title={Openai gpt-5 system card},
  author={Singh, Aaditya and Fry, Adam and Perelman, Adam and Tart, Adam and Ganesh, Adi and El-Kishky, Ahmed and McLaughlin, Aidan and Low, Aiden and Ostrow, AJ and Ananthram, Akhila and others},
  journal={arXiv preprint arXiv:2601.03267},
  year={2025}
}

@inproceedings{zhou2025towards,
  title={Where, what, why: Towards explainable driver attention prediction},
  author={Zhou, Yuchen and Tang, Jiayu and Xiao, Xiaoyan and Lin, Yueyao and Liu, Linkai and Guo, Zipeng and Fei, Hao and Xia, Xiaobo and Gou, Chao},
  booktitle={2025 IEEE/CVF International Conference on Computer Vision (ICCV)},
  pages={2675--2685},
  year={2025},
  organization={IEEE}
}

@inproceedings{yuan2026embodiedr1,
  title={Embodied-r1: Reinforced embodied reasoning for general robotic manipulation},
  author={Yuan, Yifu and Cui, Haiqin and Huang, Yaoting and Chen, Yibin and Ni, Fei and Dong, Zibin and Li, Pengyi and Zheng, Yan and Tang, Hongyao and Hao, Jianye},
  booktitle={International Conference on Learning Representations},
  volume={2026},
  pages={8984--9009},
  year={2026}
}

@article{dai2026gazemoe,
  title={GazeMoE: Perception of Gaze Target with Mixture-of-Experts},
  author={Dai, Zhuangzhuang and Lu, Zhongxi and Zakka, Vincent G and Manso, Luis J and Calero, Jose M Alcaraz and Li, Chen},
  journal={arXiv preprint arXiv:2603.06256},
  year={2026}
}

@article{wang2026enhancing,
  title={Enhancing Gaze Reasoning in Vision Foundation Models for Gaze Following},
  author={Wang, Shijing and Huang, Yaping and Cui, Chaoqun and Wong, David and Cheng, Yihua and Neophytou, Alexandros and Chang, Hyung Jin},
  journal={arXiv preprint arXiv:2605.22607},
  year={2026}
}

@inproceedings{cao2026gaze,
  title={Gaze Target Estimation Anywhere with Concepts},
  author={Cao, Xu and Yang, Houze and Gunda, Vipin and Zhou, Zhongyi and Xu, Tianyu and Kowdle, Adarsh and Kim, Inki and Rehg, James M},
  booktitle={Proceedings of the IEEE/CVF Conference on Computer Vision and Pattern Recognition},
  pages={31304--31315},
  year={2026}
}

@inproceedings{mathew2026gazevlm,
  title={Gazevlm: A vision-language model for multi-task gaze understanding},
  author={Mathew, Athul M and Hermassi, Haithem and Kadavil, Thariq and Khan, Arshad Ali},
  booktitle={Proceedings of the IEEE/CVF Conference on Computer Vision and Pattern Recognition},
  pages={9241--9250},
  year={2026}
}

@article{mi2026multi,
  title={Multi-scale Object-Aware Gaze Estimation via Geometric Reasoning},
  author={Mi, Jiajie and Liu, Xinyu and Song, Mengke and Chen, Chenglizhao},
  journal={arXiv preprint arXiv:2606.29334},
  year={2026}
}

@article{zuo2026gaze2act,
  title={Gaze2act: Gaze-conditioned vision-language-action policies for interactive robot manipulation},
  author={Zuo, Kuangji and Li, Gen and Lyu, Bofan and Lu, Yanshuo and Ma, Boyu and Han, Shijia and Zhou, Xinyu and Yuan, Xichen and Zhou, Chuhao and Bai, Jiaqi and others},
  journal={arXiv preprint arXiv:2605.30282},
  year={2026}
}

@article{wang2025vl4gaze,
  title={VL4Gaze: Unleashing Vision-Language Models for Gaze Following},
  author={Wang, Shijing and Cui, Chaoqun and Huang, Yaping and Chang, Hyung Jin and Cheng, Yihua},
  journal={arXiv preprint arXiv:2512.20735},
  year={2025}
}

@article{tan2026robobrain,
  title={Robobrain 2.5: Depth in sight, time in mind},
  author={Tan, Huajie and Zhou, Enshen and Li, Zhiyu and Xu, Yijie and Ji, Yuheng and Chen, Xiansheng and Chi, Cheng and Wang, Pengwei and Jia, Huizhu and Ao, Yulong and others},
  journal={arXiv preprint arXiv:2601.14352},
  year={2026}
}

@article{dang2026rynnbrain,
  title={Rynnbrain: Open embodied foundation models},
  author={Dang, Ronghao and Guo, Jiayan and Hou, Bohan and Leng, Sicong and Li, Kehan and Li, Xin and Liu, Jiangpin and Mao, Yunxuan and Wang, Zhikai and Yuan, Yuqian and others},
  journal={arXiv preprint arXiv:2602.14979},
  year={2026}
}

@article{hao2025mimo,
  title={Mimo-embodied: X-embodied foundation model technical report},
  author={Hao, Xiaoshuai and Zhou, Lei and Huang, Zhijian and Hou, Zhiwen and Tang, Yingbo and Zhang, Lingfeng and Li, Guang and Lu, Zheng and Ren, Shuhuai and Meng, Xianhui and others},
  journal={arXiv preprint arXiv:2511.16518},
  year={2025}
}

@article{zhang2025pelican,
  title={Pelican-vl 1.0: A foundation brain model for embodied intelligence},
  author={Zhang, Yi and Liu, Che and Ren, Xiancong and Ni, Hanchu and Zhang, Shuai and Ding, Zeyuan and Hu, Jiayu and Shan, Hanzhe and Niu, Zhenwei and Liu, Zhaoyang and others},
  journal={arXiv preprint arXiv:2511.00108},
  year={2025}
}

@article{yuan2026embodied,
  title={Embodied-R1. 5: Evolving Physical Intelligence via Embodied Foundation Models},
  author={Yuan, Yifu and Huang, Yaoting and Yao, Xianze and Li, Yutong and Zhang, Shuoheng and Han, Linqi and Li, Pengyi and Sun, Jiangeng and Jia, Wenting and Zhang, Zhao and others},
  journal={arXiv preprint arXiv:2606.11324},
  year={2026}
}

@inproceedings{carion2026sam,
  title={Sam 3: Segment anything with concepts},
  author={Carion, Nicolas and Gustafson, Laura and Hu, Yuan-Ting and Debnath, Shoubhik and Hu, Ronghang and Suris Coll-Vinent, Didac and Ryali, Chaitanya and Alwala, Kalyan Vasudev and Khedr, Haitham and Huang, Andrew and others},
  booktitle={International conference on learning representations},
  volume={2026},
  pages={138846--138923},
  year={2026}
}

@article{lin2025depth,
  title={Depth anything 3: Recovering the visual space from any views},
  author={Lin, Haotong and Chen, Sili and Liew, Junhao and Chen, Donny Y and Li, Zhenyu and Shi, Guang and Feng, Jiashi and Kang, Bingyi},
  journal={arXiv preprint arXiv:2511.10647},
  year={2025}
}

@article{xu2026deepseek,
  title={Deepseek-v4: Towards highly efficient million-token context intelligence},
  author={Xu, Anyi and Lin, Bangcai and Xue, Bing and Wang, Bingxuan and Xu, Bingzheng and Wu, Bochao and Zhang, Bowei and Lin, Chaofan and Dong, Chen and Ling, Chenchen and others},
  journal={arXiv preprint arXiv:2606.19348},
  year={2026}
}

@article{zeng2026glm,
  title={Glm-5: from vibe coding to agentic engineering},
  author={Zeng, Aohan and Lv, Xin and Hou, Zhenyu and Du, Zhengxiao and Zheng, Qinkai and Chen, Bin and Yin, Da and Ge, Chendi and Huang, Chenghua and Xie, Chengxing and others},
  journal={arXiv preprint arXiv:2602.15763},
  year={2026}
}

@misc{xiaimodelcard_card-4p6-4cd2dc57,
  author       = {{xAI}},
  title        = {Model Card: {Grok} 4.6},
  year         = {2026},
  month        = aug,
  howpublished = {Model card},
  url          = {https://media.x.ai/v1/website/card-4p6-4cd2dc57.pdf},
  note         = {Revision 2026-08-17}
}

@inproceedings{he2022masked,
  title={Masked autoencoders are scalable vision learners},
  author={He, Kaiming and Chen, Xinlei and Xie, Saining and Li, Yanghao and Doll{\'a}r, Piotr and Girshick, Ross},
  booktitle={2022 IEEE/CVF conference on computer vision and pattern recognition (CVPR)},
  pages={15979--15988},
  year={2022},
  organization={IEEE}
}

@article{yang2025qwen3,
  title={Qwen3 technical report},
  author={Yang, An and Li, Anfeng and Yang, Baosong and Zhang, Beichen and Hui, Binyuan and Zheng, Bo and Yu, Bowen and Gao, Chang and Huang, Chengen and Lv, Chenxu and others},
  journal={arXiv preprint arXiv:2505.09388},
  year={2025}
}

@article{wang2025internvl3,
  title={Internvl3. 5: Advancing open-source multimodal models in versatility, reasoning, and efficiency},
  author={Wang, Weiyun and Gao, Zhangwei and Gu, Lixin and Pu, Hengjun and Cui, Long and Wei, Xingguang and Liu, Zhaoyang and Jing, Linglin and Ye, Shenglong and Shao, Jie and others},
  journal={arXiv preprint arXiv:2508.18265},
  year={2025}
}

@article{zhu2025internvl3,
  title={Internvl3: Exploring advanced training and test-time recipes for open-source multimodal models},
  author={Zhu, Jinguo and Wang, Weiyun and Chen, Zhe and Liu, Zhaoyang and Ye, Shenglong and Gu, Lixin and Tian, Hao and Duan, Yuchen and Su, Weijie and Shao, Jie and others},
  journal={arXiv preprint arXiv:2504.10479},
  year={2025}
}

@article{hu2021lora,
  title={Lora: Low-rank adaptation of large language models},
  author={Hu, Edward J and Shen, Yelong and Wallis, Phillip and Allen-Zhu, Zeyuan and Li, Yuanzhi and Wang, Shean and Wang, Lu and Chen, Weizhu},
  journal={arXiv preprint arXiv:2106.09685},
  year={2021}
}

@article{loshchilov2017decoupled,
  title={Decoupled weight decay regularization},
  author={Loshchilov, Ilya and Hutter, Frank},
  journal={arXiv preprint arXiv:1711.05101},
  year={2017}
}

@article{wang2026hy,
  title={Hy-Embodied-VLM-1.0: Efficient Physical-World Agents},
  author={Wang, Ziyi and Yu, Xumin and Rao, Yongming and Ling, Yonggen and Li, Yunheng and Wang, Oran and Gao, Mingqi and Zhou, Yuchen and Liang, Yves and Liu, Zuyan and others},
  journal={arXiv preprint arXiv:2607.12894},
  year={2026}
}

@article{zhou2025behavior,
  title={Behavior-aware knowledge-embedded model for driver attention prediction},
  author={Zhou, Yuchen and Gou, Chao and Guo, Zipeng and Cheng, Yihua and Chang, Hyung Jin},
  journal={IEEE Transactions on Circuits and Systems for Video Technology},
  volume={35},
  number={10},
  pages={10199--10212},
  year={2025},
  publisher={IEEE}
}

@inproceedings{zhou2024learning,
  title={Learning from observer gaze: Zero-shot attention prediction oriented by human-object interaction recognition},
  author={Zhou, Yuchen and Liu, Linkai and Gou, Chao},
  booktitle={2024 IEEE/CVF Conference on Computer Vision and Pattern Recognition (CVPR)},
  pages={28390--28400},
  year={2024},
  organization={IEEE}
}

@article{kleinke1986gaze,
  title={Gaze and eye contact: a research review.},
  author={Kleinke, Chris L},
  journal={Psychological bulletin},
  volume={100},
  number={1},
  pages={78},
  year={1986},
  publisher={American Psychological Association}
}

@article{frischen2007gaze,
  title={Gaze cueing of attention: visual attention, social cognition, and individual differences.},
  author={Frischen, Alexandra and Bayliss, Andrew P and Tipper, Steven P},
  journal={Psychological bulletin},
  volume={133},
  number={4},
  pages={694},
  year={2007},
  publisher={American Psychological Association}
}

@article{grossmann2017eyes,
  title={The eyes as windows into other minds: An integrative perspective},
  author={Grossmann, Tobias},
  journal={Perspectives on Psychological Science},
  volume={12},
  number={1},
  pages={107--121},
  year={2017},
  publisher={Sage Publications, Sage CA: Los Angeles, CA}
}

@article{kendon1967some,
  title={Some functions of gaze-direction in social interaction},
  author={Kendon, Adam},
  journal={Acta psychologica},
  volume={26},
  pages={22--63},
  year={1967},
  publisher={Elsevier}
}

@inproceedings{hu2023gfie,
  title={Gfie: A dataset and baseline for gaze-following from 2d to 3d in indoor environments},
  author={Hu, Zhengxi and Yang, Yuxue and Zhai, Xiaolin and Yang, Dingye and Zhou, Bohan and Liu, Jingtai},
  booktitle={2023 IEEE/CVF Conference on Computer Vision and Pattern Recognition (CVPR)},
  pages={8907--8916},
  year={2023},
  organization={IEEE}
}

@article{chen2026they,
  title={Where are they looking in the operating room?},
  author={Chen, Keqi and Baributsa, S{\'e}raphin and Schewski, Lilien and Srivastav, Vinkle and Mutter, Didier and Beldi, Guido and Keller, Sandra and Padoy, Nicolas},
  journal={International Journal of Computer Assisted Radiology and Surgery},
  pages={1--10},
  year={2026},
  publisher={Springer}
}

@article{peng2026eye,
  title={In the eye of mllm: Benchmarking egocentric video intent understanding with gaze-guided prompting},
  author={Peng, Taiying and Hua, Jiacheng and Liu, Miao and Lu, Feng},
  journal={Advances in Neural Information Processing Systems},
  volume={38},
  year={2026}
}

@inproceedings{tonini2023object,
  title={Object-aware gaze target detection},
  author={Tonini, Francesco and Dall'Asen, Nicola and Beyan, Cigdem and Ricci, Elisa},
  booktitle={Proceedings of the IEEE/CVF International Conference on Computer Vision},
  pages={21860--21869},
  year={2023}
}

@inproceedings{pani2026gaze,
  title={Gaze-Regularized Vision-Language-Action Models for Robotic Manipulation},
  author={Pani, Anupam and Yang, Yanchao},
  booktitle={Proceedings of the IEEE/CVF Conference on Computer Vision and Pattern Recognition (CVPR) Workshops},
  pages={11332--11342},
  year={2026}
}

@inproceedings{lai2026fam,
  title={FAM-HRI: Foundation-Model Assisted Multimodal Human Robot Interaction Combining Gaze and Speech},
  author={Lai, Yuzhi and Yuan, Shenghai and Li, Peizheng and Zhang, Boya and Kiefer, Benjamin and Deng, Tianchen and Zell, Andreas},
  booktitle={2026 IEEE International Conference on Robotics and Automation (ICRA)},
  pages={10521--10534},
  year={2026}
}

@inproceedings{tang2026letp,
  title={LETP: Coupling Attention Localization and Cognitive Reasoning for Ego-Centric Multi-Task Driving Scene Perception},
  author={Tang, Jiayu and Zhou, Yuchen and Xiong, Chen and Gou, Chao},
  booktitle={ICASSP 2026-2026 IEEE International Conference on Acoustics, Speech and Signal Processing (ICASSP)},
  pages={19797--19801},
  year={2026},
  organization={IEEE}
}

@article{tang2026spatialsv,
  title={SpatialSV: Internalizing Interpretable 3D Spatial Awareness in MLLMs via Task-Oriented Visual Supervision},
  author={Tang, Jiayu and Zhou, Yuchen and Gou, Chao},
  journal={arXiv preprint arXiv:2606.19915},
  year={2026}
}

@inproceedings{zhou2026logic,
  title={Logic unseen: Revealing the logical blindspots of vision-language models},
  author={Zhou, Yuchen and Tang, Jiayu and Yang, Shuo and Xiao, Xiaoyan and Dai, Yuqin and Yang, Wenhao and Gou, Chao and Xia, Xiaobo and Chua, Tat-Seng},
  booktitle={Proceedings of the AAAI Conference on Artificial Intelligence},
  volume={40},
  number={34},
  pages={29062--29070},
  year={2026}
}

\end{document}